\documentclass[10pt,journal,compsoc]{IEEEtran}

\ifCLASSOPTIONcompsoc
  \usepackage[nocompress]{cite}
\else
  \usepackage{cite}
\fi

\usepackage{mathptmx}
\usepackage{graphicx}
\usepackage{times}
\usepackage{caption}
\usepackage{morefloats}
\usepackage[tight,normalsize,sf,SF]{subfigure}

\usepackage{amsmath,amscd}
\usepackage{amsfonts}
\usepackage[ruled,vlined]{algorithm2e}
\usepackage{algorithmic}
\usepackage{epsfig}
\usepackage{enumerate}
\usepackage{multirow}
\usepackage{array}
\usepackage{diagrams}
\usepackage{diagbox}

\usepackage{color,soul} 

\makeatletter
 \def\SOUL@hlpreamble{%
 \setul{}{2.4ex}
 \let\SOUL@stcolor\SOUL@hlcolor
 \SOUL@stpreamble
 }
\makeatother

\begin{document}

\title{LMap: Shape-Preserving Local Mappings for Biomedical Visualization}

\author{Saad~Nadeem,~Xianfeng~Gu,~and~Arie~Kaufman,~\IEEEmembership{Fellow,~IEEE}
\IEEEcompsocitemizethanks{\IEEEcompsocthanksitem Saad Nadeem, Xianfeng Gu and Arie Kaufman are with the Department of Computer Science, Stony Brook University, Stony Brook, NY 11794-2424.
\protect\\
E-mail: \{sanadeem, gu, ari\}@cs.stonybrook.edu.
}
\thanks{}}

\IEEEtitleabstractindextext{
\begin{abstract}
Visualization of medical organs and biological structures is a challenging task because of their complex geometry and the resultant occlusions. Global spherical and planar mapping techniques simplify the complex geometry and resolve the occlusions to aid in visualization. However, while resolving the occlusions these techniques do not preserve the geometric context, making them less suitable for mission-critical biomedical visualization tasks. In this paper, we present a shape-preserving local mapping technique for resolving occlusions locally while preserving the overall geometric context. More specifically, we present a novel visualization algorithm, LMap, for conformally parameterizing and deforming a selected local region-of-interest (ROI) on an arbitrary surface. The resultant shape-preserving local mappings help to visualize complex surfaces while preserving the overall geometric context. The algorithm is based on the robust and efficient extrinsic Ricci flow technique, and uses the dynamic Ricci flow algorithm to guarantee the existence of a local map for a selected ROI on an arbitrary surface. We show the effectiveness and efficacy of our method in three challenging use cases: (1) multimodal brain visualization, (2) optimal coverage of virtual colonoscopy centerline flythrough, and (3) molecular surface visualization.
\end{abstract}

}

\maketitle

\IEEEdisplaynontitleabstractindextext

\IEEEpeerreviewmaketitle

\section{Introduction}

\begin{figure*}[t!]
\begin{center}
\begin{tabular}{ccccc}
\includegraphics[width=0.23\textwidth]{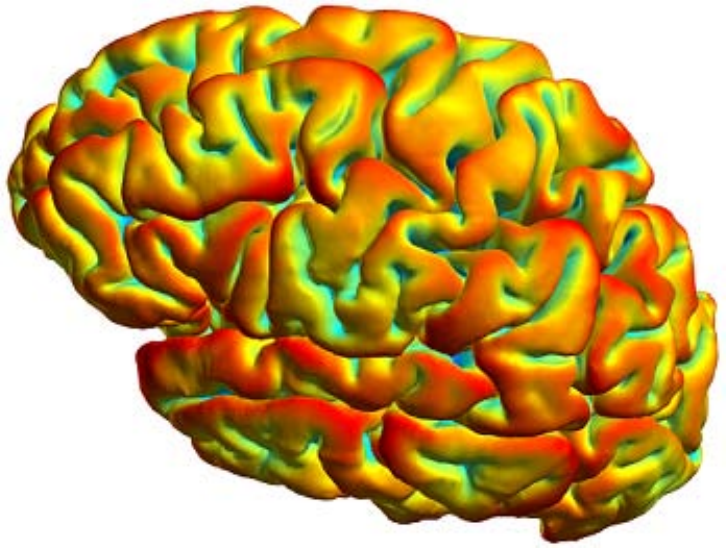}&
\includegraphics[width=0.175\textwidth]{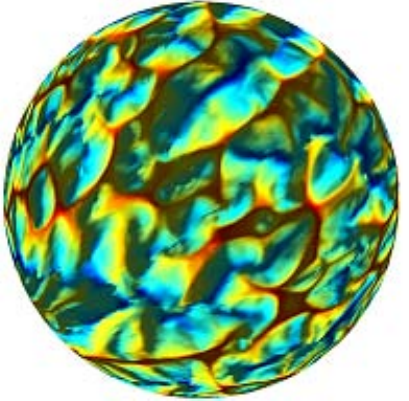}&
\includegraphics[width=0.23\textwidth]{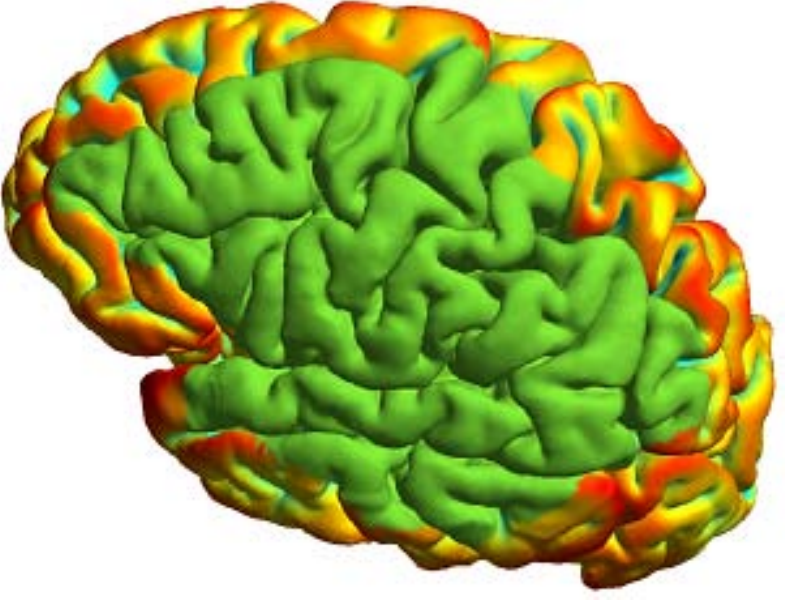}&
\includegraphics[width=0.23\textwidth]{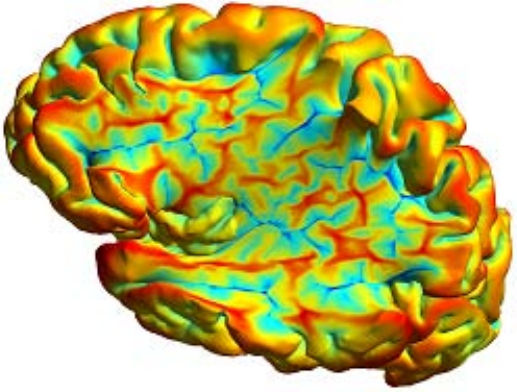}&
\includegraphics[width=0.027\textwidth]{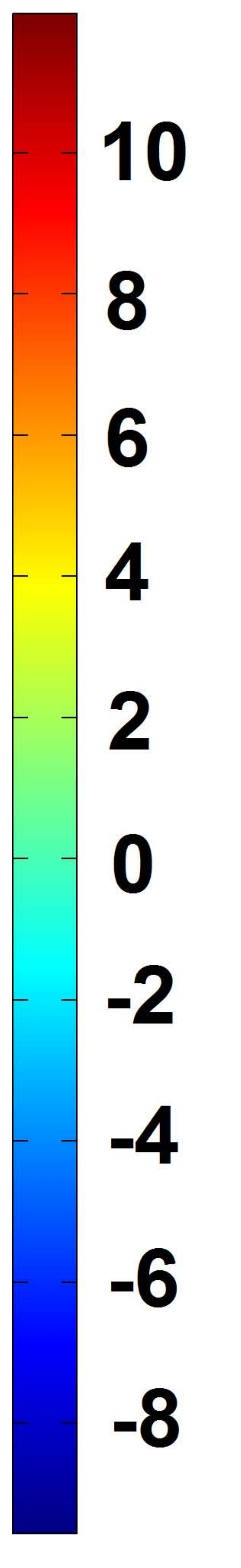}\\
(a) & (b) & (c) & (d) & \\
\includegraphics[width=0.23\textwidth]{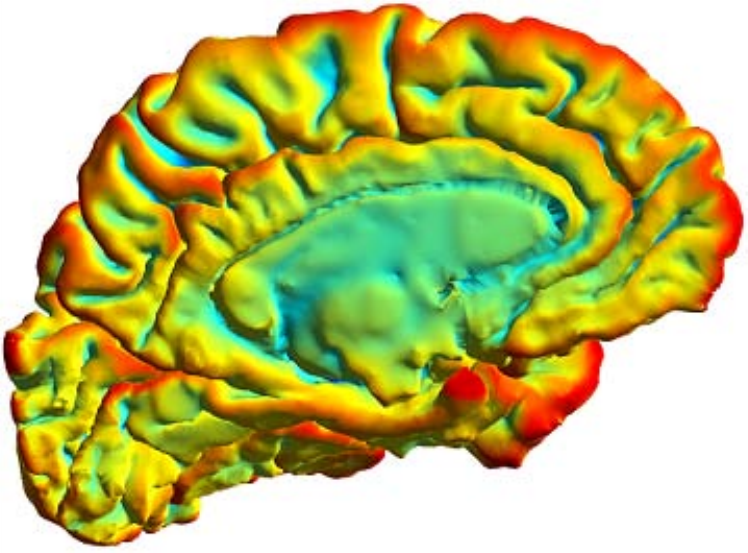}&
\includegraphics[width=0.175\textwidth]{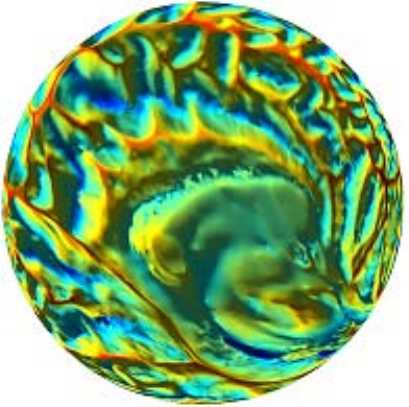}&
\includegraphics[width=0.23\textwidth]{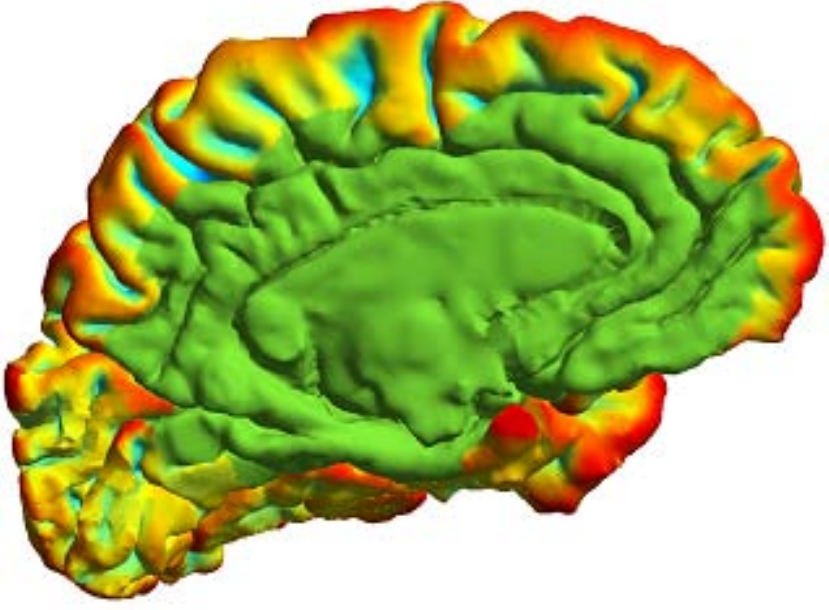}&
\includegraphics[width=0.23\textwidth]{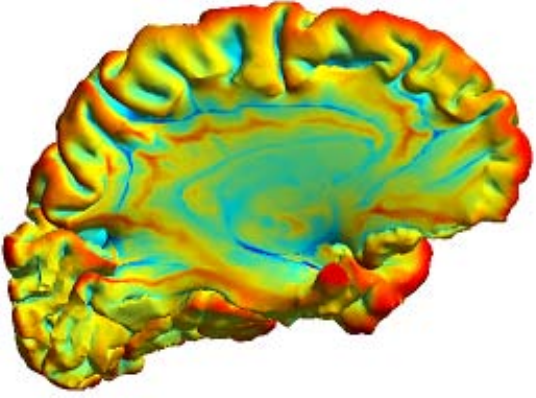}&
\includegraphics[width=0.027\textwidth]{figures/spherical_local_comparison/colormap}\\
(e) & (f) & (g) & (h) & \\
\end{tabular}
\end{center}
\vspace{-3mm}
\caption{Spherical and local mappings for brain cortical surface overlaid with mean curvature. (a) Front and (e) back of the brain cortical surface overlaid with mean curvature. (b) Front and (f) back of the angle-area preserving spherical mapping \cite{nadeem2016spherical}. (c) Front and (g) back selected regions-of-interest, colored in green. (d) Front and (h) back of the local mapping result from our algorithm. The dark blue lines in locally mapped surfaces (d) and (h) correspond to the sulcal fundi which have been shown to be robust and reproducible cortical landmarks for registration and group-level analysis. LMap computation for the selected ROIs (78952 vertices for the front ROI and 82291 vertices for the back ROI) took 2.7 seconds. 
\label{fig:spherical_local_comp}}
\end{figure*}

Visualization is critical for diagnosing various medical conditions (e.g., different types of cancer) through non-invasive imaging. However, 3D visualization of medical organs and various biological structures is challenging because of their complex 3D geometry and the resultant occlusions. For example, deep folds in the brain are occluded by the gyri (ridges) on the brain cortical surface (Figure~\ref{fig:spherical_local_comp}a). Global spherical and planar mapping techniques help to resolve these occlusions by mapping the complex 3D geometry to a canonical representation such as a sphere or a plane, respectively, with minimal angle and/or area distortions. The removal of both occlusions and the geometric context makes it difficult to correlate the features between the original and the mapped surfaces (Figure~\ref{fig:spherical_local_comp}b), making these techniques less useful for mission-critical biomedical visualization tasks. 

In effect, spherical and planar mesh mapping techniques are mostly restricted or relegated to segmentation, registration, and classification tasks. Moreover, these techniques assume a certain topology for the input surface, for example, spherical mapping techniques mandate a genus 0 closed surface input, whereas planar mapping techniques require a genus 0 surface with a boundary. These restrictions force the users to conform the data to the assumed input requirements via pre-processing (e.g., hole-filling, handle removal, feature extraction, etc.) to make these techniques work. For example, Krone et al. \cite{krone2017molecular} had to remove important channel information from the molecular surfaces for their spherical mapping technique to work, and Zeng et al. \cite{zeng2010supine} had to perform topological denoising (removal of fake handles) on the colon surface and extract consistent teniae coli (often not possible due to collapsed regions or segmentation artifacts \cite{nadeem2017corresponding}) for their flattening approach. Similar problems underline the multimodal domain where the information from all the modalities is computed in the native space and transformation of the original geometry (to a sphere or a plane) requires re-computation of all the modalities in the mapped space \cite{nadeem2016multimodal}. 

Ideally, a local mapping algorithm is required to deal with the shortcomings of the global mapping techniques. We prescribe the following desirable properties for our local mapping algorithm (modified from the ones outlined in Rocha et al. \cite{rocha2017decal} for local parameterization): (1) it should be independent of the surface parameterization since many scientific datasets are not represented parameterically; (2) it should not rely on any global information from the underlying mesh since this is expensive to compute and depends on the mesh type; (3) it should be efficient since we can require a large number of parameterizations and deformations of several regions-of-interest (ROIs) on the surface; (4) it should be angle-preserving (conformal) for the selected ROI and should not affect the rest of the surface; and (5) it should work for arbitrary surfaces (open/closed, smooth/convoluted, arbitrary-genus).

In this work, we present a technique, LMap, to parameterize and conformally deform local ROIs on arbitrary surfaces for occlusion-free visualization while preserving the overall geometric context (Figure~\ref{fig:spherical_local_comp}d). More specifically, for a selected ROI on a mesh, LMap uses a novel extrinsic Ricci flow algorithm for parameterizing and conformally deforming this selected ROI. The presented extrinsic Ricci flow algorithm entails the following steps: (1) compute local Gaussian curvature on the selected ROI, (2) obtain new edge lengths for the computed Gaussian curvature using dynamic Ricci flow \cite{nadeem2016spherical}, and (3) deform the position of each ROI vertex on the mesh, according to the new edge lengths, to obtain the conformally flattened ROI on the given mesh. The resultant output mesh resolves occlusions for selected ROIs while keeping the rest of the surface intact and untouched. Two important implications of the presented LMap algorithm are as follows: (1) the use of dynamic Ricci flow guarantees the existence of a stable (robust to perturbations in the selected ROIs) local map for a selected ROI on an arbitrary surface, and (2) the use of Gaussian curvature for the local ROI results in a stable local map, as opposed to the use of mean curvature for local ROIs which can lead to self-intersections (as shown in Figure~\ref{fig:spin_transformations}h).

In essence, LMap does not require any global information from the underlying mesh to deform the selected ROIs, and LMap is not dependent on the underlying topology or the type of the mesh. Moreover, unlike recent approaches which can only parameterize sphere-masked smooth local regions on arbitrary surfaces \cite{rocha2017decal}, LMap is applicable for smooth as well as convoluted local (sharp bend) ROIs on arbitrary surfaces. It also does not require any prior hole-filling or topological denoising of the input surface. Intuitively, LMap acts like a magic lens for selected ROIs on arbitrary surfaces, where it can be used to locally parameterize, and overlay textures and different types of information from multimodal and multivariate data. LMap also allows for local shape-preserving deformations for better visualization of complex ROIs. 

We show the efficacy and effectiveness of LMap in three challenging use cases: (1) multimodal brain visualization, (2) optimal coverage in virtual colonoscopy centerline flythroughs, and (3) molecular surface visualization. In the context of multimodal brain visualization, neuroscientists want to study specific cortical structures (which are highly convoluted and difficult to visualize in their original form) and the connectivity patterns between these structures. Unlike previous spherical parameterization methods \cite{nadeem2016multimodal,nadeem2016spherical} which mandate the transformation of the all the modalities to the underlying spherical space, LMap allows the simplified visualization of these specific structures and the connectivity patterns in the native space while preserving the overall geometric context. For virtual colonoscopy use case, radiologists want to localize and study the polyps (precursors of colon cancer) as accurately as possible. Current angle-preserving colon flattening approaches \cite{marino2011context,nadeem2017corresponding,zeng2010supine} induce considerable area distortion, require tenaie coli extraction (for cutting) and distort the geometry significantly, making the polyp localization difficult. In contrast, LMap can be used to only deform the detected haustral folds \cite{nadeem2017corresponding} in place while leaving the original geometry of the polyps and the overall colon intact, thus allowing for accurate polyp localization and measurements. Finally, in the context of molecular surface visualization, different cavities on the boundary and in the interior of the molecules need to be visualized as potential binding sites for other molecules. Current simplification approaches \cite{krone2017molecular}, in this context, use spherical parameterization to visualize the individual boundary cavities at the expense of removing the critical interior cavities and filling the resultant holes which induces considerable angle and area distortion during the parameterization; higher genus surfaces can induce considerably higher angle and area distortion due to higher loss in interior information. LMap, on the other hand, can be used for conformally deforming only the boundary cavities without removing the critical interior cavities and without affecting the rest of the molecular surface.

The main contributions of this paper are summarized as follows:
\begin{itemize}
  \item We present a novel local mapping algorithm, LMap, for visualizing local regions-of-interest on arbitrary genus closed/open surfaces. LMap deforms local regions efficiently with minimal angular distortion using robust and theoretically-sound extrinsic Ricci flow. Moreover, the use of established dynamic discrete surface Ricci flow guarantees the existence of the local for the prescribed Gaussian curvature, and greatly improves the stability of the algorithm.  
  \item We show the efficacy and effectiveness of LMap in three challenging use cases: (1) multimodal brain visualization, (2) optimal coverage in virtual colonoscopy centerline flythroughs, and (3) molecular surface visualization. We also compare the LMap results with the state-of-the-art visualization algorithms in these domains.
\end{itemize}

\section{Related Work}

Various global mapping techniques have been proposed in the past for medical visualizations, especially for colon, brain, and molecular surfaces. Most of these techniques require removal of segmentation artifacts, topological denoising (e.g., removal of fake handles), hole-filling (e.g., removal of channels mandates hole-filling in parameterized molecular surface visualization), and extraction of difficult-to-find features (e.g., consistent teniae coli extraction for colon flattening).

Colon flattening is a method in which the entire inner surface of the colon is displayed as a 2D image and thus all the occlusions are resolved but the geometric context is not preserved. The flattening requires the consistent teniae coli extraction which is often not possible and the folds in the resultant 2D image are cut at sharp bends \cite{nadeem2017corresponding}, reducing the overall usefulness of this flattened result in the medical visualization context. Initial attempts to flatten the colon surface include iterative methods based on electrical field lines \cite{wang1998gi}, cartographic \cite{paik2000visualization}, and cylindrical projections \cite{bartroli2001virtual,bartroli2001nonlinear}. However, most of these methods do not preserve the local shape. Conformal mapping, an approach where the local angles are preserved, has recently been used for colon flattening with promising results \cite{haker2000nondistorting,hong2006pipeline,gurijala2013colon,zeng2010supine}. The conformally flattened colon was used in the detection of colonic folds \cite{hong2006pipeline} and for supine-prone colon registration \cite{zeng2010supine}. To overcome the need of topological denoising for conformal colon flattening, Gurijala et al. \cite{gurijala2013colon} used a heat diffusion metric and obtained promising results. A geometric context-preserving colon flattening approach was presented by Marino et al. \cite{marino2011context} which was angle-preserving but induced large area distortion; however, the approach could not handle topological noise and required the prior consistent teniae coli extraction, mostly done manually. In our earlier work \cite{nadeem2017corresponding}, we presented an automatic method to segment folds and extract consistent teniae coli for flattening, doing away with the tedious cutting process. However, segmentation artifacts remain a problem since it is difficult to find a consistent cutting loop through the collapsed regions of the colon. Moreover, current colon flattening approaches are only angle-preserving and hence, the induced area-distortion can be large, making them less useful for polyp visualization on the resultant 2D flattened images. LMap can be used to only deform the detected haustral folds \cite{nadeem2017corresponding} in place while leaving the original geometry of the polyps and the overall colon intact, thus allowing for accurate polyp localization and measurements.

Conformal mapping techniques have also been used for brain visualization, such as the circle packing-based method \cite{hurdal2004cortical}, finite element method \cite{angenent1999conformal,ju2005quantitative,tosun2008geometry}, and spherical harmonic map method \cite{choi2015flash,gu2004genus,wang2005optimization}.
Zhao et al. \cite{zhao2013area} presented an angle-area preserving planar brain flattening approach using optimal mass transport. In our earlier work \cite{nadeem2016spherical}, we developed an angle-area preserving 3D spherical mapping technique for the brain and other genus 0 surfaces. The problem with these approaches is that because of the loss in geometric context, they are primarily used for registration, segmentation and classification tasks but not for brain visualization. In contrast, LMap allows simplified visualization of the local regions of interest while preserving the overall geometric context

In computational chemistry, molecules are often visually compared based on local surface attributes. However, to the best of our knowledge all the previous work for making this local comparison rely on global mapping techniques, which do not preserve the overall geometric context, and hence can make it difficult to make a finer comparison in the case of large regions with uniform attribute values. Previous works leveraged spherical mapping via deformable models \cite{polyansky2014preddimer}, spherical coordinates \cite{rahi2007mapping}, conformal mapping \cite{hass2014round}, surface projections \cite{kontopoulos2016structuprint}, and map projections \cite{krone2017molecular}. Krone et al. \cite{krone2017molecular} presented a hole-filling technique to reduce the n genus surface to genus 0 and map the resultant genus 0 surface to a sphere for visualization. Unfortunately, the simplification to genus 0 surface was done at the expense of removing critical channel information and filling the resultant holes, which induces significantly high angle and area distortion; higher genus surfaces can induce considerably higher angle and area distortion due to higher loss in interior information. LMap, on the other hand, can be used for conformally deforming only the boundary cavities without removing the critical interior cavities and without affecting the rest of the molecular surface. 

The method closest to our LMap algorithm is spin transformations \cite{crane2011spin}. Spin transformations require global information, i.e. mean curvature computation for the entire mesh, to allow for conformal deformations for selected ROIs on this mesh; restricting the mean curvature computation to a local region while ignoring the rest of the surface leads to self-intersections (as shown in Figure~\ref{fig:spin_transformations}h). In contrast, LMap only requires local Gaussian curvature information to parameterize and conformally deform selected ROIs, and does not affect the rest of the surface, hence making it more useful for biomedical visualization tasks than the previous methods. Simple smoothing constrained to the boundary of the selected ROI does not preserve shape and hence, is not useful in the present context. LMap is inspired from the recent works in conformal parameterization domain (\cite{Springborn:2008,aigerman:2015}).

\section{Computational Algorithm}

This section introduces the computational algorithm for the intrinsic and extrinsic Ricci flow. The intrinsic Ricci flow \cite{nadeem2016spherical} is used to obtain the new lengths for the target Gaussian curvature and the extrinsic Ricci flow is used to compute the corresponding deformed mesh. More details on the intrinsic Ricci flow can be found in our previous paper \cite{nadeem2016spherical}.

\subsection{Intrinsic Surface Ricci Flow}
In practice, surfaces are represented as triangular meshes. A \emph{discrete metric} on a mesh is
the edge length function, denoted as $l: E \to \mathbb{R}^+$, which satisfies the triangle inequality. The \emph{discrete Gauss curvature} is the angle deficit, defined on vertices, $K: V \to \mathbb{R}$,
\begin{equation}
K(v)= \left\{
\begin{array}{rl}
2\pi - \sum_{jk} \theta_i^{jk}, & v \not\in \partial M\\
 \pi - \sum_{jk} \theta_i^{jk}, & v \in \partial M\\
\end{array}
\right.,
\label{eqn:discrete_curvature}
\end{equation}
where $\theta_i^{jk}$ is the corner angle at $v_i$ in the face $[v_i,v_j,v_k]$, and $\partial M$ represents the boundary of the mesh.

The discrete Gaussian curvature is determined by the discrete Riemannian metric via the cosine law,
\begin{equation}
    l_i^2 = l_j^2 + l_k^2 - 2l_jl_k \cos\theta_i.
    \label{eqn:cosine_law}
\end{equation}
According to the Gauss-Bonnet theorem, the total Gaussian curvature is a topological invariant and equals the product of $2\pi$ and the Euler characteristic number $\chi$,
\begin{equation}
    \sum_{v \not\in \partial \Sigma} K(v) + \sum_{v\in \partial \Sigma} K(v) = 2\pi \chi(\Sigma).
    \label{eqn:discrete_Gauss_Bonnet}
\end{equation}

The \emph{cotangent edge weight} plays an important role. Given an interior edge $[v_i,v_j]$  adjacent to two faces $[v_i,v_j,v_k]$ and $[v_j,v_i,v_l]$, the cotangent weight is defined as
\begin{equation}
    w_{ij} = \cot \theta_k^{ij} +\cot\theta_l^{ji}.
    \label{eqn:cotangent_weight_1}
\end{equation}
If the edge is on the boundary, adjacent to the face $[v_i,v_j,v_k]$, then the cotangent weight is
\begin{equation}
    w_{ij} = \cot \theta_k^{ij}.
    \label{eqn:cotangent_weight_2}
\end{equation}
A triangulation of the mesh is called \emph{Delaunay} if all cotangent edge weights are non-negative.

Given a triangular mesh $M$, the \emph{discrete conformal factor} is a function defined on each vertex $\mathbf{u}:V\to\mathbb{R}$,  the length of an edge $[v_i,v_j]$ is given by
\begin{equation}
    l_{ij} = e^{u_i}\beta_{ij}e^{u_j},
    \label{eqn:edge_length_for_euclidean_yamabe_flow}
\end{equation}
where $\beta_{ij}$ is the initial edge length.

\hspace{-5mm}\textbf{Discrete Surface Ricci Flow:}  The discrete surface Ricci flow is defined as
\begin{equation}
\frac{du_i(t)}{dt} = \bar{K}_i - K_i(t),
\label{eqn:discrete_ricci_flow}
\end{equation}
\noindent where $\bar{K}_i$ is the target curvature at the vertex $v_i$, and the discrete metric is given by Eqn. (\ref{eqn:edge_length_for_euclidean_yamabe_flow}). 

Furthermore, the discrete Ricci flow is the negative gradient flow
of the discrete Ricci energy:
\begin{equation}
E_\Sigma(\mathbf{u}) = \int^{\mathbf{u}} \sum_{i=1}^n (\bar{K}_i-K_i) du_i.
 \label{eqn:mesh_ricci_energy}
\end{equation}

The gradient of the Ricci energy is $(\bar{K}_i-K_i)^T$, the Hessian matrix consists of cotangent edge weights, given in Eqn.~(\ref{eqn:cotangent_weight_1}),
\begin{equation}
\frac{\partial K_i}{\partial u_j} = \frac{\partial K_j}{\partial
u_i} = w_{ij},
\label{eqn:Euclidean_Hessian_1}
\end{equation}
\noindent and the diagonal elements are
\begin{equation}
\frac{\partial K_i}{\partial u_i} = -\sum_j w_{ij}. \label{eqn:Euclidean_Hessian_2}
\end{equation}

\hspace{-5mm}\textbf{Dynamic Surface Ricci Flow \cite{nadeem2016spherical}:} Given a target curvature $\bar{K}:V\to\mathbb{R}$, satisfying the Gauss-Bonnet condition (\ref{eqn:discrete_Gauss_Bonnet}), conventional Ricci flow does not guarantee the existence of the solution. During the conventional discrete surface Ricci flow, some triangles may become degenerated; namely the triangle inequality does not hold on some faces, hence the flow terminates. In order to overcome the instability of conventional surface Ricci flow, we use dynamic surface Ricci flow \cite{nadeem2016spherical}: during the Ricci flow, the triangulation of the mesh is updated to be Delaunay with respect to the current Riemannian metric.

The Delaunay triangulation can be obtained by a finite number of edge flipping operations: for each edge, we flatten the two adjacent faces to form a planar quadrilateral, then swap the edge to the shorter diagonal. During the optimization, the triangulation is updated to be Delaunay at each step; details can be found in Algorithm~\ref{alg:yamabe}. Preserving Delaunay triangulation is crucial to ensure the existence of the solution to the dynamic Ricci flow. Given a target curvature $\bar{K}$ satisfying the Gauss-Bonnet condition in Eqn. (\ref{eqn:discrete_Gauss_Bonnet}), and for each vertex $\bar{K}_i\in (-\infty,2\pi)$, there exists a solution to the dynamic Ricci flow \cite{gu2016}. The solution is unique up to a constant.

\begin{algorithm}
\caption{Intrinsic Surface Ricci Flow}
\label{alg:yamabe}
\begin{algorithmic}
\REQUIRE{The input mesh $M$, target curvature $\bar{K}$, threshold $\epsilon$}
\ENSURE{The edge length which realizes the target curvature}

\STATE Compute the initial edge lengths $\{\beta_{ij}\}$; \\
\STATE Initialize the conformal factor to be zeros;\\
\WHILE {true}
\STATE Compute the edge lengths using Eqn. (\ref{eqn:edge_length_for_euclidean_yamabe_flow});
\STATE Update the triangulation to be Delaunay by edge flipping;
\STATE Compute the corner angles using Eqn. (\ref{eqn:cosine_law});
\STATE Compute the cotangent edge weights using Eqns.~(\ref{eqn:cotangent_weight_1}) and (\ref{eqn:cotangent_weight_2});
\STATE Compute the vertex curvature using Eqn. (\ref{eqn:discrete_curvature});
\IF {$\forall |\bar{K}_i-K_i(\mathbf{h})| < \epsilon$}
\STATE    Break;
\ENDIF
\STATE Compute the gradient of the Ricci flow using Eqn. (\ref{eqn:discrete_ricci_flow});
\STATE Compute the Hessian of the Ricci energy using Eqns. (\ref{eqn:Euclidean_Hessian_1}) and (\ref{eqn:Euclidean_Hessian_2});
\STATE Solve the linear system
\[
    Hess(\mathbf{u}) \delta \mathbf{u} = \nabla E(\mathbf{u})
\]
$\mathbf{u} \leftarrow \mathbf{u} + \delta\mathbf{u}$;
\ENDWHILE
\RETURN the edge length $\{l_{ij}\}$
\end{algorithmic}
\end{algorithm}

\begin{figure*}[t!]
\begin{center}
\begin{tabular}{ccccc}
\includegraphics[width=0.02\textwidth]{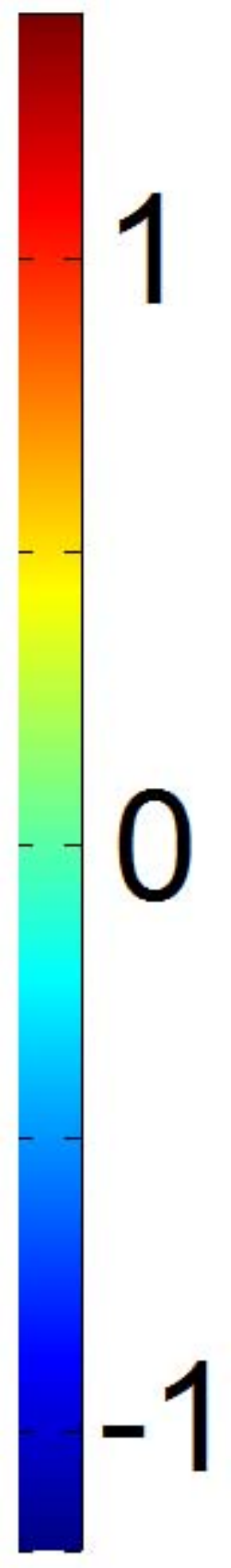}&
\includegraphics[width=0.22\textwidth]{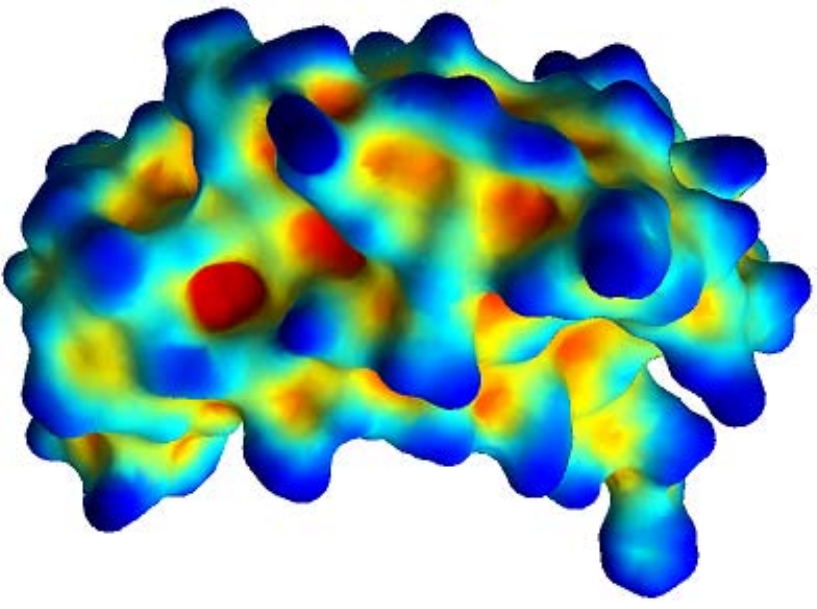}&
\includegraphics[width=0.22\textwidth]{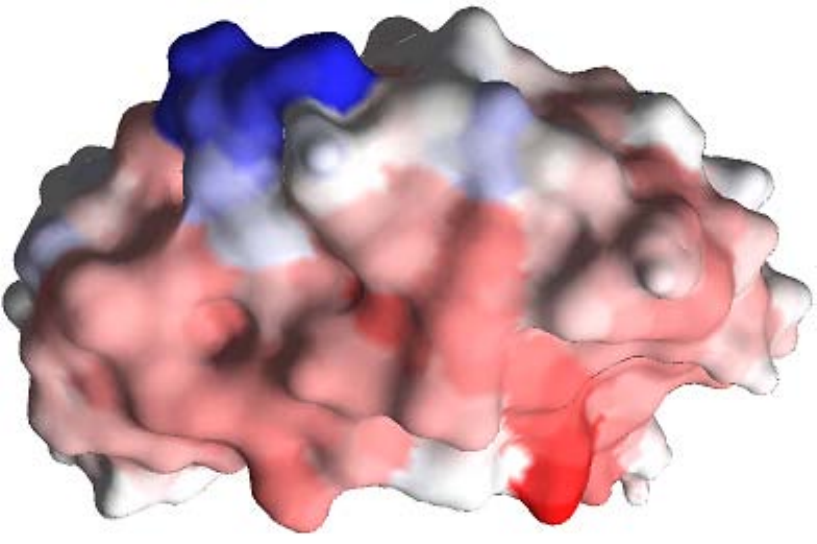}&
\includegraphics[width=0.22\textwidth]{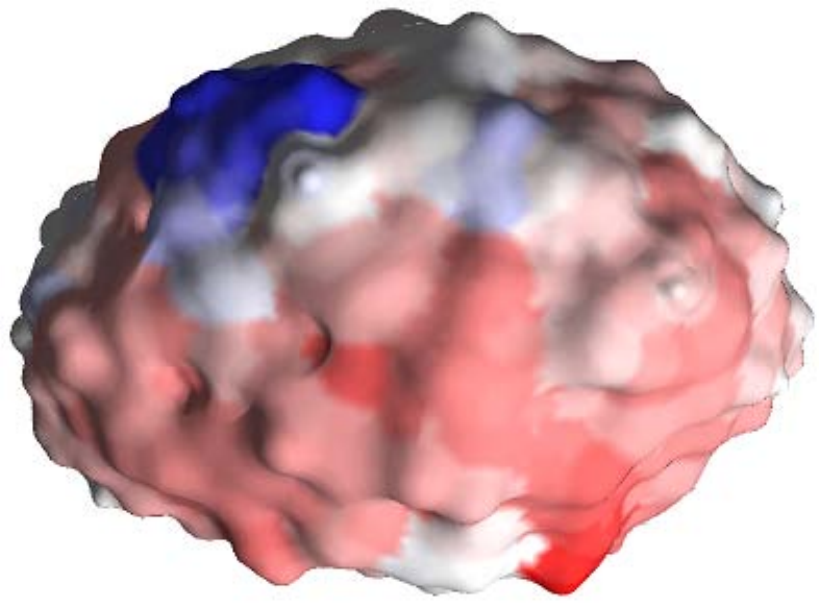}&
\includegraphics[width=0.22\textwidth]{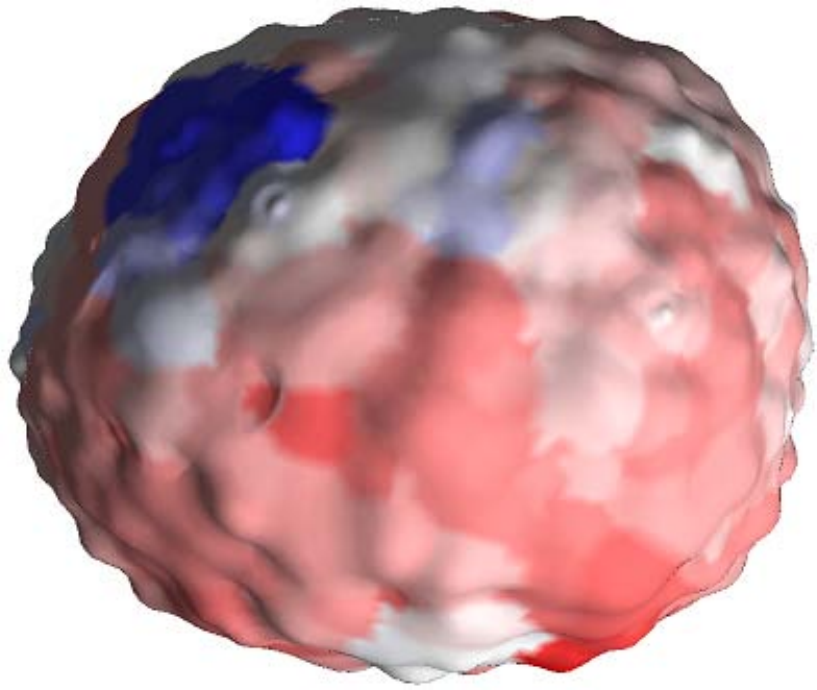}\\
& (a) & (b) & (c) & (d)\\
\includegraphics[width=0.02\textwidth]{figures/spin_transformations/colormap_curv}&
\includegraphics[width=0.22\textwidth]{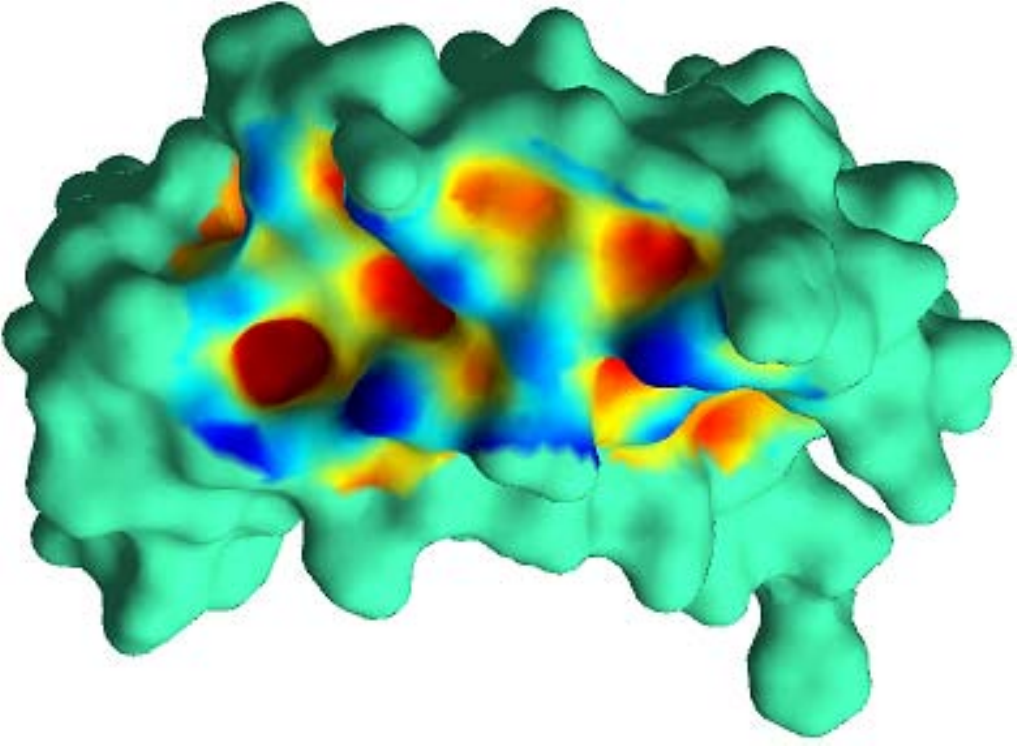}&
\includegraphics[width=0.22\textwidth]{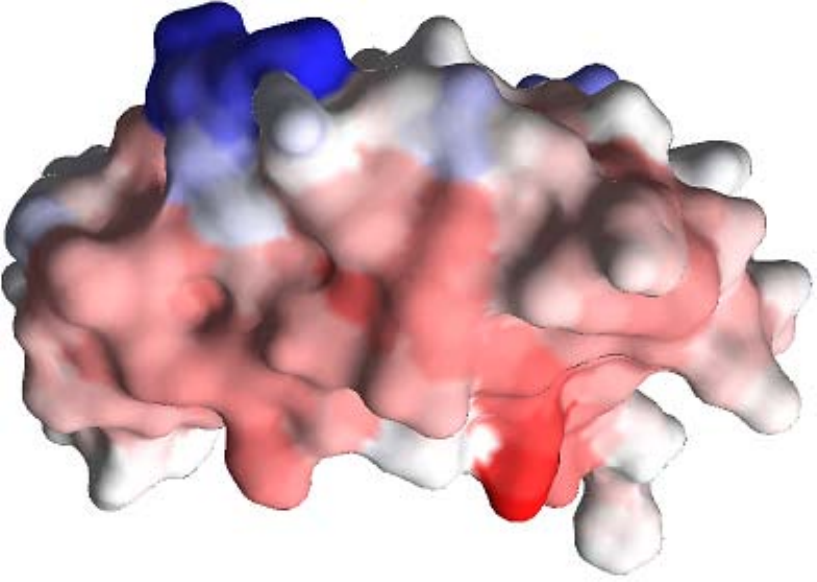}&
\includegraphics[width=0.22\textwidth]{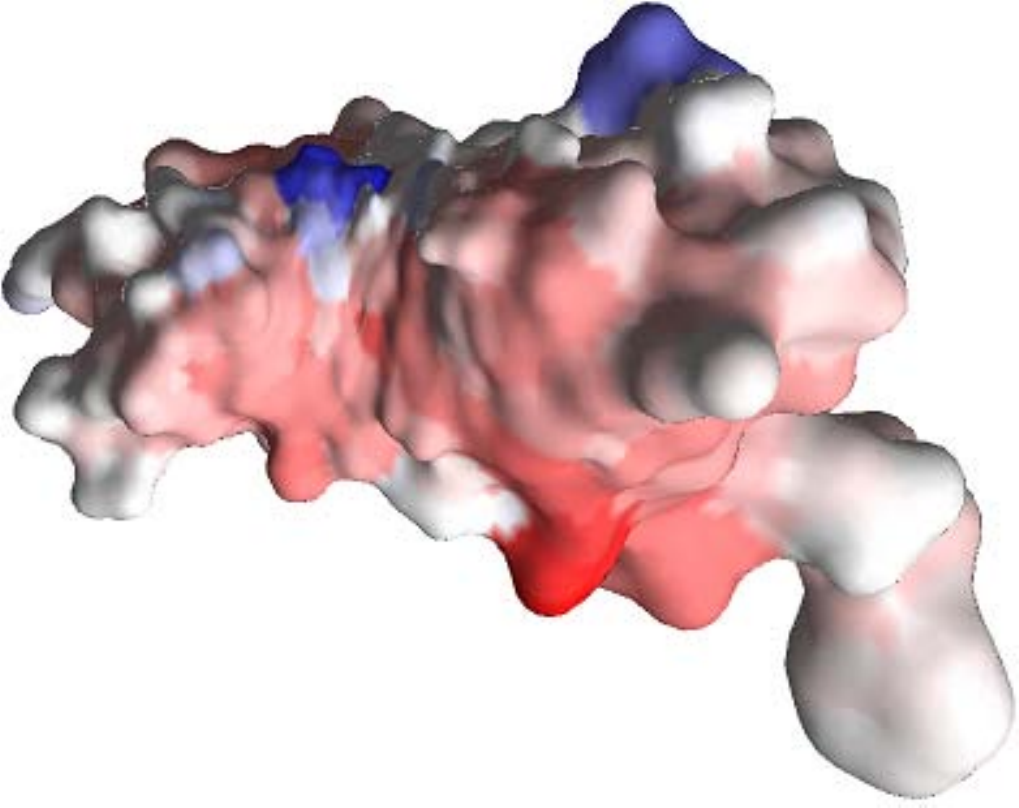}&
\includegraphics[width=0.22\textwidth]{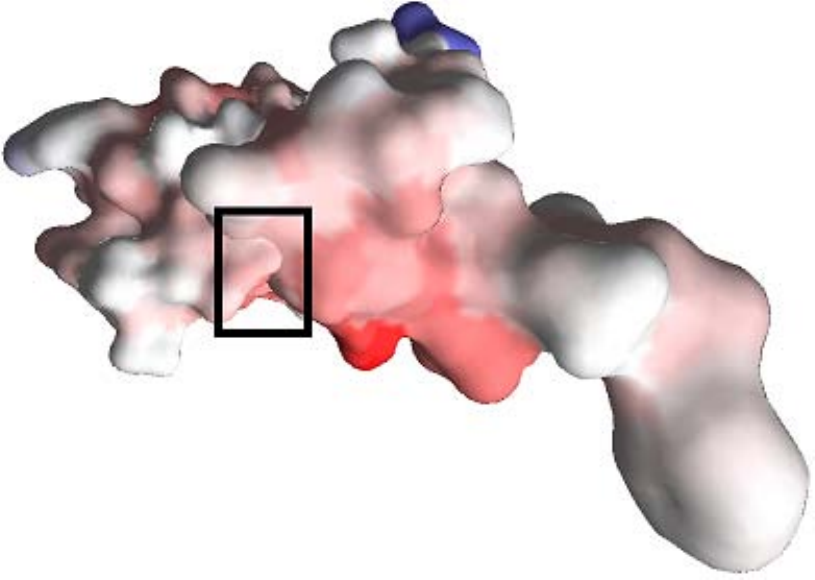}\\
& (e) & (f) & (g) & (h)\\
&
\includegraphics[width=0.22\textwidth]{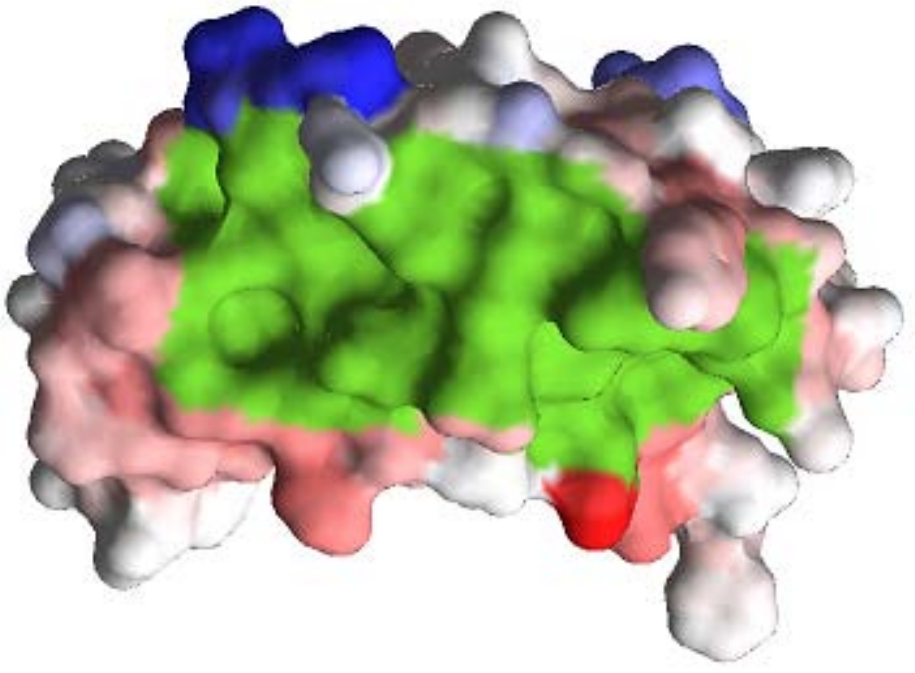}&
\includegraphics[width=0.22\textwidth]{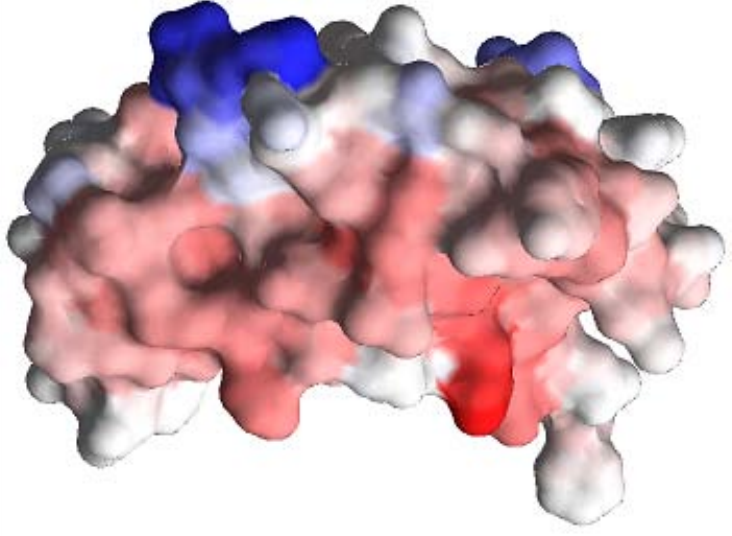}&
\includegraphics[width=0.22\textwidth]{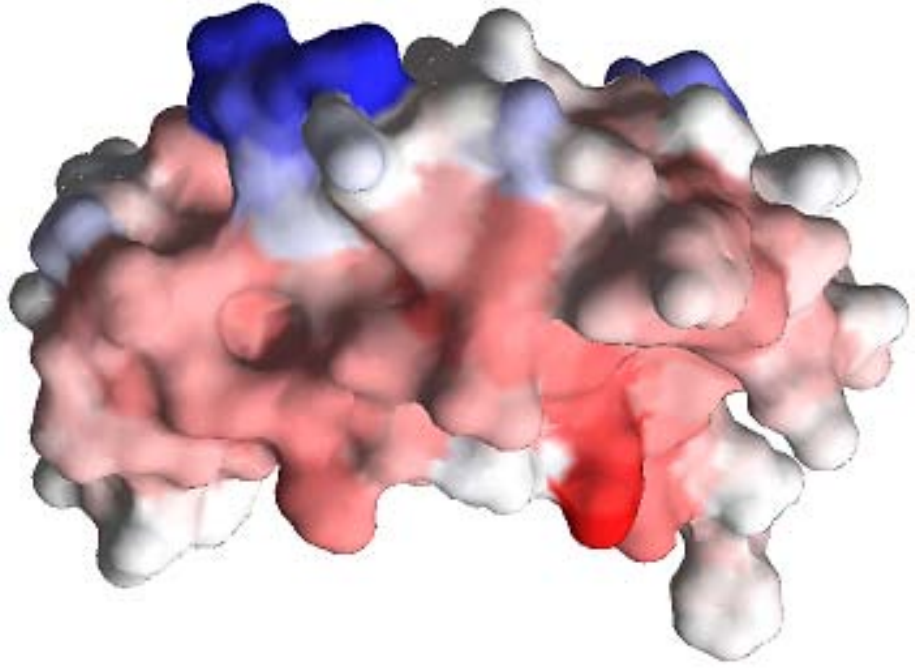}&
\includegraphics[width=0.22\textwidth]{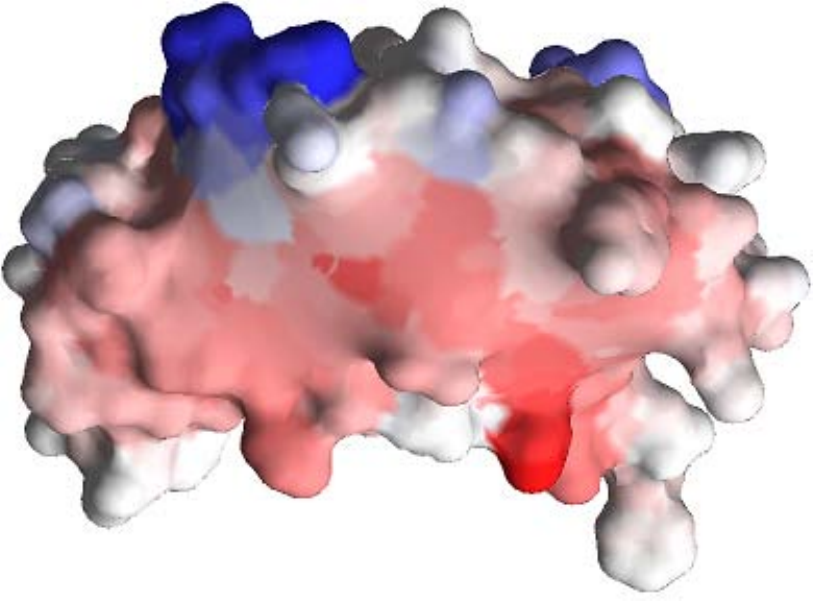}\\
& (i) & (j) & (k) & (l)
\end{tabular}
\end{center}
\caption{Spin transformations comparison for the molecular surface in Figure~\ref{fig:molecular_surface}a. (a) Mean curvature and \emph{global} spin transformations with varying scale factors (b) 2, (c) 6, and (d) 10. (e) Mean curvature is set to zero everywhere except the selected ROI and corresponding \emph{local} spin transformations with varying scale factors (f) 2, (g) 6, and (h) 10. (h) Surface self-intersects (black bounding box). (i) Selected ROI and the corresponding LMap deformations (using Gaussian curvature) after (j) 1, (k) 2, and (l) 5 steps. 
\label{fig:spin_transformations}}
\end{figure*}

\subsection{Extrinsic Surface Ricci Flow}

Given a triangle mesh $M=(V,E,F)$, we select a region of interest (ROI) $\Omega \subset M$. We choose the step length as $1/n$, where $n$ is a positive integer. At the $k$-th step, we set the target curvatures for vertices inside $\Omega$,
\begin{equation}
    \bar{K}(v_i) = \left(1-\frac{k}{n}\right) K(v_i), ~\forall v_i \in \Omega,
    \label{eqn:target_curvature}
\end{equation}
where $K(v_i)$ is the initial discrete Gaussian curvature. We use the intrinsic discrete surface Ricci flow algorithm (Algorithm~\ref{alg:yamabe}) to compute the corresponding edge length, denoted as $l^k: E\to\mathbb{R}_{>0}$. Then we adjust the positions of the vertices to satisfy the edge length constraints.

At the $k$-th step, the mesh is $M^k$, the position for the vertex $v_i$ is $p_i^k$, and the deformation is along the normal field of the surface. Let $\lambda^k: V\to \mathbb{R}$ be a function defined on the vertex set, the deformation of $v_i$ is represented as
\begin{equation}
    p_i^{k+1} \leftarrow p_i^k + \lambda_i^k n_i^k,
    \label{eqn:deformation}
\end{equation}
where $n_i^k$ is the normal to the mesh $M^k$ at the vertex $v_i$. The area of the face $[v_i,v_j,v_k]$ is given by
\[
    s_{ijl}^k = \frac{1}{2}|(p_j^k-p_i^k)\times(p_l^k-p_i^k)|.
\]
The normal to the face can be computed as
\[
    n_{ijl}^k  =  \frac{1}{2s_{ijl}^k} (p_j^k-p_i^k)\times(p_l^k-p_i^k).
\]
The normal to the vertex $v_i$ is given by
\begin{equation}
    d_i^k = \sum_{jl} s_{ijl}^k n_{ijl}^k,~n_i^k = \frac{d_i^k}{|d_i^k|}.
\end{equation}
The edge length of an edge $[v_i,v_j]$ is $l^k_{ij}$, the square of the length is $L^k_{ij}$, then the energy for the optimization is given by
\begin{equation}
    E(\lambda^k) := \sum_{[v_i,v_j]\in E}
    \left[\langle p_i^{k+1} - p_j^{k+1}, p_i^{k+1} - p_j^{k+1}\rangle - L^k_{ij}\right]^2.
    \label{eqn:energy}
\end{equation}
By direct computation, the gradient of the energy can be calculated as
\begin{equation}
\frac{\partial E(\lambda^k)}{\partial \lambda^k_i } = \sum_{v_i\sim v_j} (L_{ij}^{k+1}-L_{ij}^k)\langle n_i^k, p_i^k - p_j^k \rangle
\label{eqn:gradient}
\end{equation}
The algorithmic details can be found in Algorithm~\ref{alg:extrinsic}. The vertices outside the selected ROI are fixed in place and hence, are not affected by the local mapping of the given ROI.

\begin{algorithm}
\caption{Extrinsic Surface Ricci Flow}
\label{alg:extrinsic}
\begin{algorithmic}
\REQUIRE{The input mesh $M$ and the ROI $\Omega\subset M$}
\ENSURE{The deformed mesh $M$ such that $\Omega$ is flat}
\STATE $k\leftarrow 0$
\WHILE {$k<n$}
\STATE Set the target curvature for vertex in $\Omega$ using Eqn.~(\ref{eqn:target_curvature})
\STATE Use Alg. \ref{alg:yamabe} to compute the edge lengths $l^k$
\STATE Minimize the energy in Eqn.~(\ref{eqn:energy}) using gradient descent method, using the formula Eqn.~(\ref{eqn:gradient})
\STATE Deform the position of each vertex in $\Omega$ using Eqn.~(\ref{eqn:deformation})
$k \leftarrow k + 1$;
\ENDWHILE
\RETURN the deformed mesh
\end{algorithmic}
\end{algorithm}

\begin{figure}[ht!]
\vspace{3mm}
\begin{center}
\begin{tabular}{cc}
\includegraphics[width=0.22\textwidth]{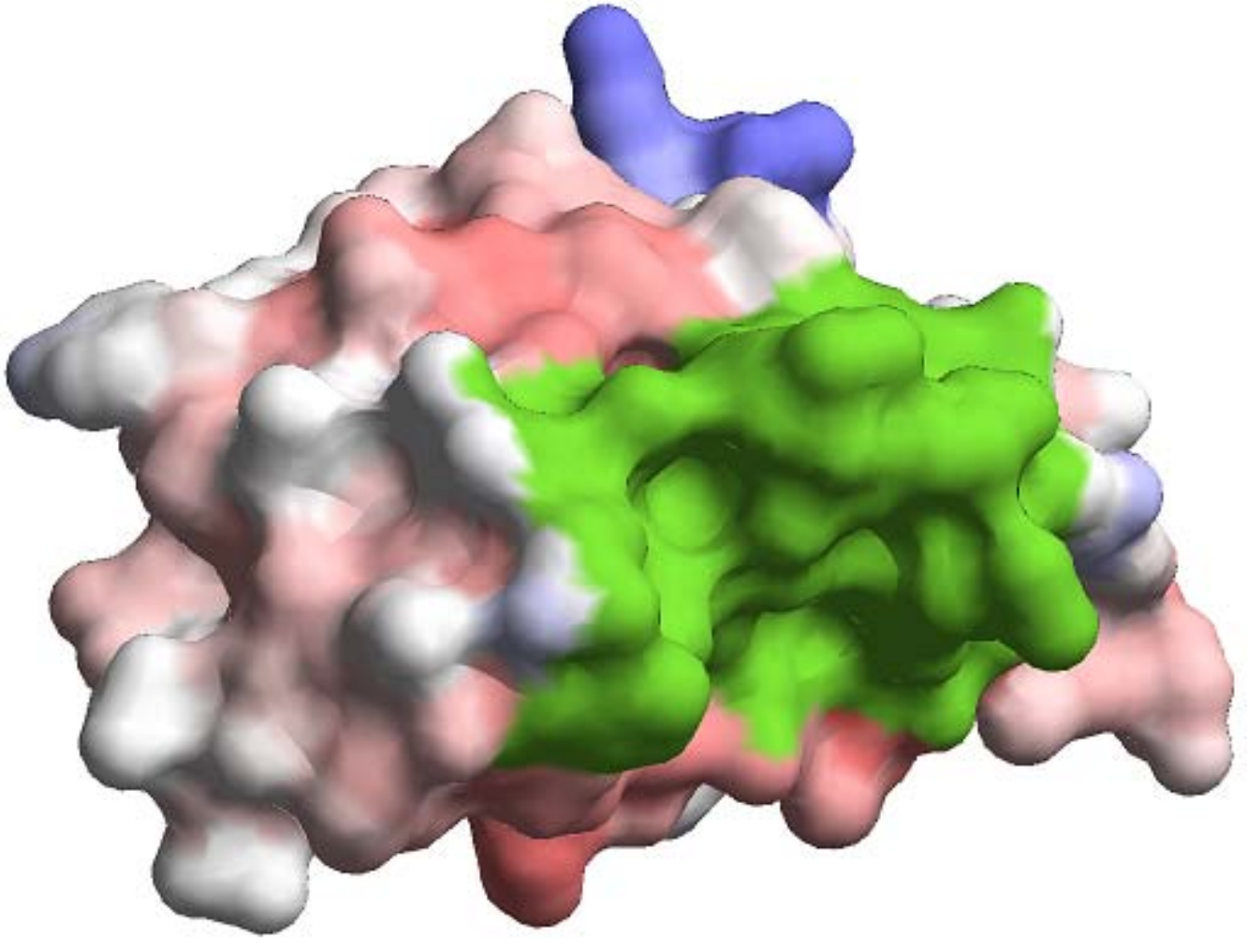}&
\includegraphics[width=0.22\textwidth]{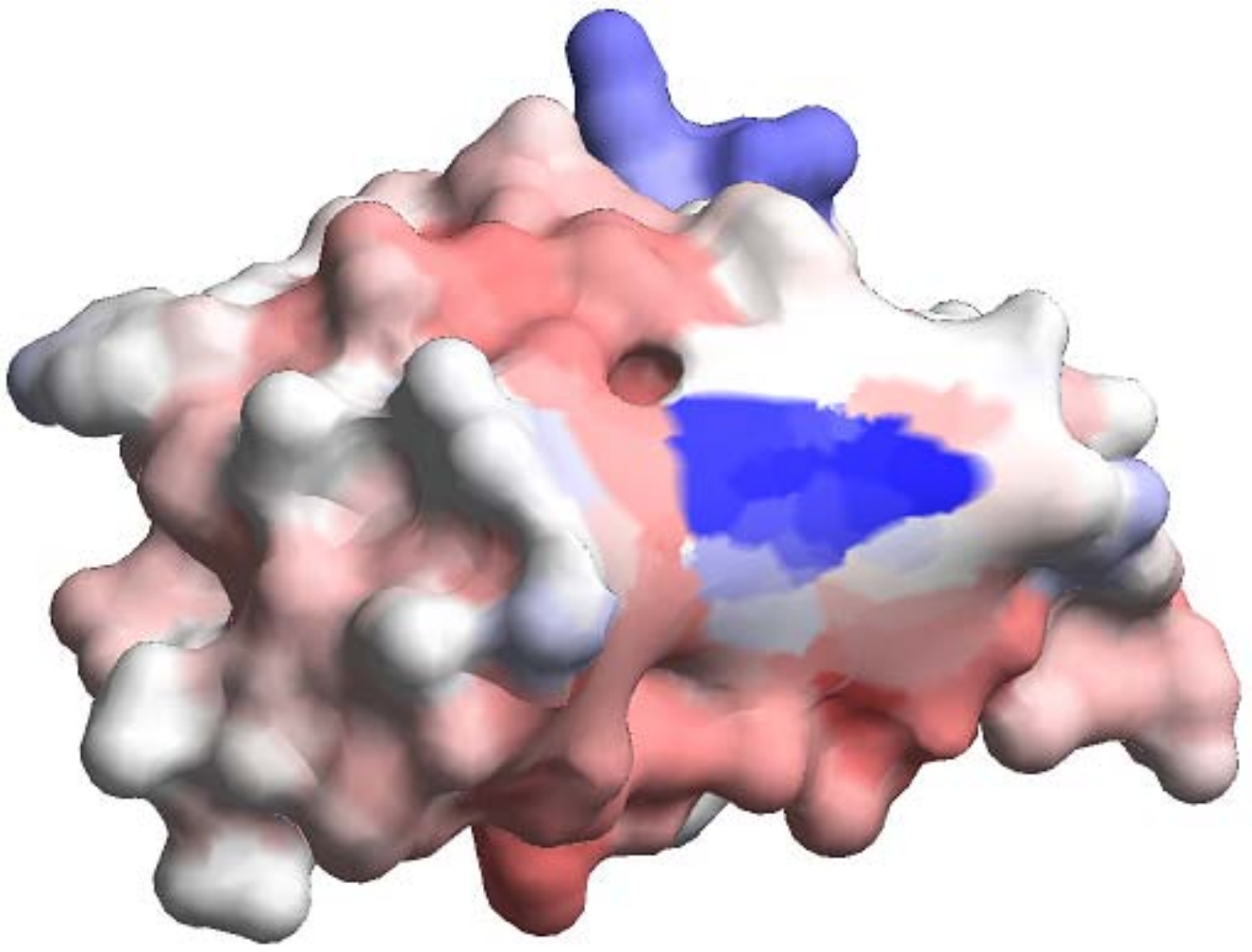}\\
(a) & (b)\\
\includegraphics[width=0.22\textwidth]{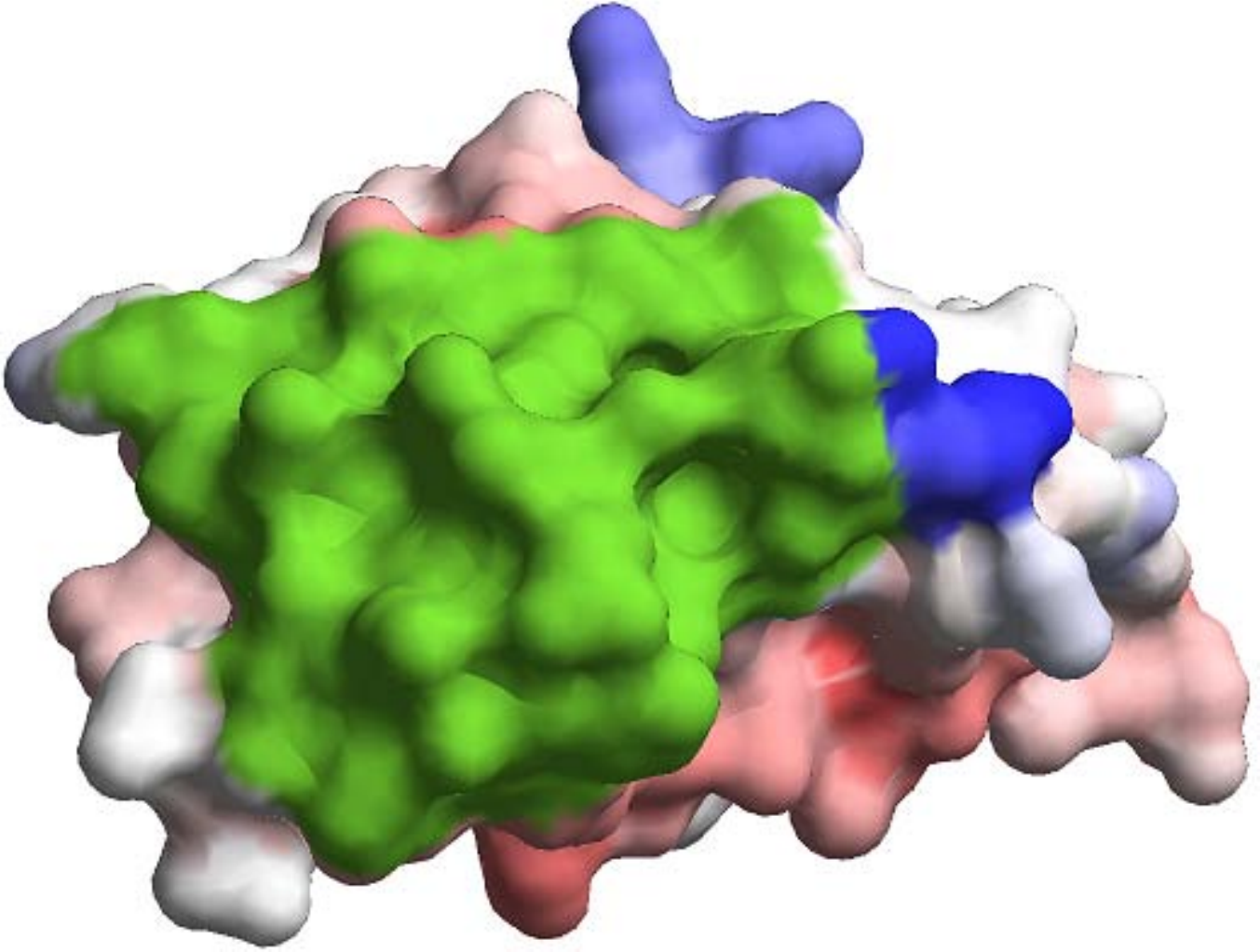}&
\includegraphics[width=0.22\textwidth]{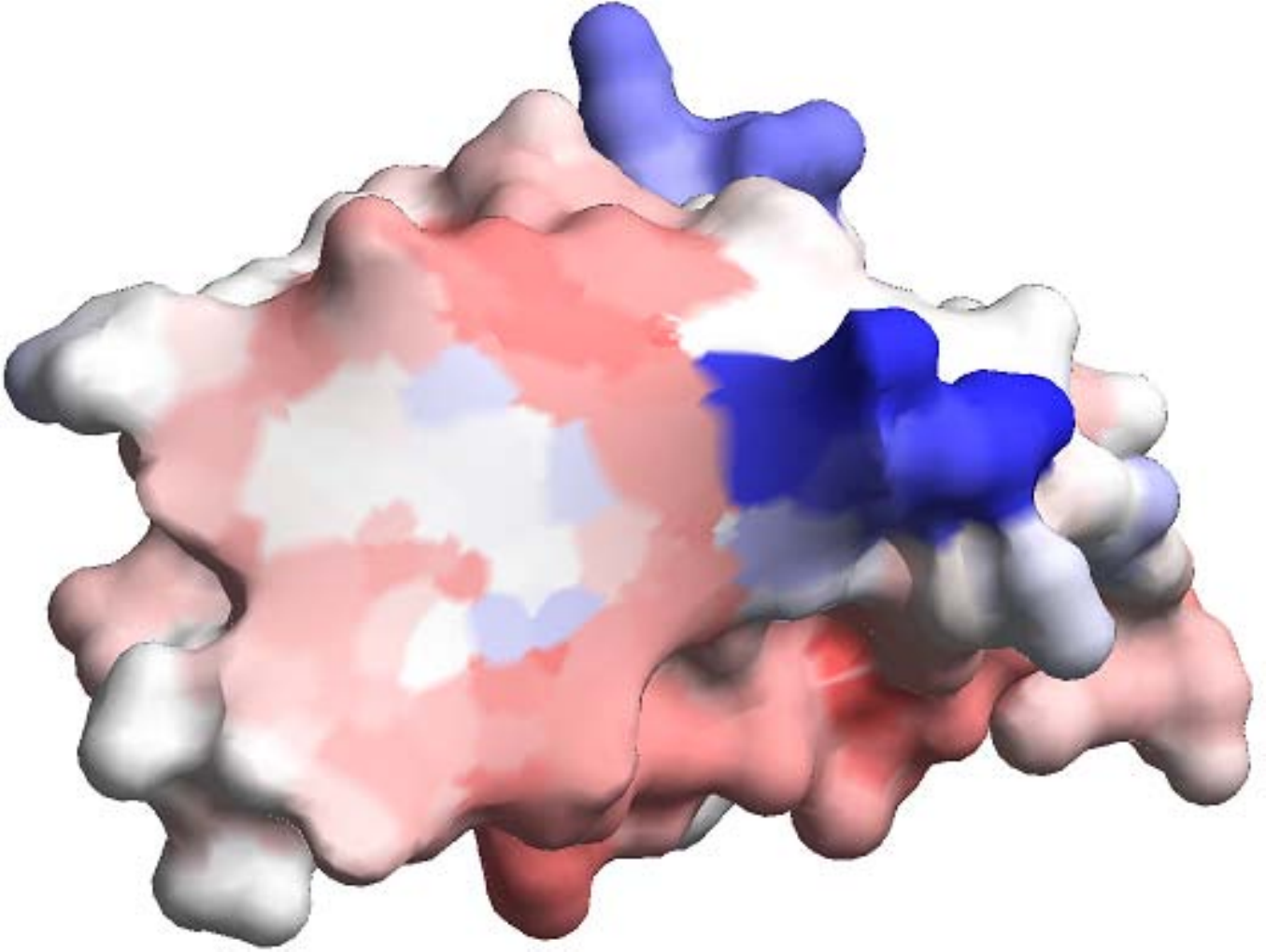}\\
(c) & (d)
\end{tabular}
\end{center}
\caption{Stability of LMap with ROI perturbations. Selected ROIs, (a) and (c), and their corresponding local mappings, (b) and (d).  
\label{fig:lmap_stability}}
\end{figure}

\begin{figure}[ht!]
\vspace{3mm}
\begin{center}
\begin{tabular}{cc}
\includegraphics[width=0.23\textwidth]{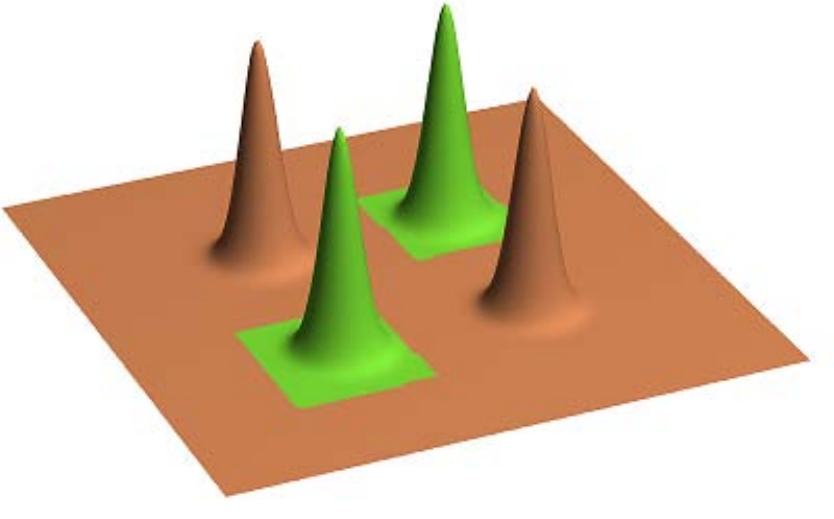}&
\includegraphics[width=0.23\textwidth]{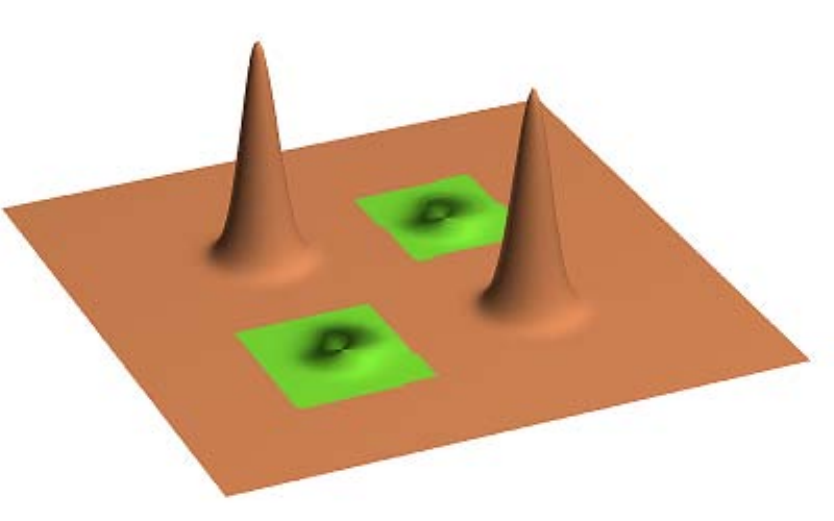}\\
(a) & (b)\\
\end{tabular}
\end{center}
\caption{Sharp Bends. (a) Selected ROIs and the (b) corresponding local mappings with normal mapped shading from (a).  
\label{fig:lmap_sharp_bends}}
\end{figure}

\begin{figure}[ht!]
\begin{center}
\begin{tabular}{cc}
\includegraphics[width=0.225\textwidth]{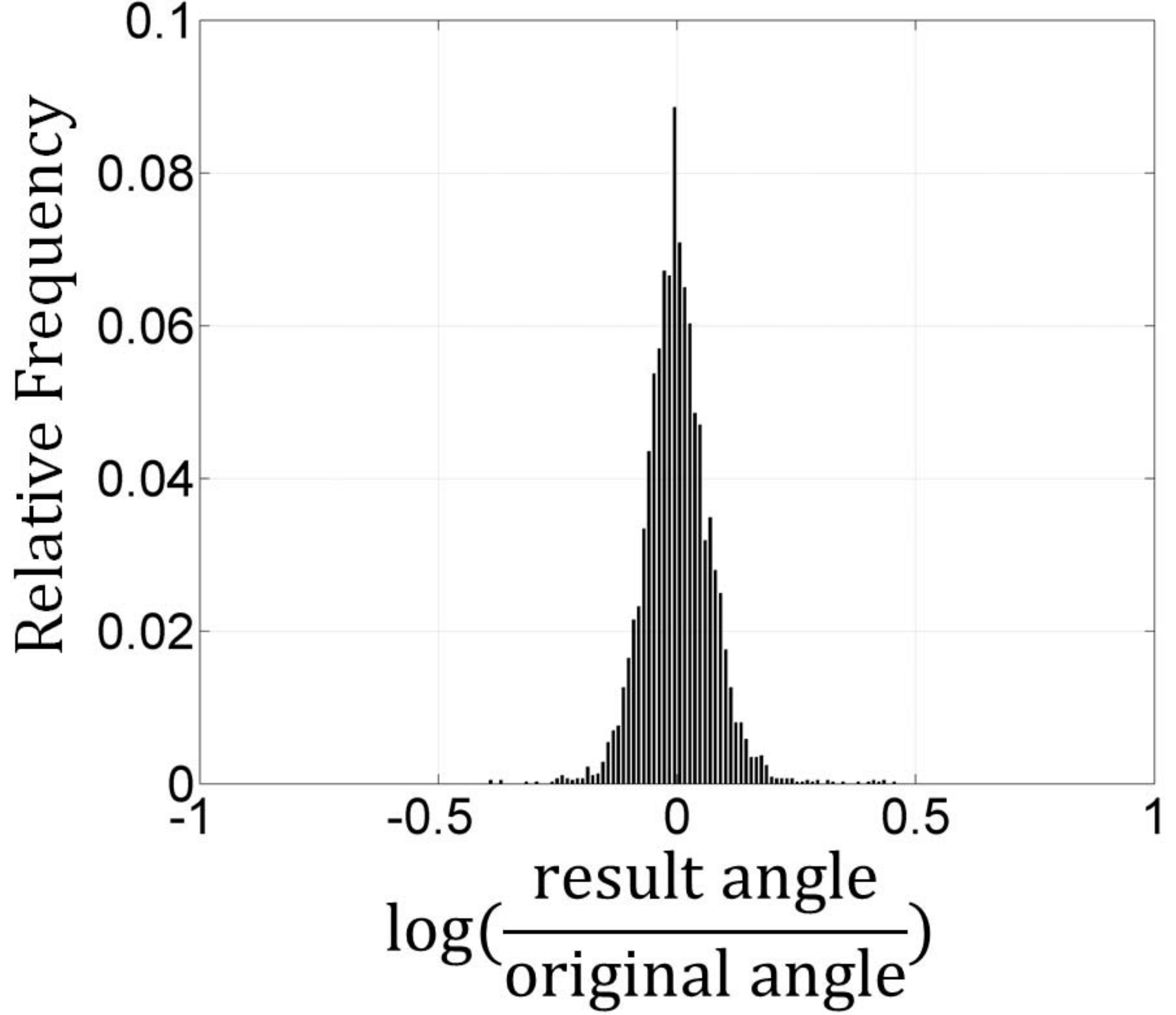}&
\includegraphics[width=0.225\textwidth]{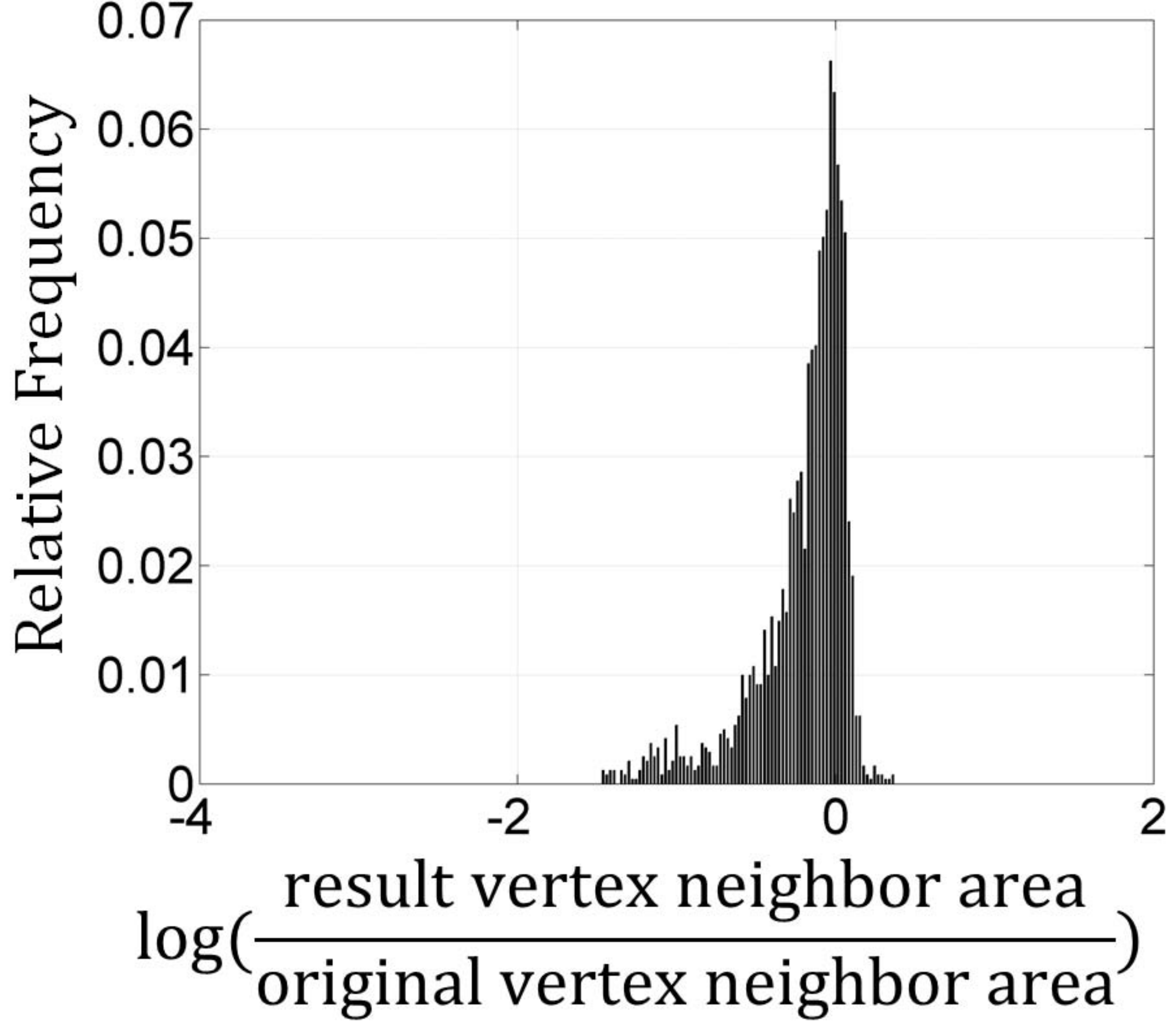}\\
(a) Angle Distortion LMap & (b) Area Distortion LMap\\
& \\
\includegraphics[width=0.225\textwidth]{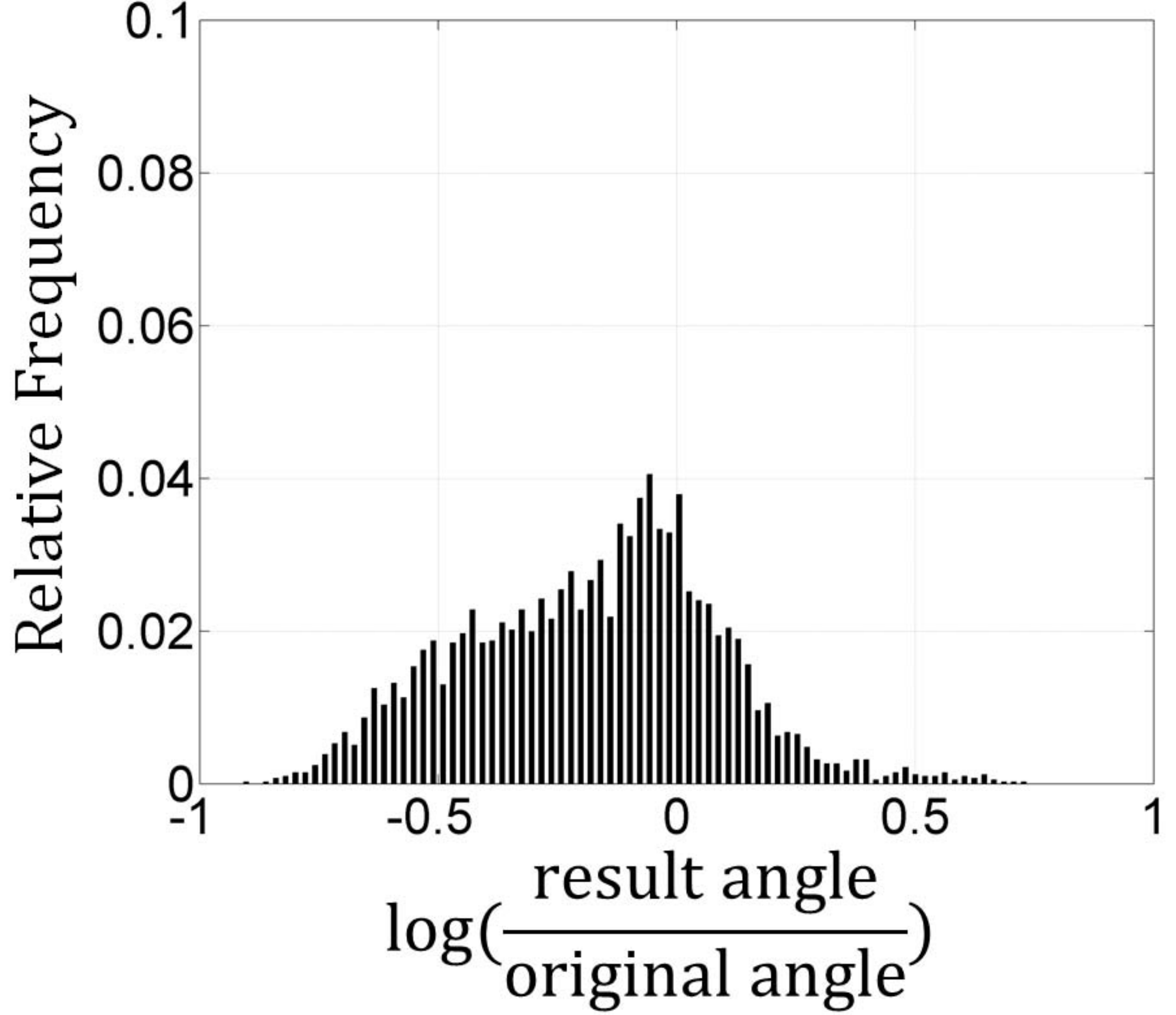}&
\includegraphics[width=0.225\textwidth]{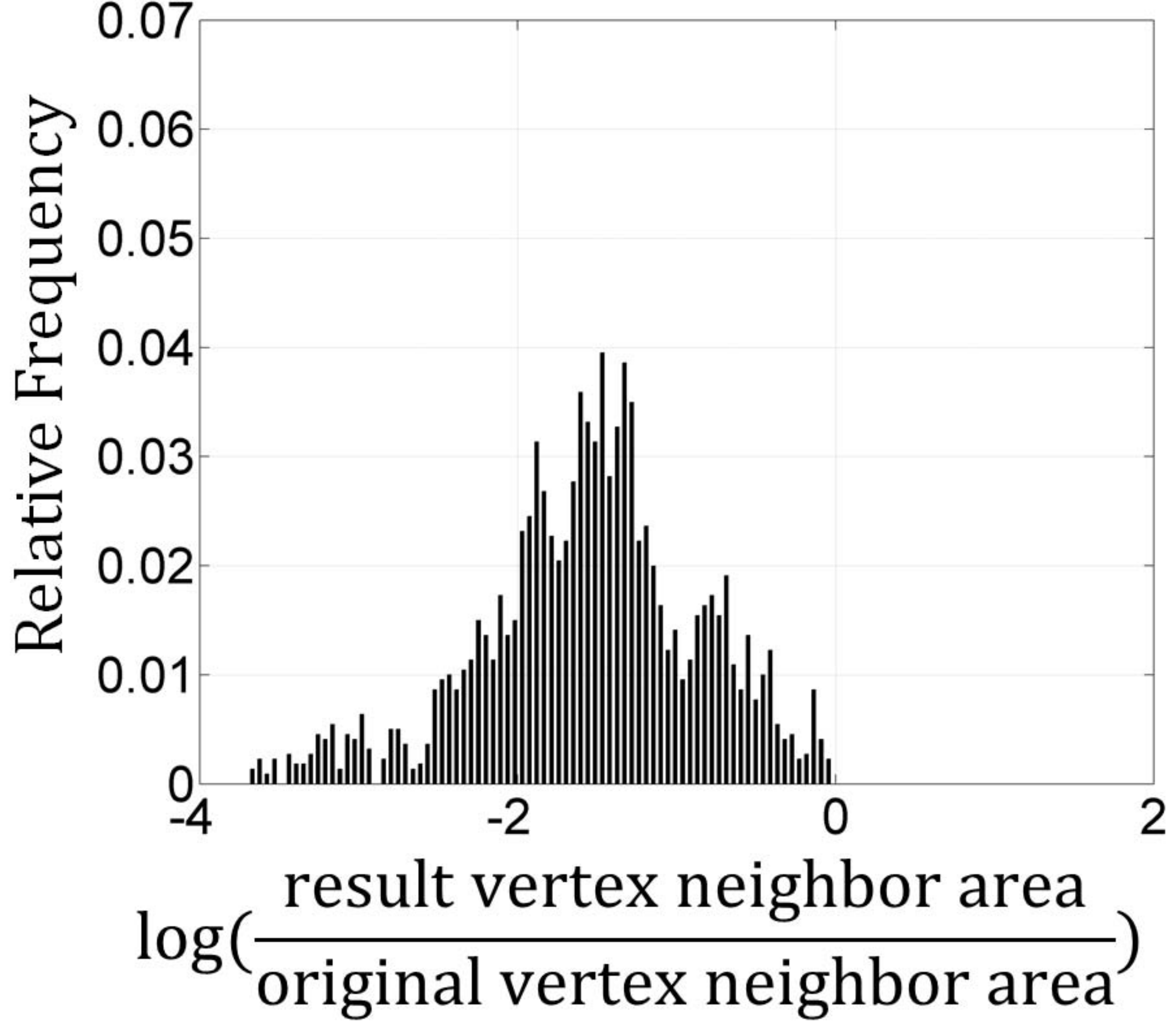}\\
(c) Angle Distortion Local & (d) Area Distortion Local\\
Spin Transformation & Spin Transformation \\
& \\
\includegraphics[width=0.225\textwidth]{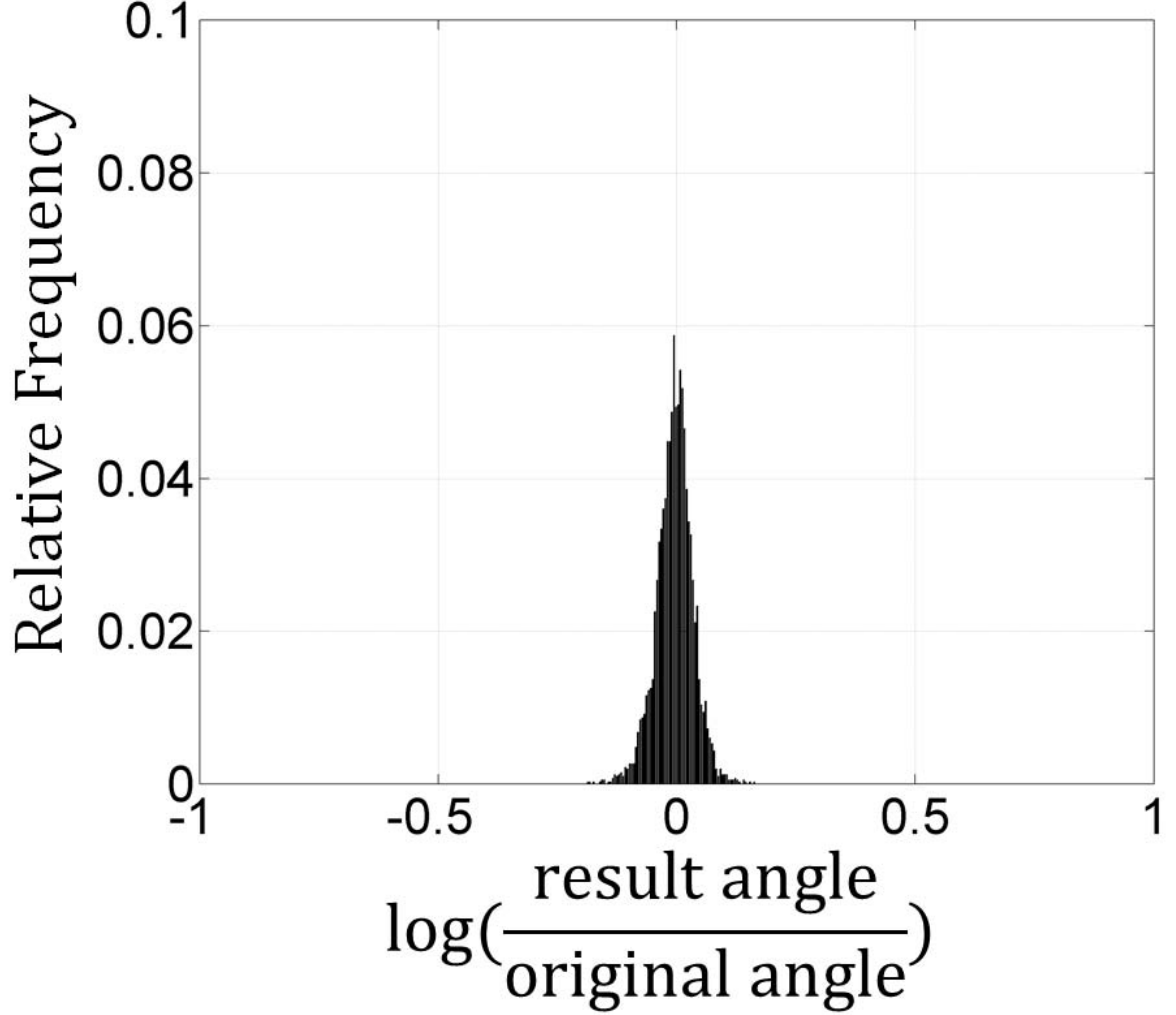}&
\includegraphics[width=0.225\textwidth]{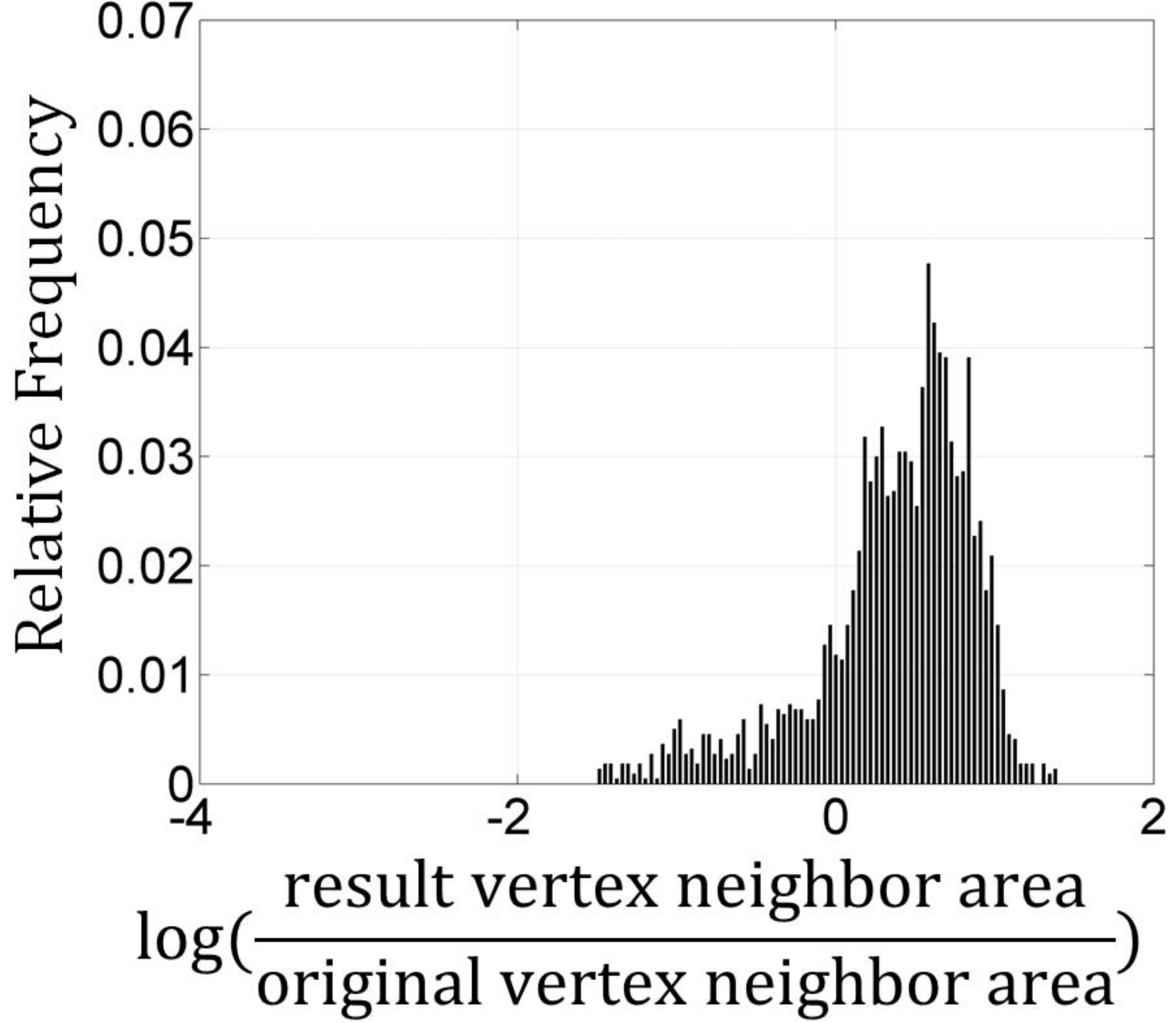}\\
(e) Angle Distortion Global & (f) Area Distortion Global\\
Spin Transformation & Spin Transformation\\
\end{tabular}
\end{center}
\caption{Area and angle distortion histograms. (a) Angle and (b) area distortion histograms for LMap on the selected ROI in Figure~\ref{fig:spin_transformations}l. (c) Angle and (d) area distortion histograms for local spin transformation on the same ROI with scale factor 6, as shown in Figure~\ref{fig:spin_transformations}g. (e) Angle and (f) area distortion histograms for global spin transformation on the same ROI with scale factor 6, as shown in Figure~\ref{fig:spin_transformations}c.
\label{fig:histograms}}
\end{figure}

\begin{figure*}[t!]
\begin{center}
\begin{tabular}{cccc}
\includegraphics[width=0.22\textwidth]{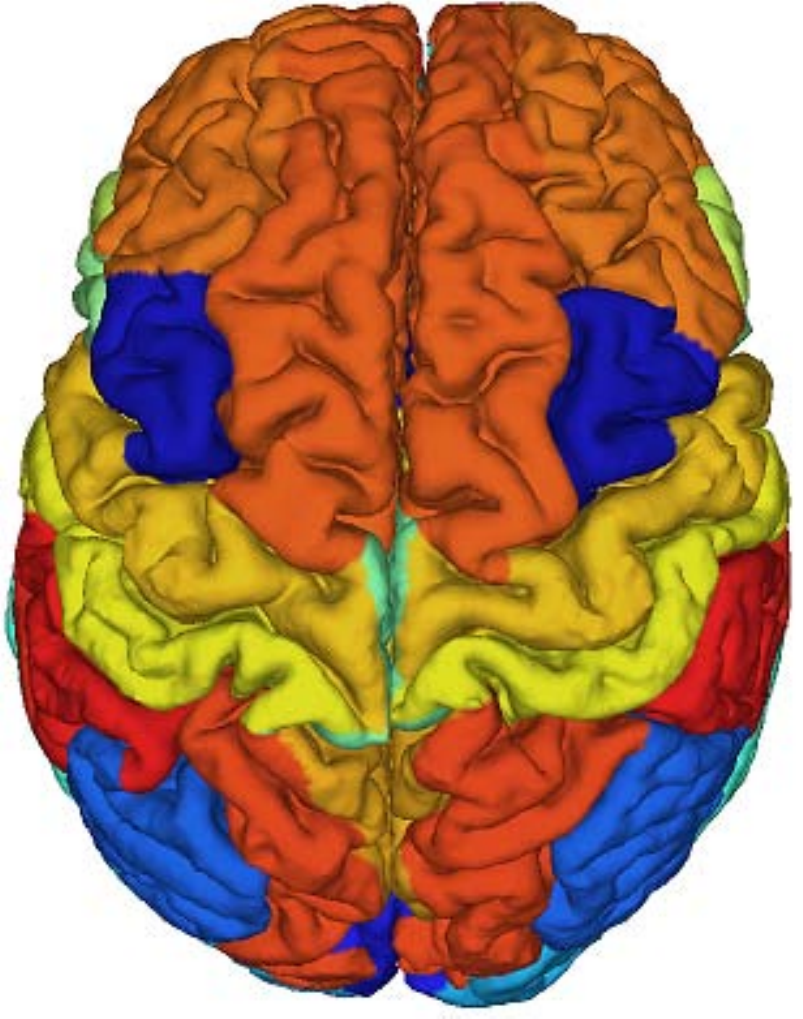}&
\includegraphics[width=0.22\textwidth]{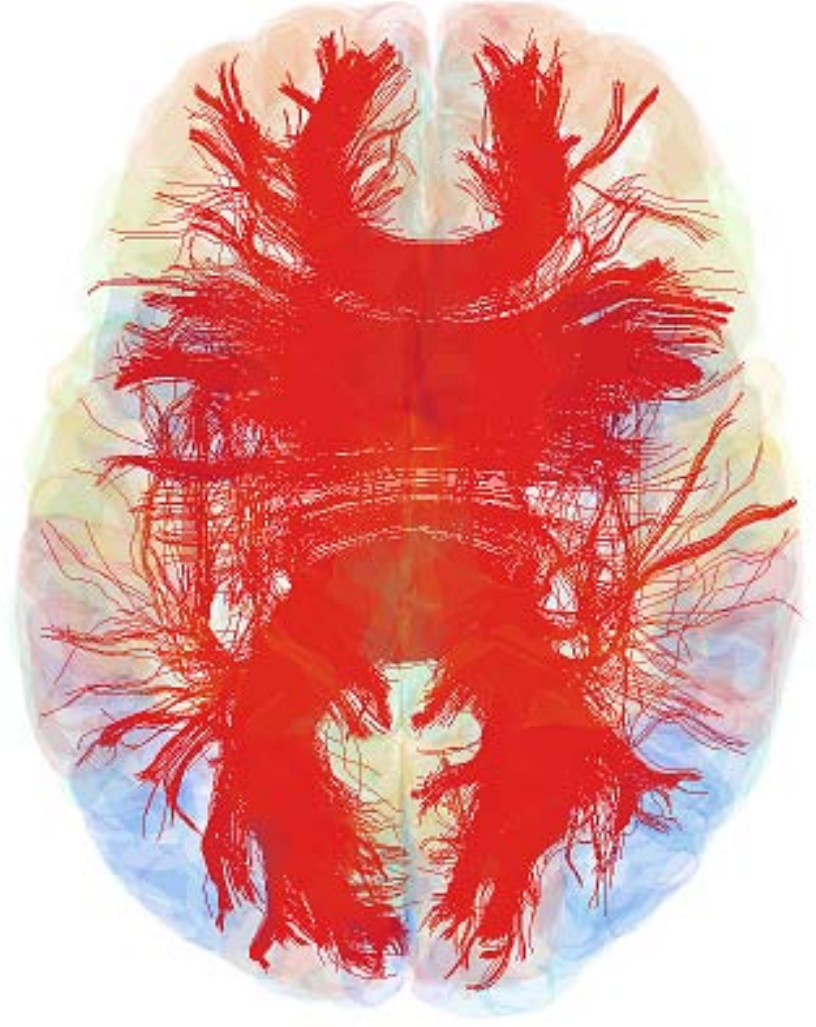}&
\includegraphics[width=0.22\textwidth]{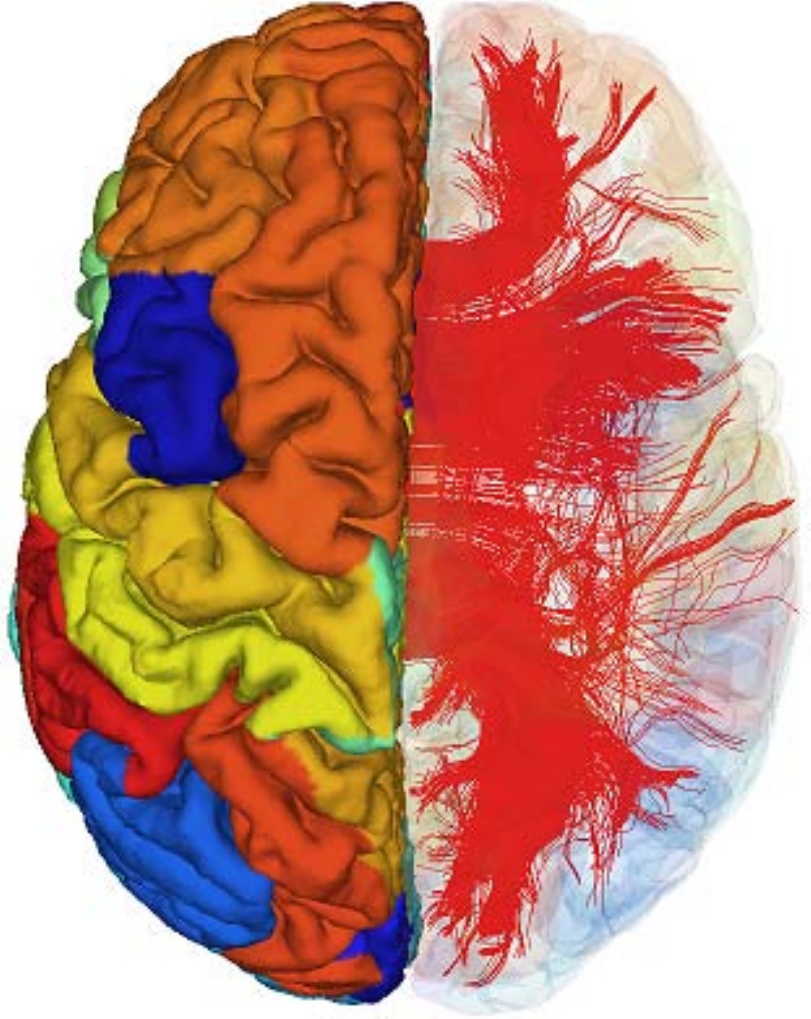}&
\includegraphics[width=0.22\textwidth]{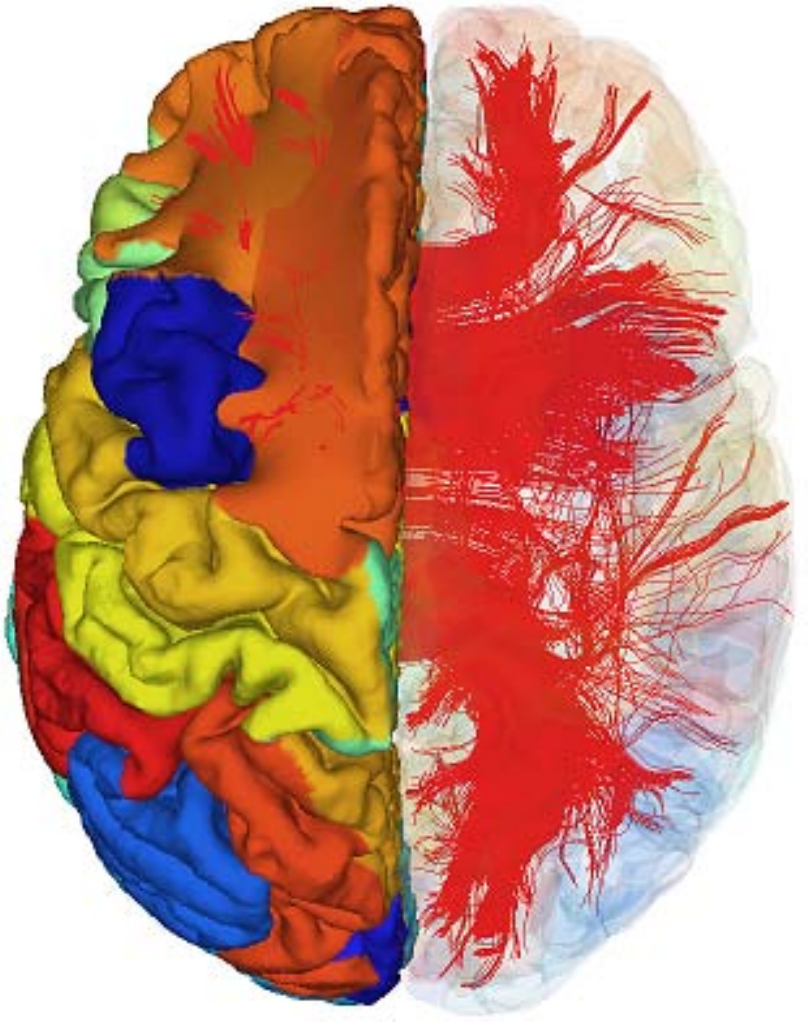}\\
(a) & (b) & (c) & (d)\\
\includegraphics[width=0.22\textwidth]{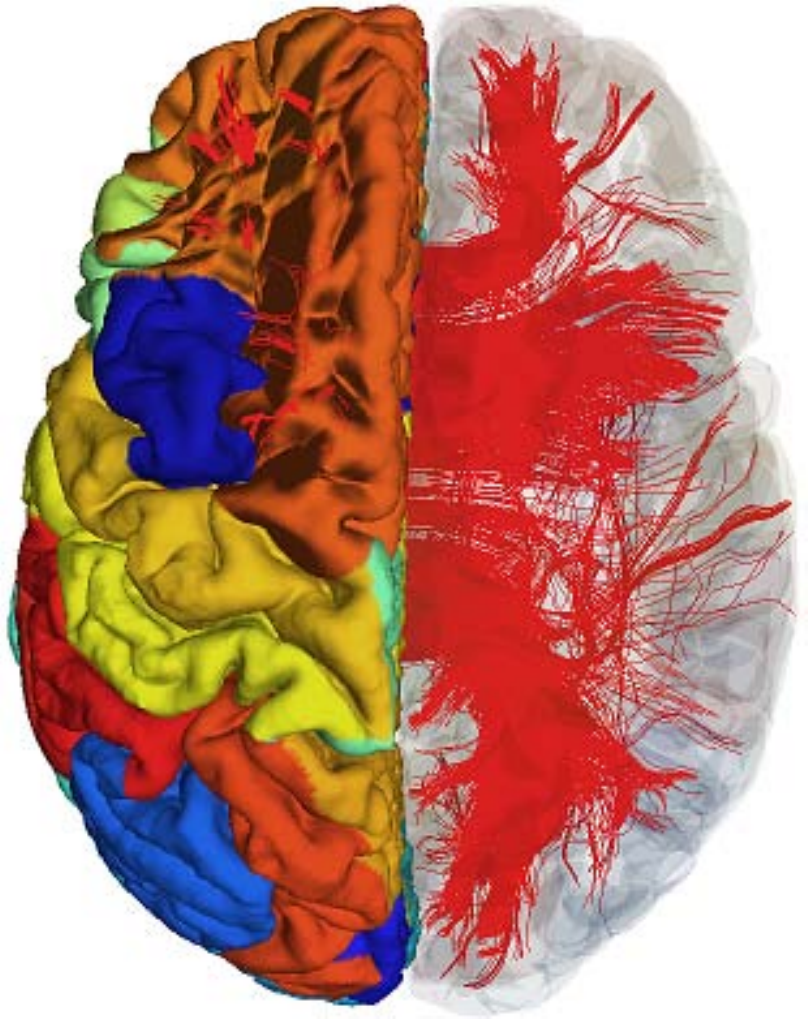}&
\includegraphics[width=0.22\textwidth]{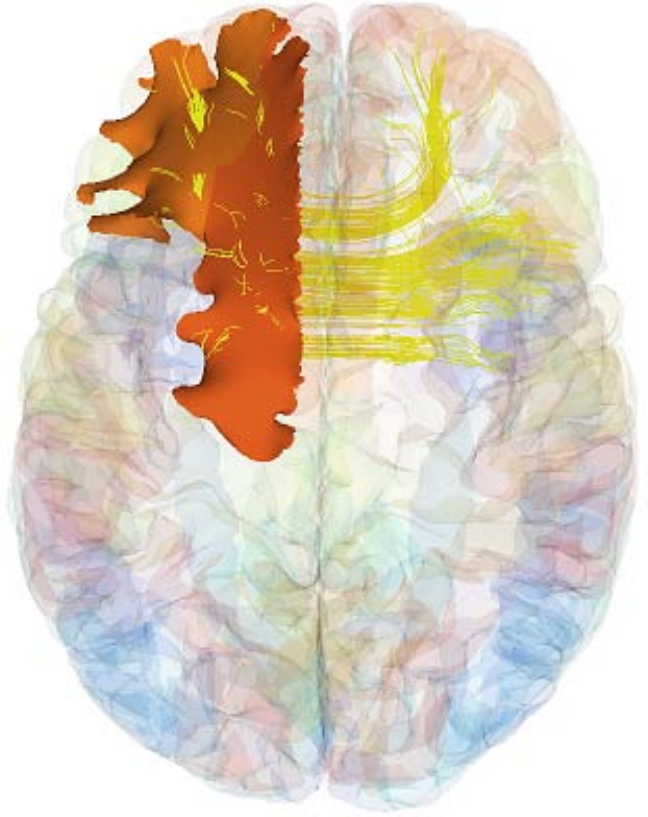}&
\includegraphics[width=0.22\textwidth]{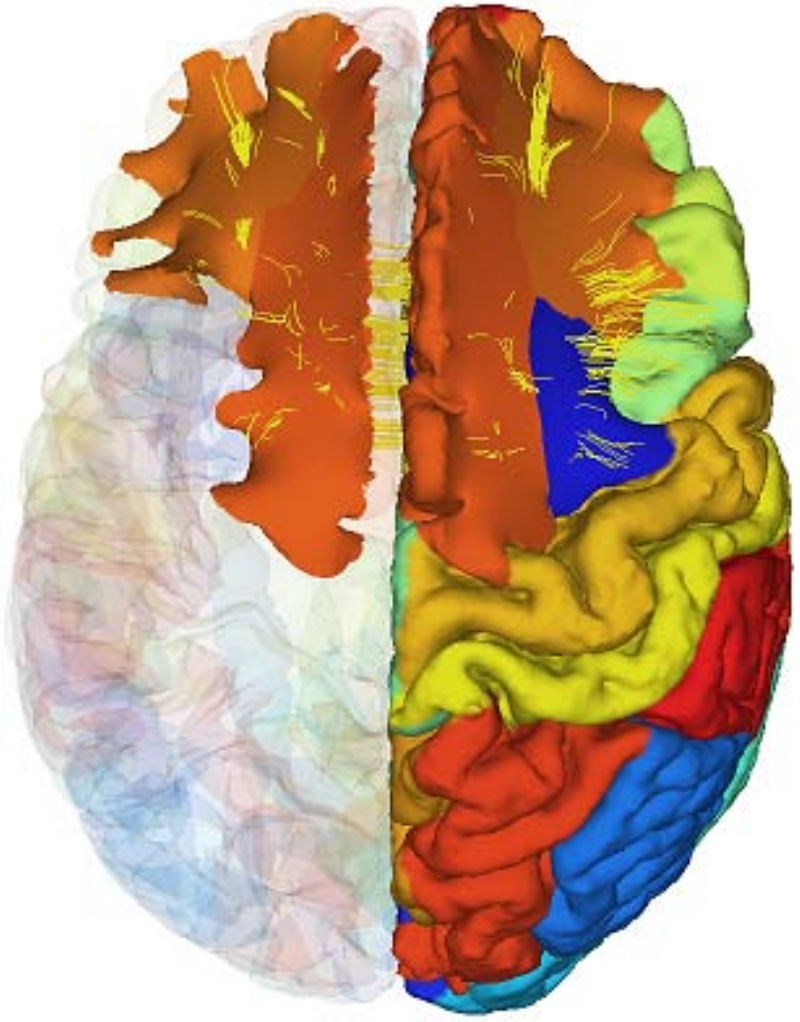}&
\includegraphics[width=0.22\textwidth]{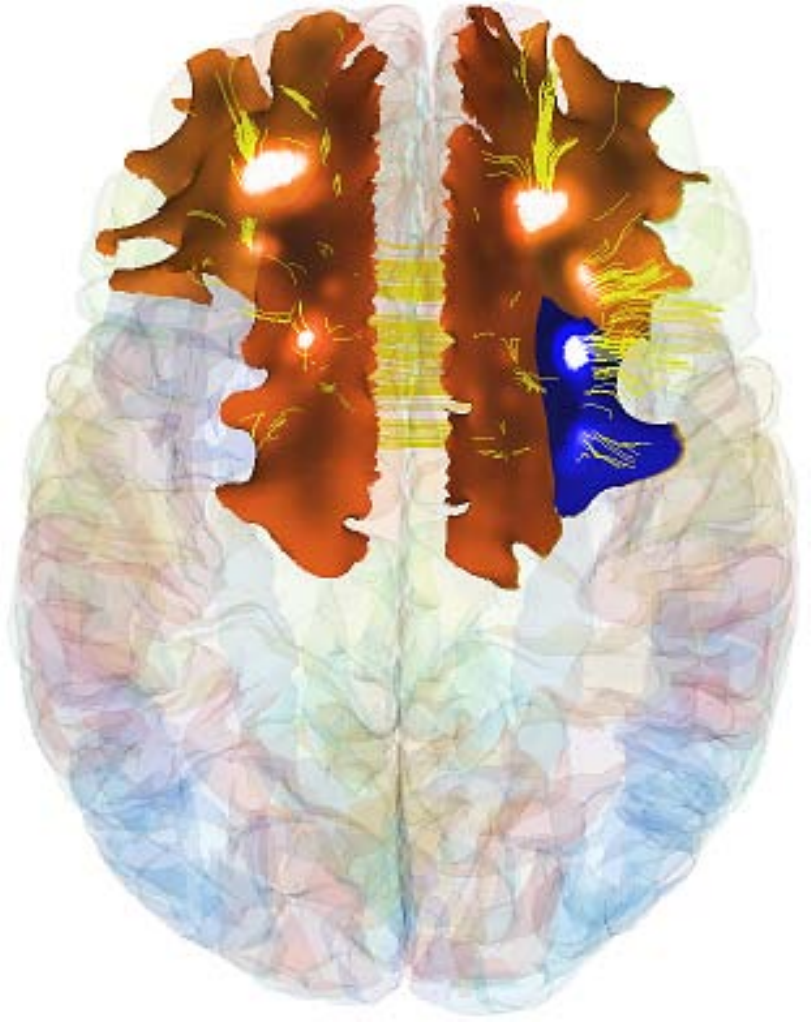}\\
(e) & (f) & (g) & (h)\\
\end{tabular}
\end{center}
\caption{Multimodal brain visualizations using local mappings. (a) Brain cortical surface overlaid with cortical parcellation. The same color regions on the left and right hemispheres represent the counterparts on the respective hemispheres. (b) Structural connectivity from the diffusion MRI data is represented as fiber tracts, seeded at the corpus callosum. Left hemisphere (c) original surface and the locally mapped surface at the selected region (d) using standard surface shading and (e) using normal mapped shading from the original mesh. (f) The longest structural fiber connections to this selected region are shown in yellow with corresponding locally mapped left hemisphere surface. (g) The corresponding connections and the locally mapped surfaces on the right hemisphere. (h) Functional connectivity data is overlaid on the locally mapped surfaces, \emph{with white representing the most functionally-activated regions and black indicating the least; the seed region is the upper white region on the left locally mapped surface.} LMap computation for the two ROIs took 1.7 secs.
\label{fig:multimodal_brain}}
\end{figure*}

\begin{figure}[ht!]
\begin{center}
\begin{tabular}{cc}
\includegraphics[width=0.23\textwidth]{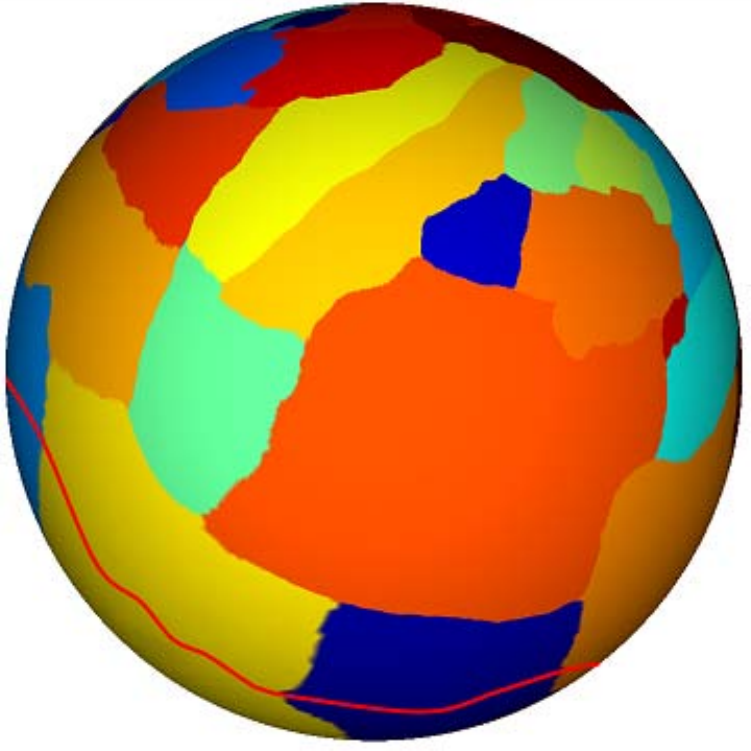}&
\includegraphics[width=0.235\textwidth]{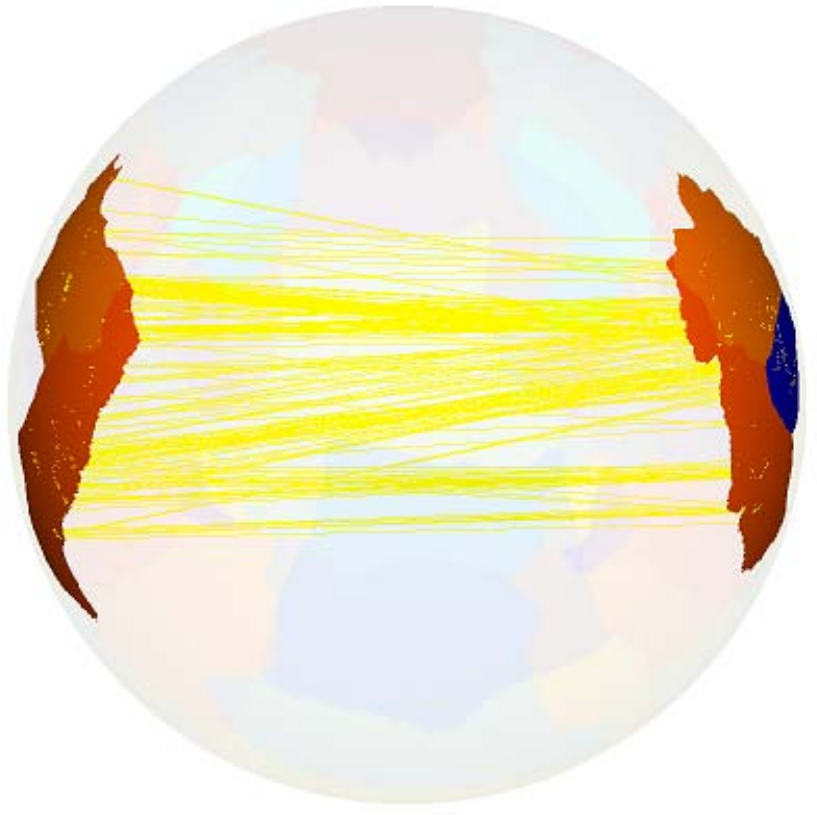}\\
(a) & (b)
\end{tabular}
\end{center}
\vspace{-3mm}
\caption{Multimodal brain visualizations comparison. (a) Angle- and area-preserving spherical parameterization \cite{nadeem2016spherical} of brain cortical surface in Figure~\ref{fig:multimodal_brain}a; the red curve represents the conformally welded boundaries of the left and right hemispheres. (b) The longest structural fiber connections to the same selected regions, as shown in Figure~\ref{fig:multimodal_brain}h.
\label{fig:multimodal_brain_comp}}
\vspace{-5mm}
\end{figure}
\subsection{Evaluation Method}
The LMap algorithm was developed using generic C++ on the Windows 7 platform. The Matlab C++ library was used for solving the linear system. All of the experiments are conducted on a workstation with a Core 2 Quad 2.50 GHz CPU with 8GB RAM. 

\begin{figure*}[t!]
\begin{center}
\begin{tabular}{cccc}
\includegraphics[width=0.23\textwidth]{figures/molecular_surface/1rwe-3-beta}&
\includegraphics[width=0.23\textwidth]{figures/molecular_surface/1rwe-3-beta_roi}&
\includegraphics[width=0.23\textwidth]{figures/molecular_surface/1rwe-3-beta_local}&
\includegraphics[width=0.23\textwidth]{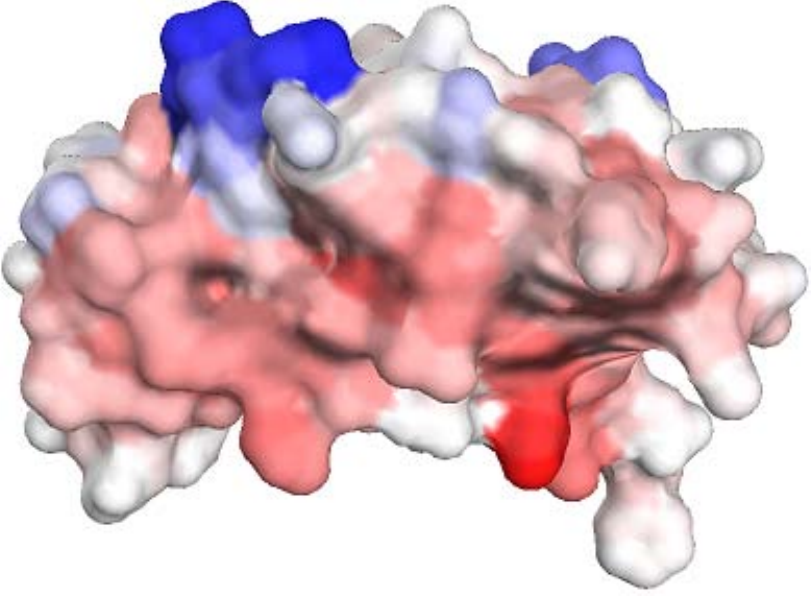}\\
(a) & (b) & (c) & (d)\\
\includegraphics[width=0.23\textwidth]{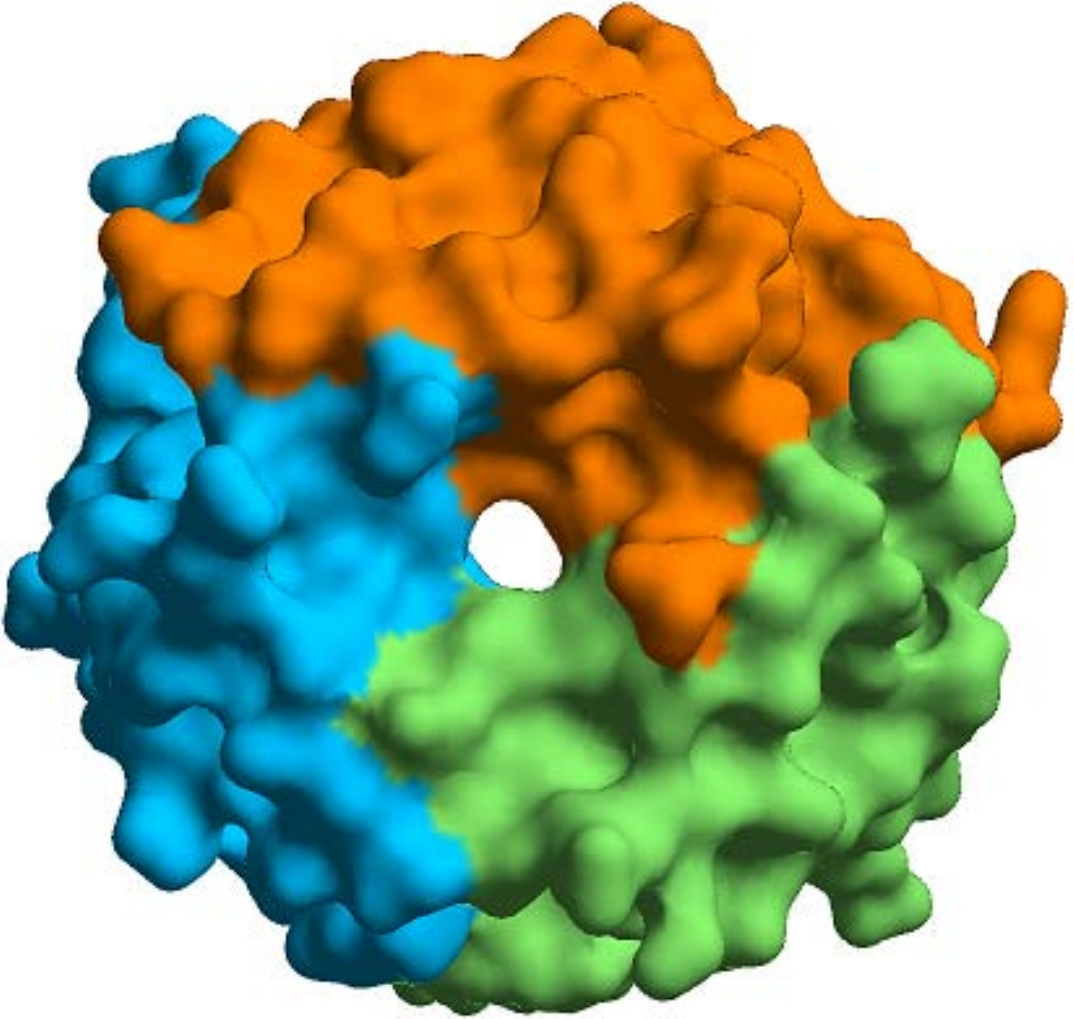}&
\includegraphics[width=0.23\textwidth]{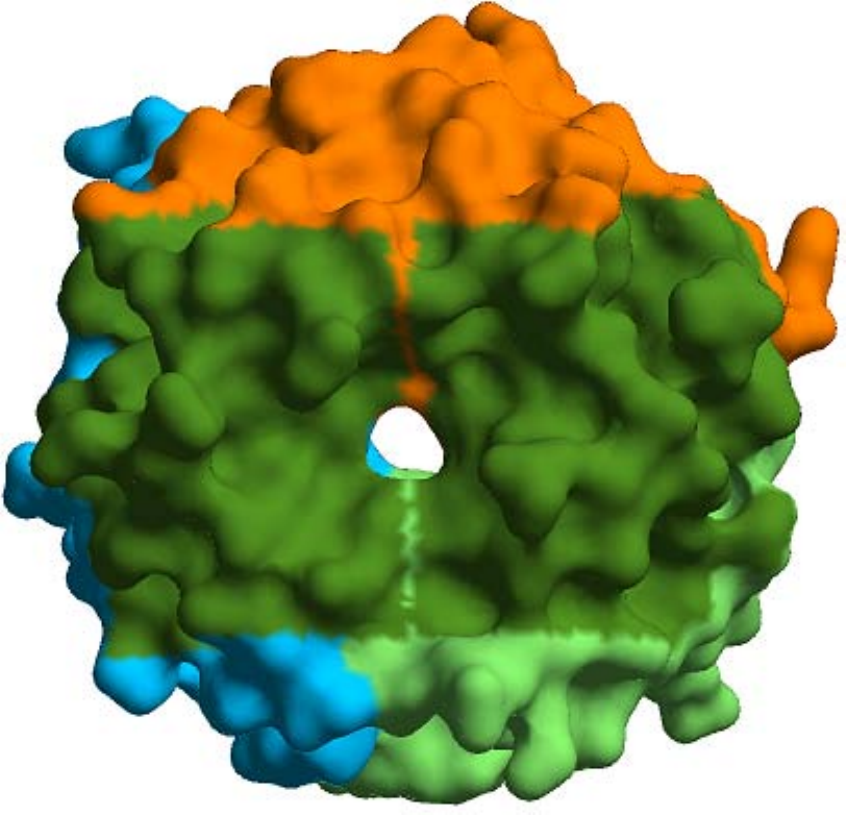}&
\includegraphics[width=0.23\textwidth]{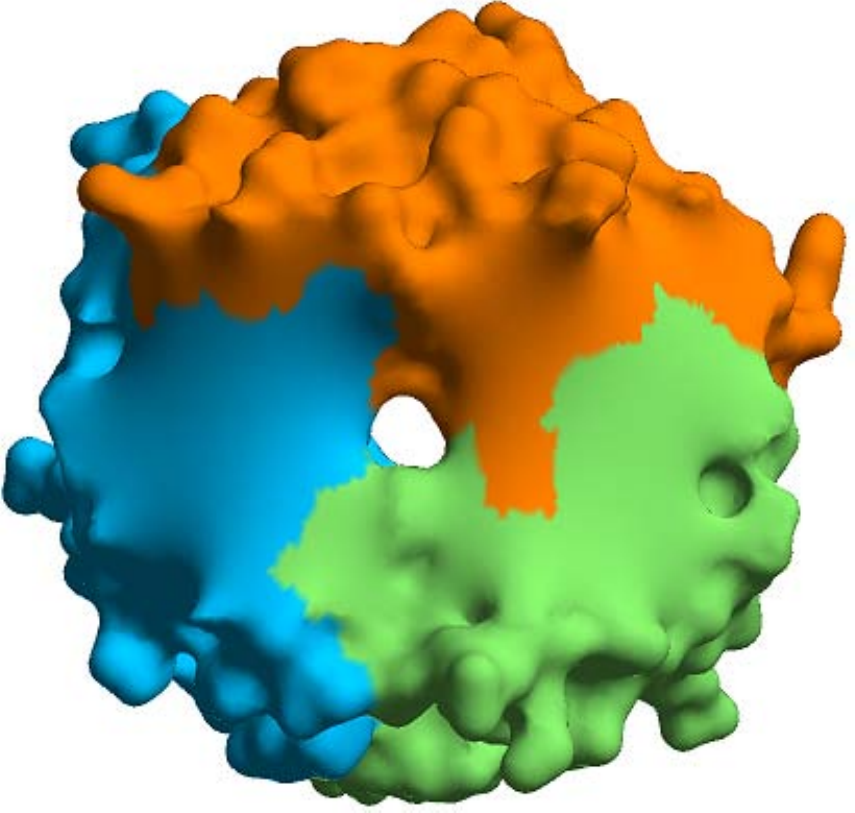}&
\includegraphics[width=0.23\textwidth]{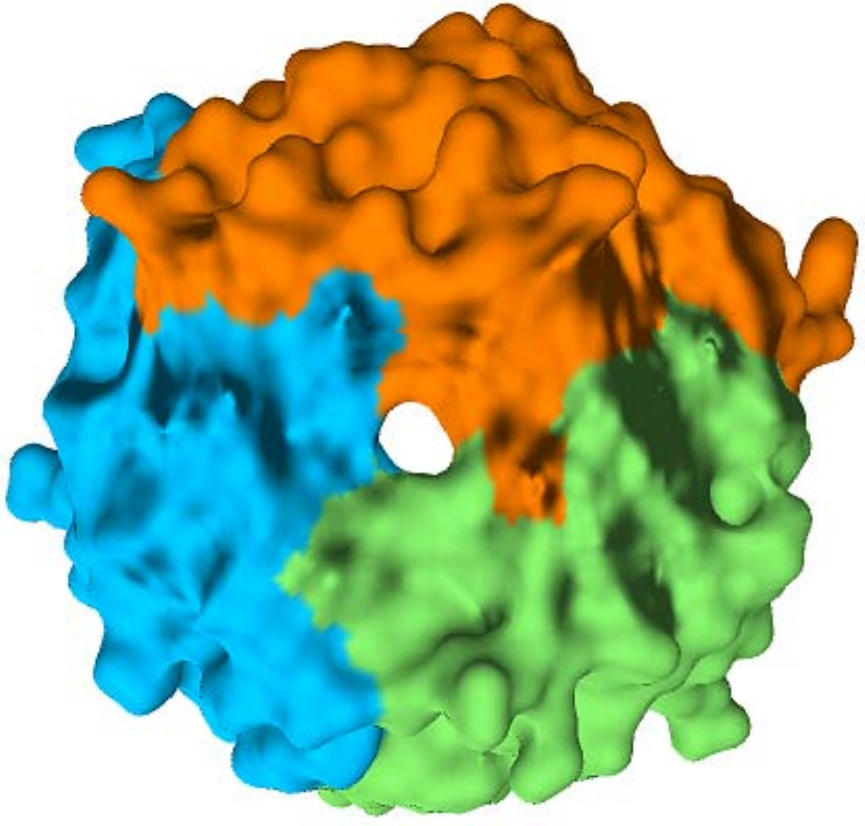}\\
(e) & (f) & (g) & (h)\\
\includegraphics[width=0.23\textwidth]{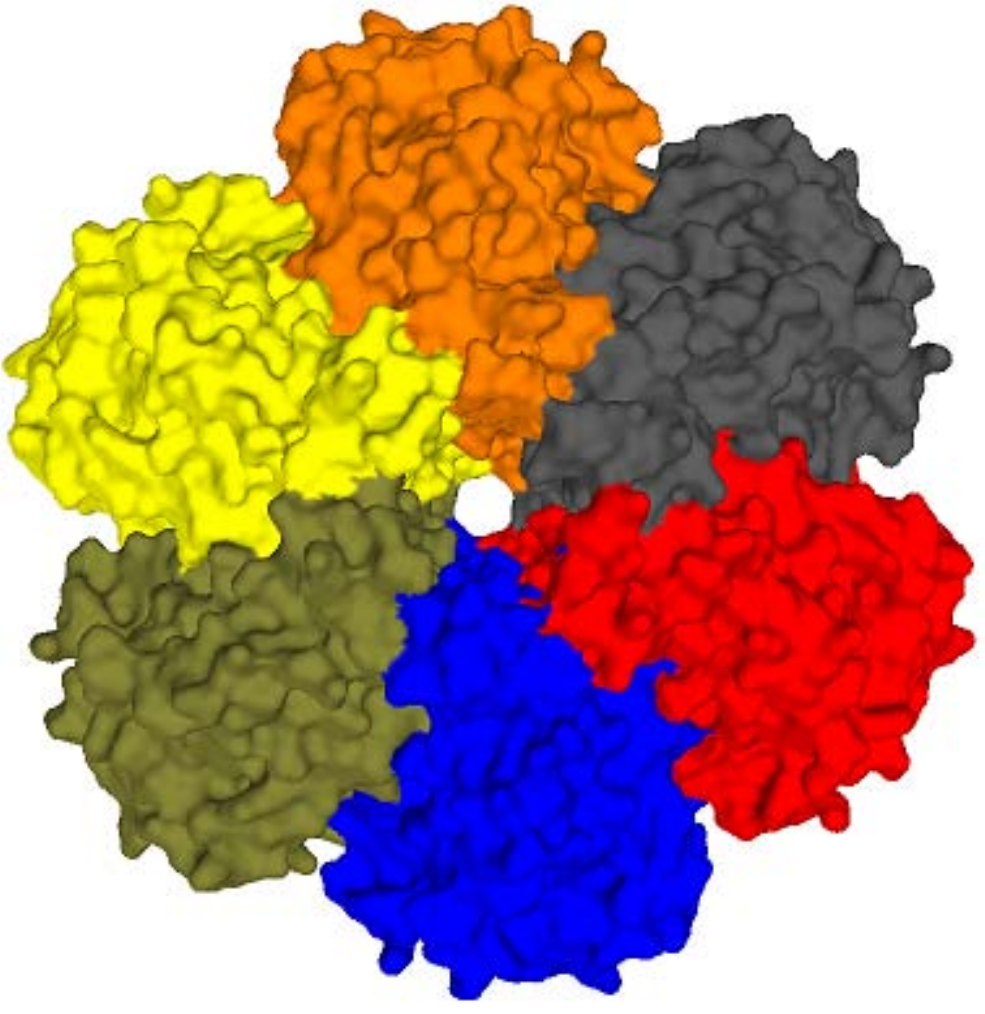}&
\includegraphics[width=0.23\textwidth]{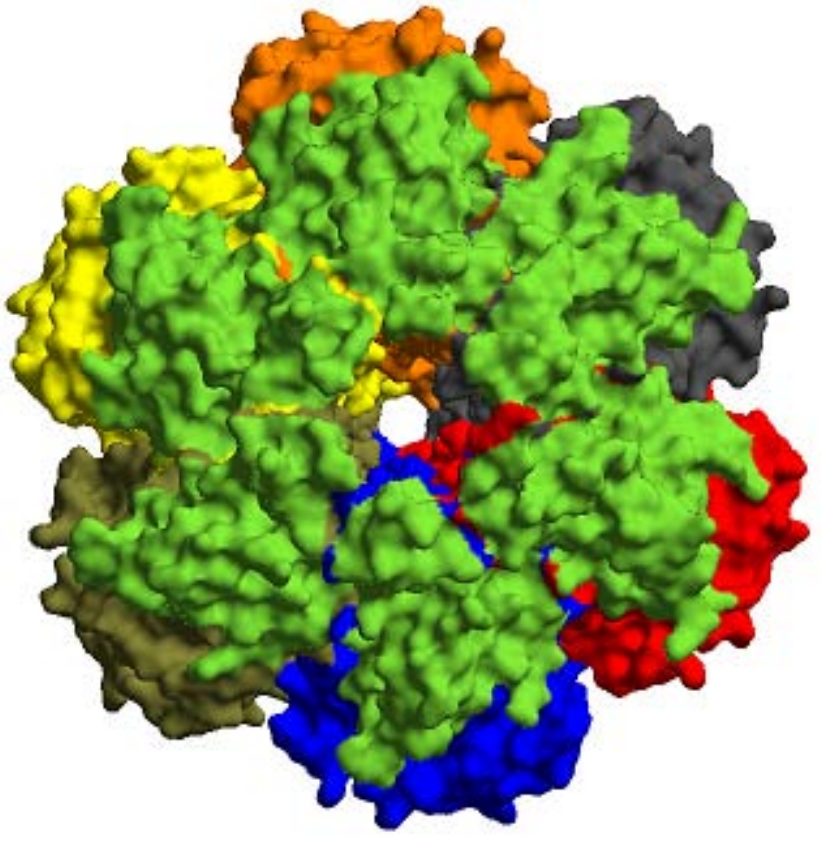}&
\includegraphics[width=0.23\textwidth]{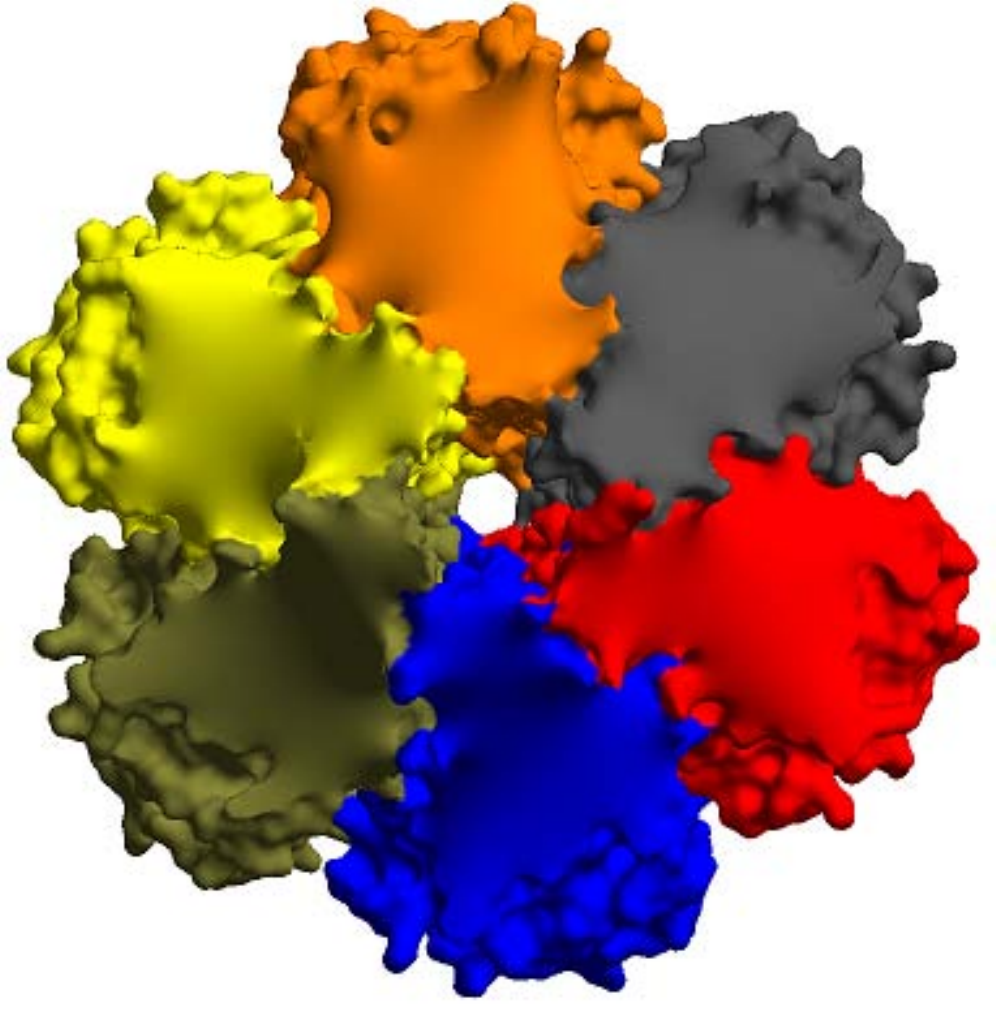}&
\includegraphics[width=0.23\textwidth]{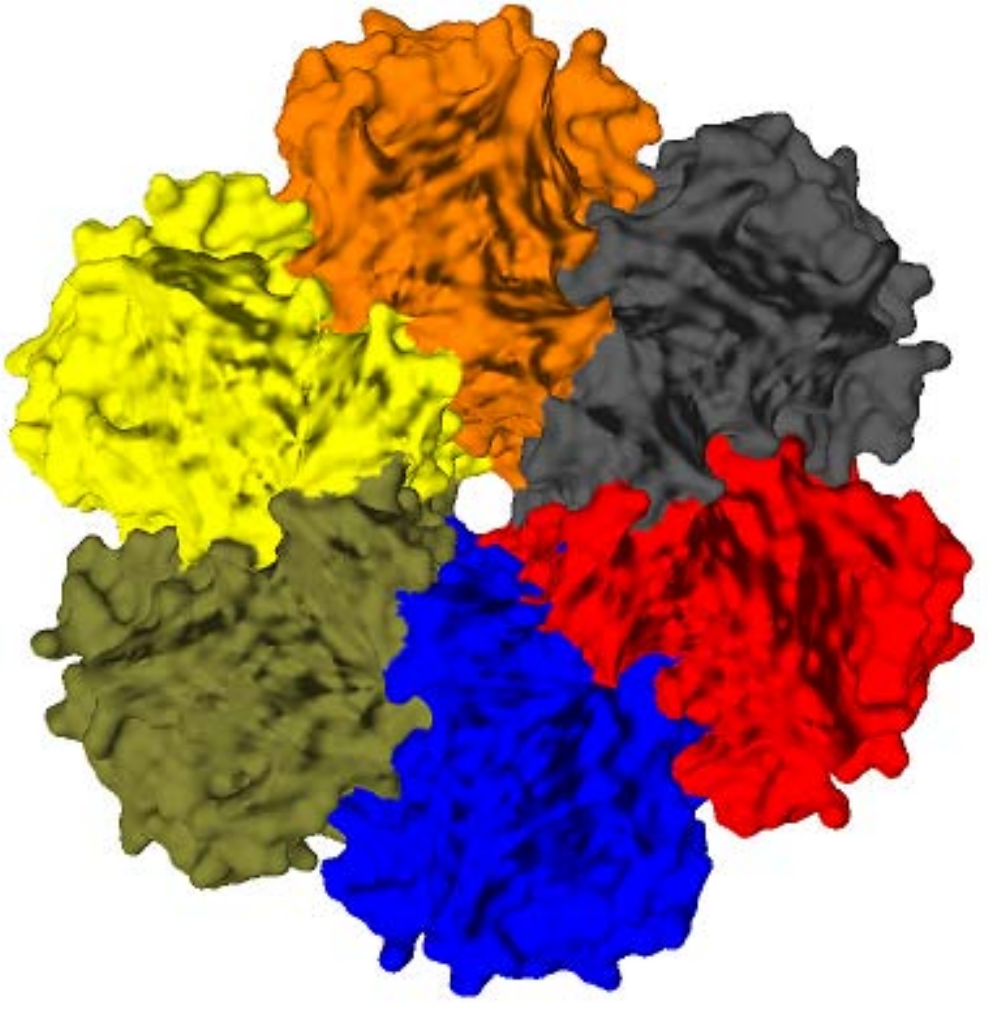}\\
(i) & (j) & (k) & (l)\\
\end{tabular}
\end{center}
\vspace{-3mm}
\caption{Molecular surface local mappings. (a) A small protein (genus 0) colored by temperature factor (PDB ID: 1RWE), with red the highest and blue the lowest temperature factors. Local mapping for a selected ROI (b) using standard surface shading (c) and using normal mapped shading from the original mesh (d). (e) Protein (genus 1) with a channel (PDB ID: 2BT9). The coloring shows the three amino acid chains that form the protein. Local mapping for selected ROIs (f) using standard surface shading (g) and using normal mapped shading from the original mesh (h). (g) Protein (genus 65) with an intricate network of channels (PDB ID: 1GKI). The coloring shows the six amino acid chains that form the protein. Local mapping for selected ROIs (j) using standard surface shading (k) and using normal mapped shading from the original mesh (l). LMap computation for ROIs in (b), (f), and (j) took 1.3, 2.0, and 3.8 secs, respectively.
\label{fig:molecular_surface}}
\vspace{-2mm}
\end{figure*}

We compare LMap against the spin transformations technique \cite{crane2011spin}. We refer to the spin transformations with mean curvature for the whole input surface as \emph{global} spin transformations and the ones with mean curvature restricted to the selected ROI as \emph{local} spin transformations. For a selected ROI on the molecular surface shown in Figure~\ref{fig:molecular_surface}b, we compute the corresponding local spin transformations where mean curvature is set to zero everywhere except the selected ROI (Figures~\ref{fig:spin_transformations}e--\ref{fig:spin_transformations}h) and global spin transformations (Figures~\ref{fig:spin_transformations}a--\ref{fig:spin_transformations}d) with varying scale factors. We can observe that the local spin transformations for a selected ROI can significantly distort the scale of the rest of the surface and can lead to self-intersections with larger scale factors. In contrast, LMap does not affect the rest of the surface while parameterizing and deforming the selected ROI, as shown in Figures~\ref{fig:spin_transformations}i--\ref{fig:spin_transformations}l. Moreover, the LMap is stable to perturbations of ROI, as shown in Figure~\ref{fig:lmap_stability}. LMap can also handle sharp bends, as demonstrated in Figure~\ref{fig:lmap_sharp_bends}.

In theory, the Ricci flow method induces a conformal mapping. A conformal map preserves angles and local shapes. The shape-preserving property is crucial for visualization purposes. In order to evaluate the conformality of the mapping result, we compute both angle and area distortions. The area distortion is computed as follows. Assume the parameterization is $\phi: M \rightarrow \mathbb{S}$. For each vertex $v_i$, the \emph{area distortion} is defined as
\[
    \epsilon_i := \log \frac{\sum_{j,k} A([\phi(v_i),\phi(v_j),\phi(v_k)])}{\sum_{j,k} A([v_i,v_j,v_k])}
\]
where $A(.)$ represents the area of a triangle, and $[v_i,v_j,v_k]$ is the triangle formed by ${v_i,v_j,v_k}$. We then plot the histograms of ${\epsilon_i}$. Similarly the \emph{angle distortion} at a corner angle is given by
\[
    \eta_{ijk} := \log \frac{\angle \phi(v_i)\phi(v_j)\phi(v_k)}{\angle v_i v_j v_k},
\]
we then plot the histograms of ${\eta_{ijk}}$. 

The angle-preserving (conformal) mapping should ideally be close to zero angle distortions everywhere, whereas the area-preserving mapping should be close to zero area distortions everywhere. We compute the area and angle distortion histograms for the local selected ROI (Figure~\ref{fig:molecular_surface}b) on mapped molecular surfaces computed using LMap (Figure~\ref{fig:molecular_surface}b), global spin transformation with scale factor 6 (Figure~\ref{fig:spin_transformations}c) and local spin transformation with scale factor 6 (Figure~\ref{fig:spin_transformations}g). As shown in Figure~\ref{fig:histograms}, for a given ROI, LMap preserves angle and area better than both global and local spin transformations.

\begin{figure*}[th!]
\begin{center}
\begin{tabular}{ccc}
\includegraphics[width=0.31\textwidth,height=0.2\textwidth]{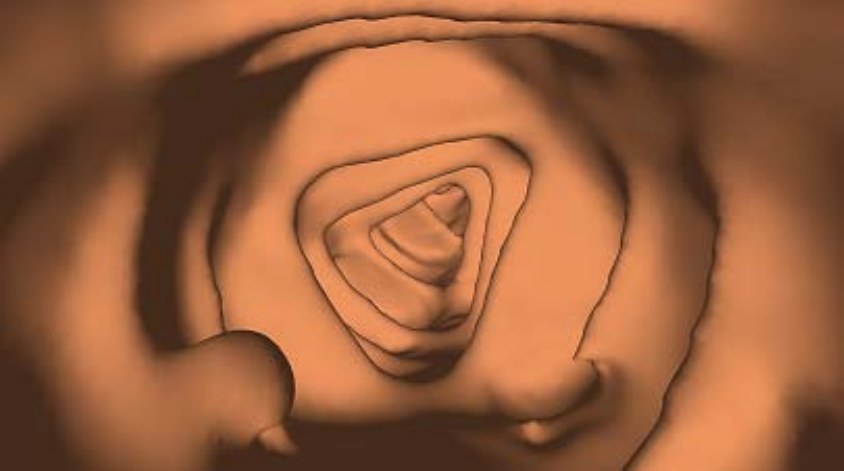}&
\includegraphics[width=0.31\textwidth,height=0.2\textwidth]{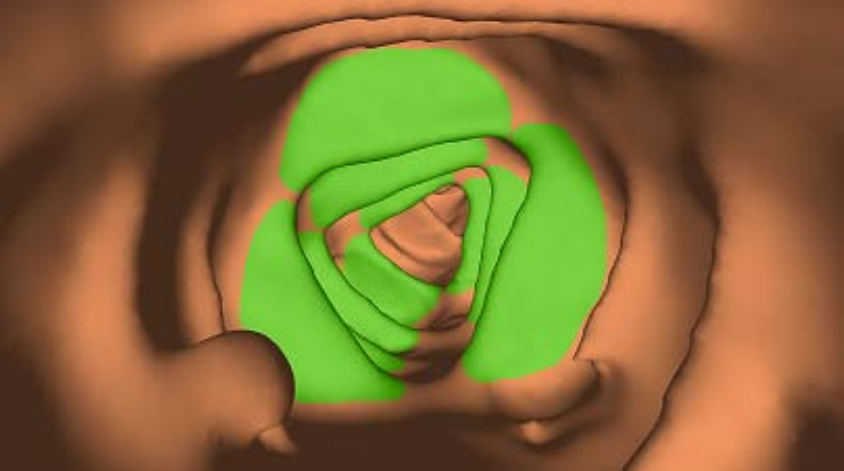}&
\includegraphics[width=0.31\textwidth,height=0.2\textwidth]{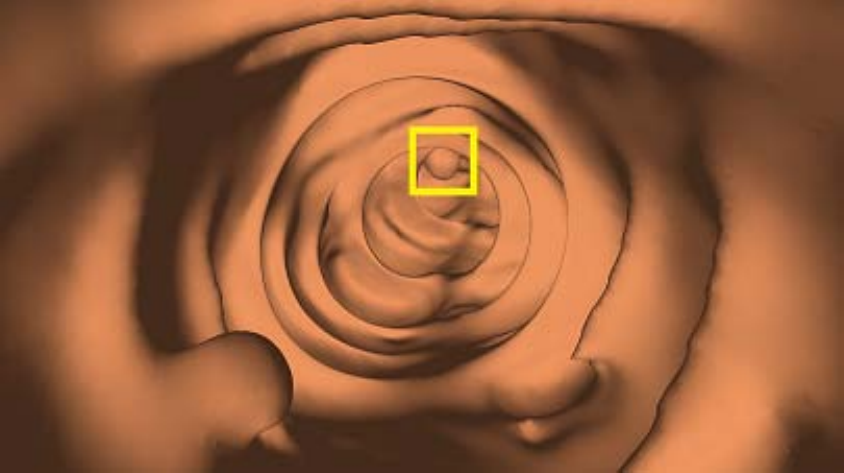}\\
(a) & (b) & (c)\\
\includegraphics[width=0.31\textwidth,height=0.2\textwidth]{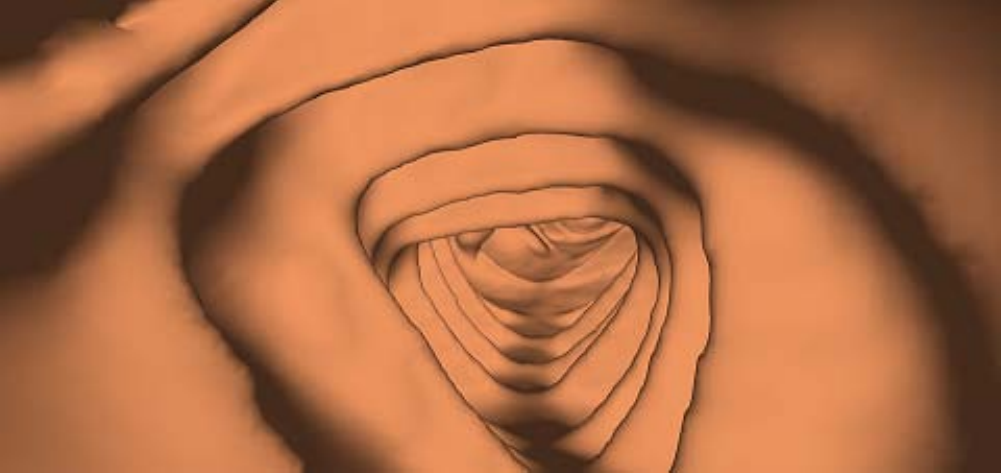}&
\includegraphics[width=0.31\textwidth,height=0.2\textwidth]{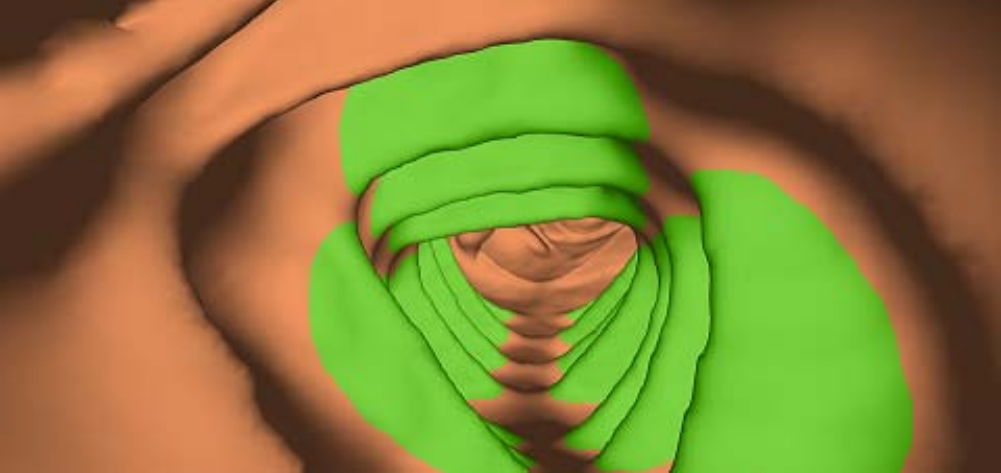}&
\includegraphics[width=0.31\textwidth,height=0.2\textwidth]{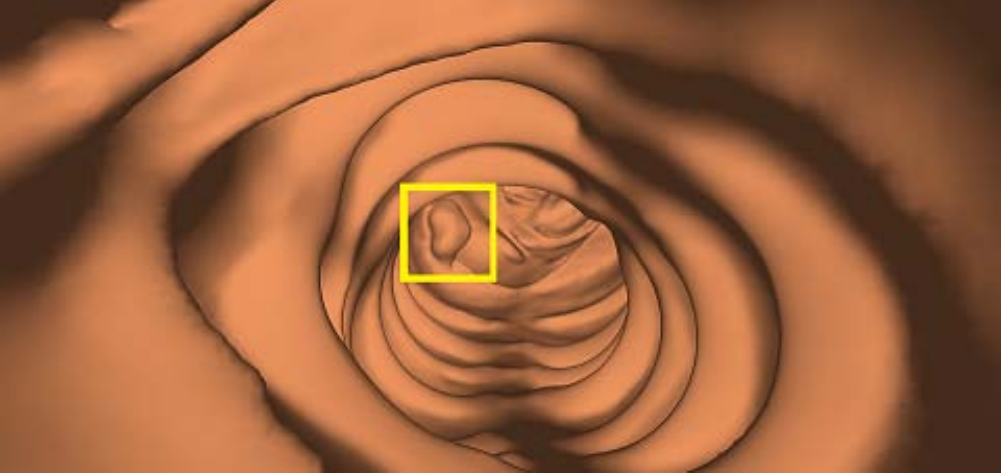}\\
(d) & (e) & (f)\\
\end{tabular}
\begin{tabular}{cccc}
\includegraphics[width=0.226\textwidth,height=0.16\textwidth]{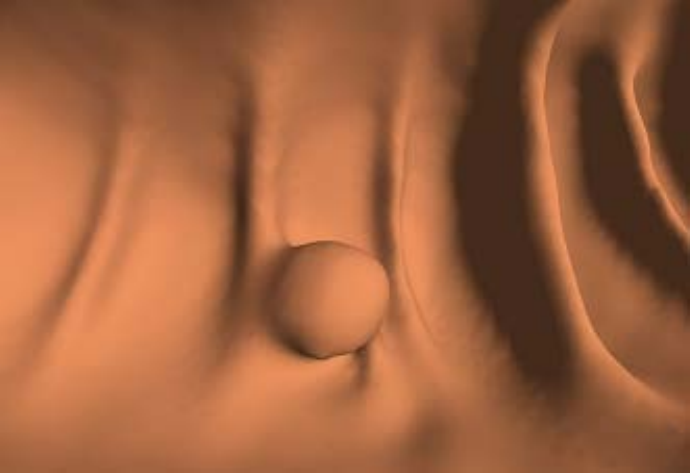}&
\includegraphics[width=0.226\textwidth,height=0.16\textwidth]{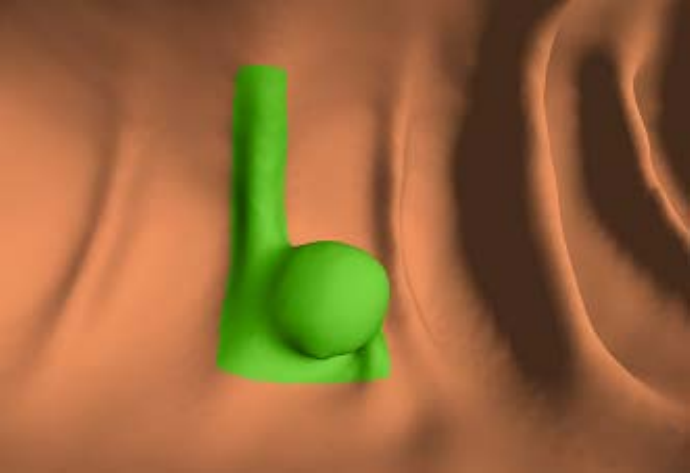}&
\includegraphics[width=0.226\textwidth,height=0.16\textwidth]{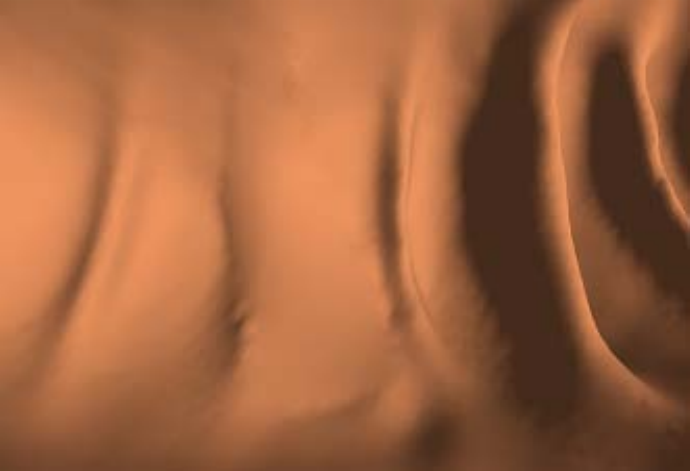}&
\includegraphics[width=0.226\textwidth,height=0.16\textwidth]{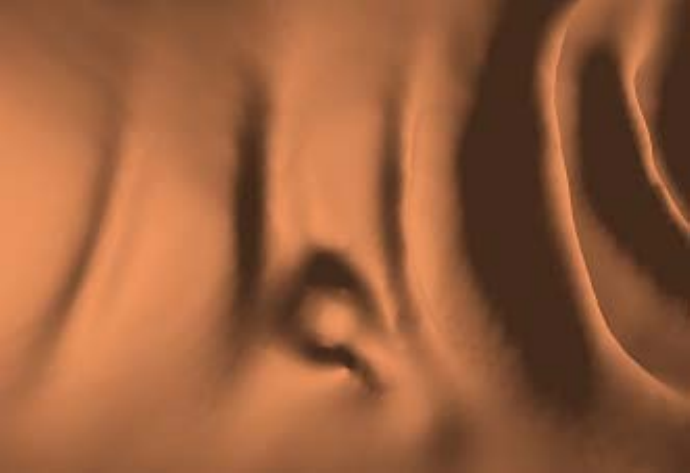}\\
(g) & (h) & (i) & (j)\\
\end{tabular}
\begin{tabular}{cc}
\includegraphics[width=0.47\textwidth]{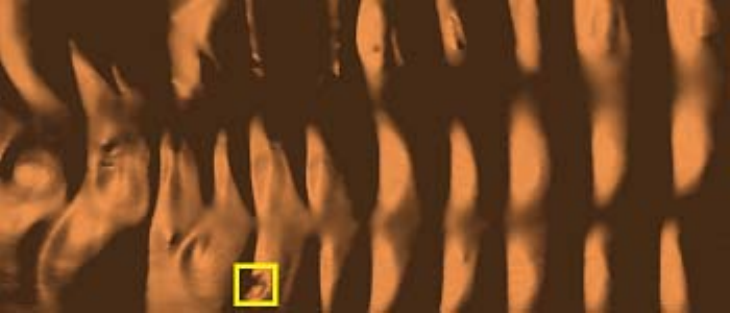}&
\includegraphics[width=0.47\textwidth]{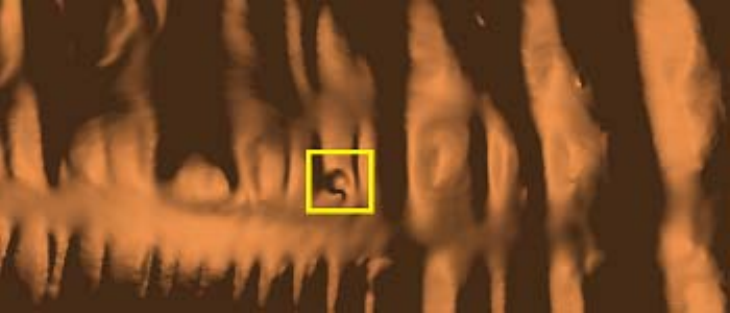}\\
(k) & (l)\\
\end{tabular}
\end{center}
\vspace{-2mm}
\caption{Optimal coverage in virtual colonoscopy. (a) Virtual colonsocopy mesh with two polyps, one in view and another one occluded by a far fold. (b) Highlighted folds between the near and the far polyps are conformally deformed via LMap and (c) the far polyp is exposed (highlighted with a yellow bounding box) while preserving the overall geometric context. The coverage of this colon segment increased from 72\% in (a) and (b) to 94\% using LMap in (c). (d) Polyp occluded by the far fold and the corresponding LMap on the (e) highlighted folds in green remove the occlusion and (f) exposes the polyp (highlighted with a yellow bounding box). The coverage of this colon segment is increased from 76\% in (d) and (e) to 91\% in (f). (g) Polyp on the fold. (h) Highlighted polyp in green and the corresponding fold are conformally deformed via LMap (and pushed towards the colon wall) to increase coverage and the result is shown (i) using standard surface shading and (j) using normal mapped shading from the original mesh. The shape of the polyp is preserved in (j), owing to the angle-preserving property of LMap. The folds in (b), (e), and (h) are automatically detected using the method presented in our earlier work \cite{nadeem2017corresponding}. (k) Angle-preserving flattening \cite{nadeem2017corresponding} of colon segment in (f) with the polyp highlighted with a yellow bounding box. (l) Flattened colon segment in (g) with the polyp highlighted with a yellow bounding box. LMap computation for ROIs in (b), (e), and (h) took 1.8, 2.3, and 1.2 secs, respectively.
\label{fig:colon_coverage}}
\end{figure*}


\begin{figure*}[t!]
\begin{center}
\begin{tabular}{ccc}
\includegraphics[width=0.25\textwidth]{figures/molecular_surface/1rwe-3-beta}&
\includegraphics[width=0.19\textwidth]{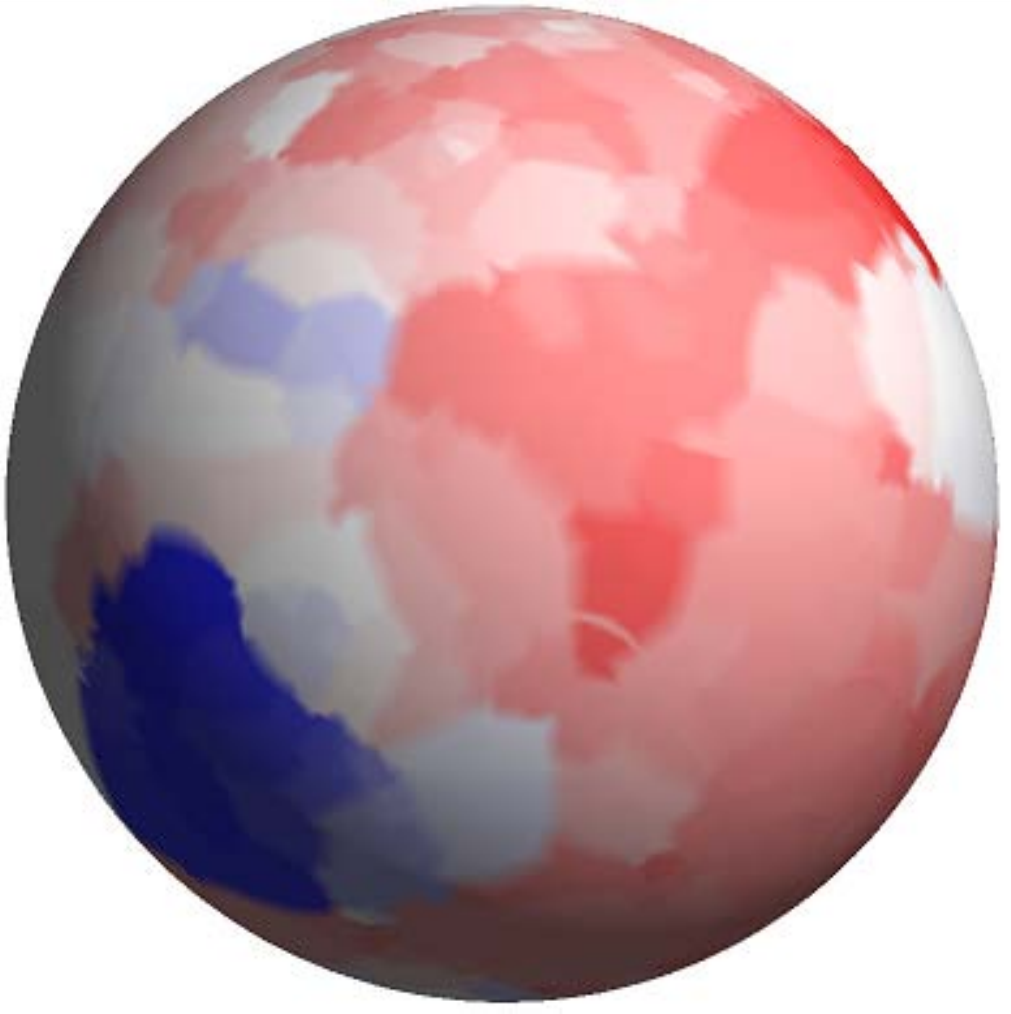}&
\includegraphics[width=0.48\textwidth]{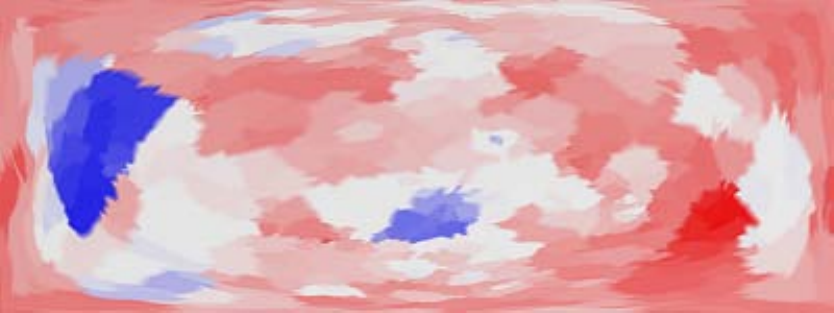}\\
(a) & (b) & (c)\\
\includegraphics[width=0.22\textwidth]{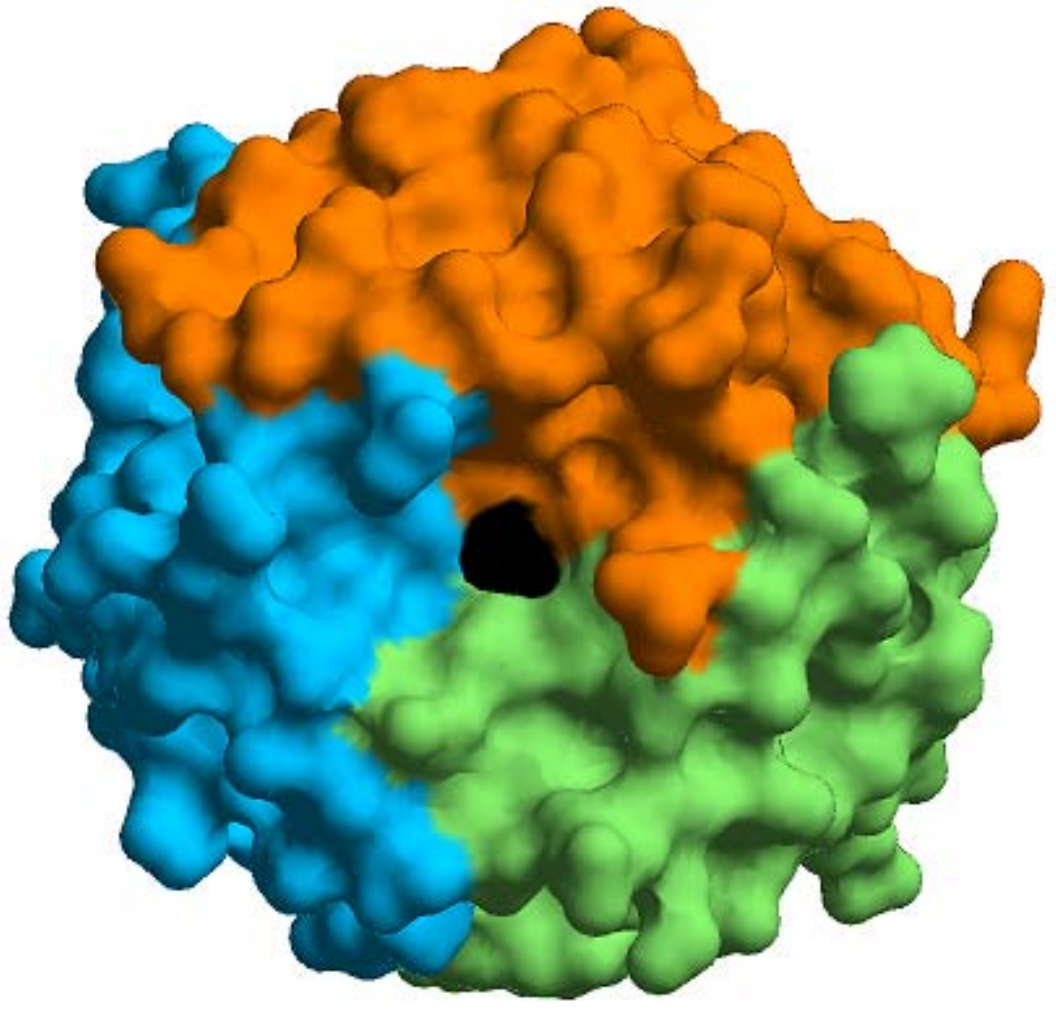}&
\includegraphics[width=0.195\textwidth]{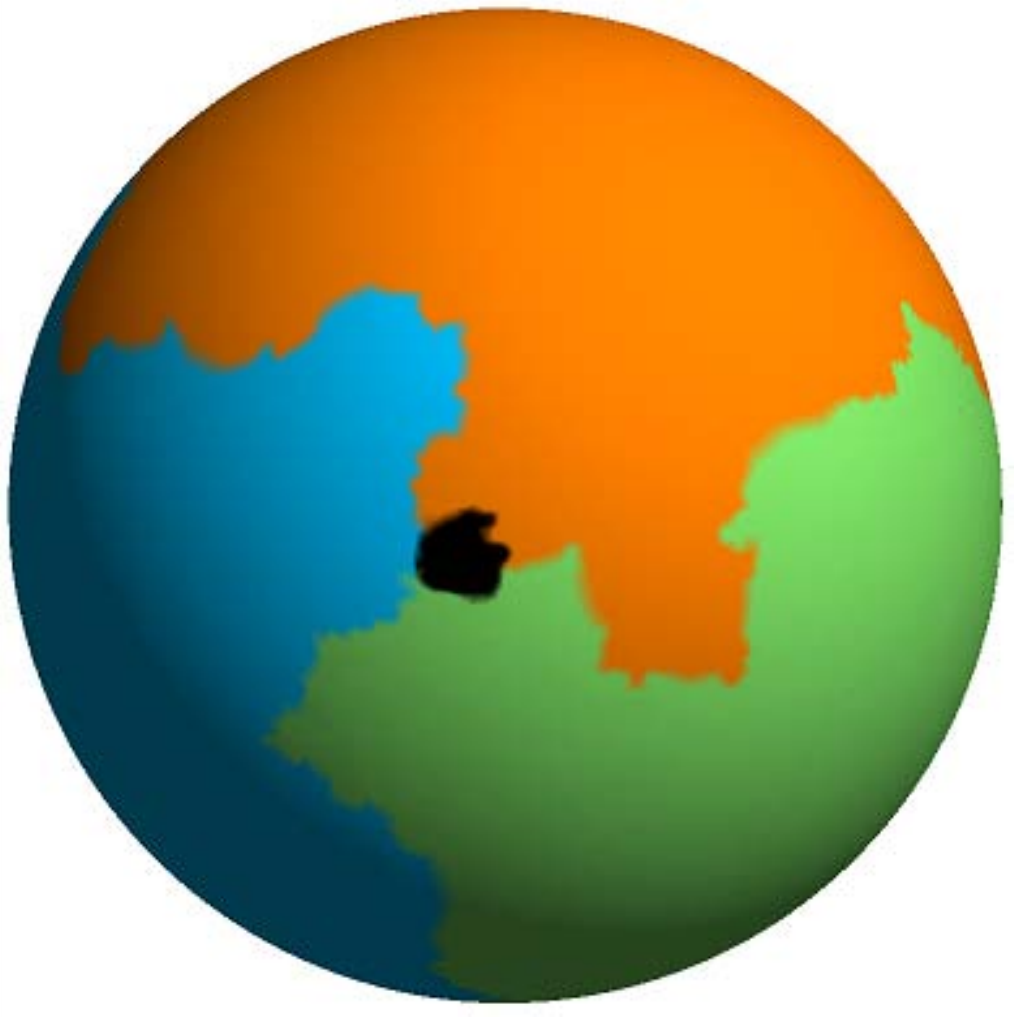}&
\includegraphics[width=0.48\textwidth]{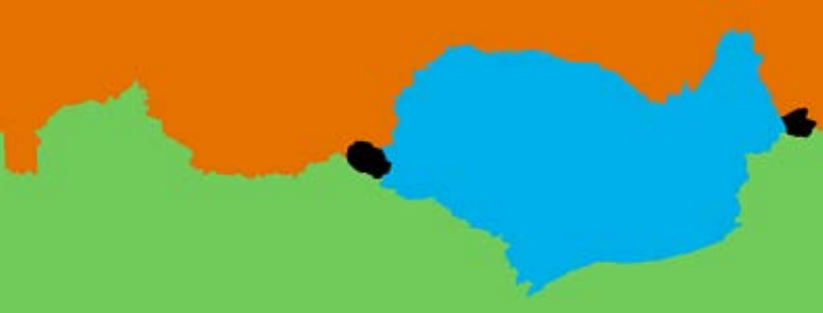}\\
(d) & (e) & (f)\\
\end{tabular}
\end{center}
\vspace{-3mm}
\caption{Molecular surface maps \cite{krone2017molecular} comparison. (a) A protein (PDB ID: 1RWE) is mapped to (b) a sphere using the force-directed method and (c) projected using the Lambert equal-area projection (as outlined in Krone et al. \cite{krone2017molecular}). (d) A protein (PDB ID: 2BT9), with channel removed and the resultant holes filled (shown in black), is mapped to (e) a sphere with the parameter-based method and (f) projected using the Plate Carr\'{e}e projection (as done in Krone et al. \cite{krone2017molecular}).
\label{fig:molecular_surface_map}}
\vspace{-3mm}
\end{figure*}

\begin{figure}[t!]
\begin{center}
\begin{tabular}{cc}
\includegraphics[width=0.23\textwidth]{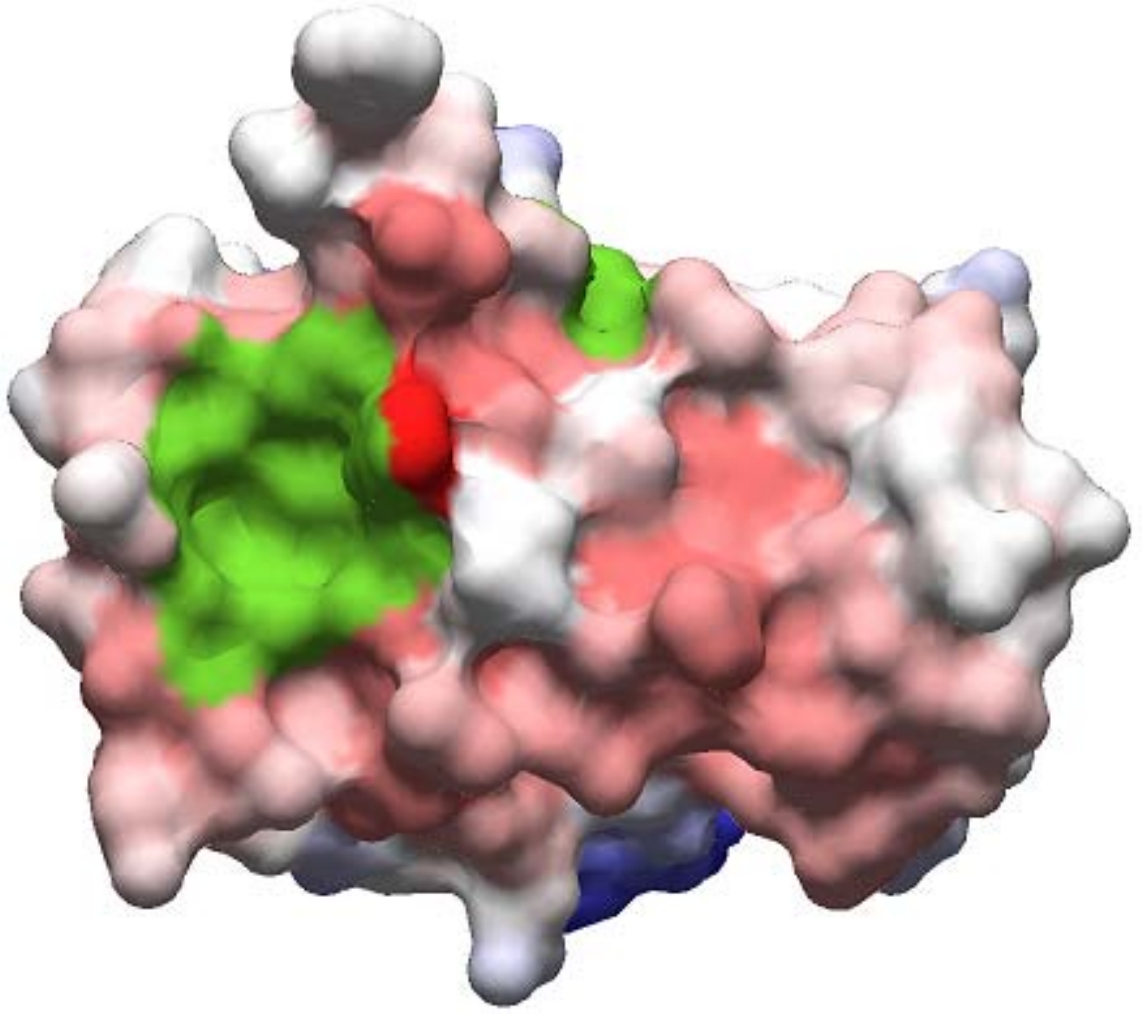}&
\includegraphics[width=0.23\textwidth]{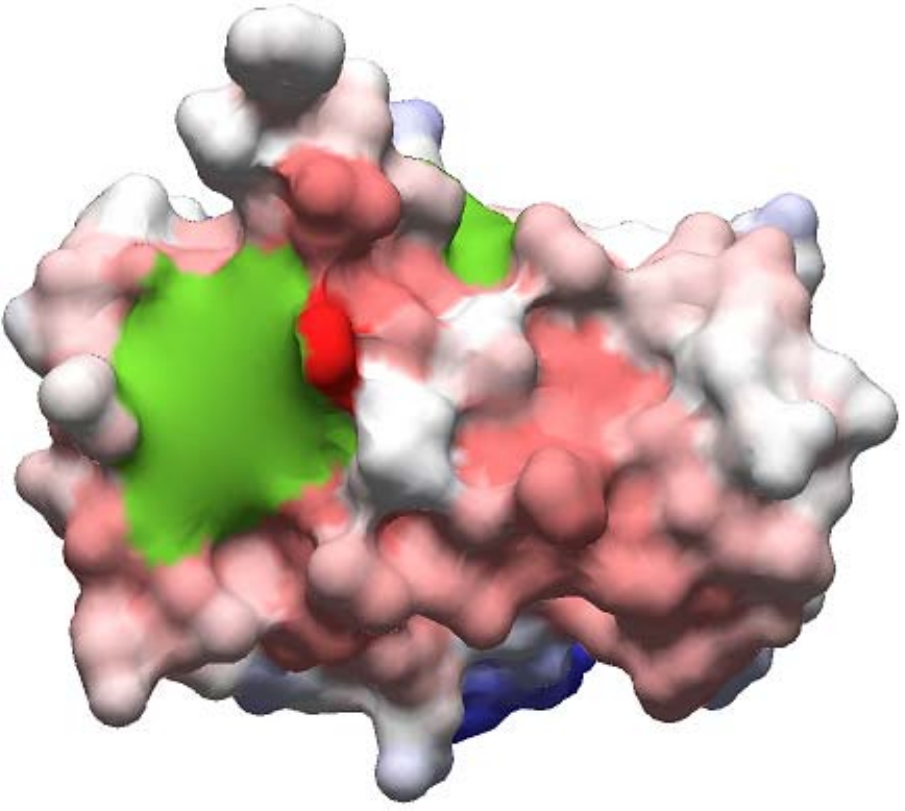}\\
(a) & (b)\\
\includegraphics[width=0.23\textwidth]{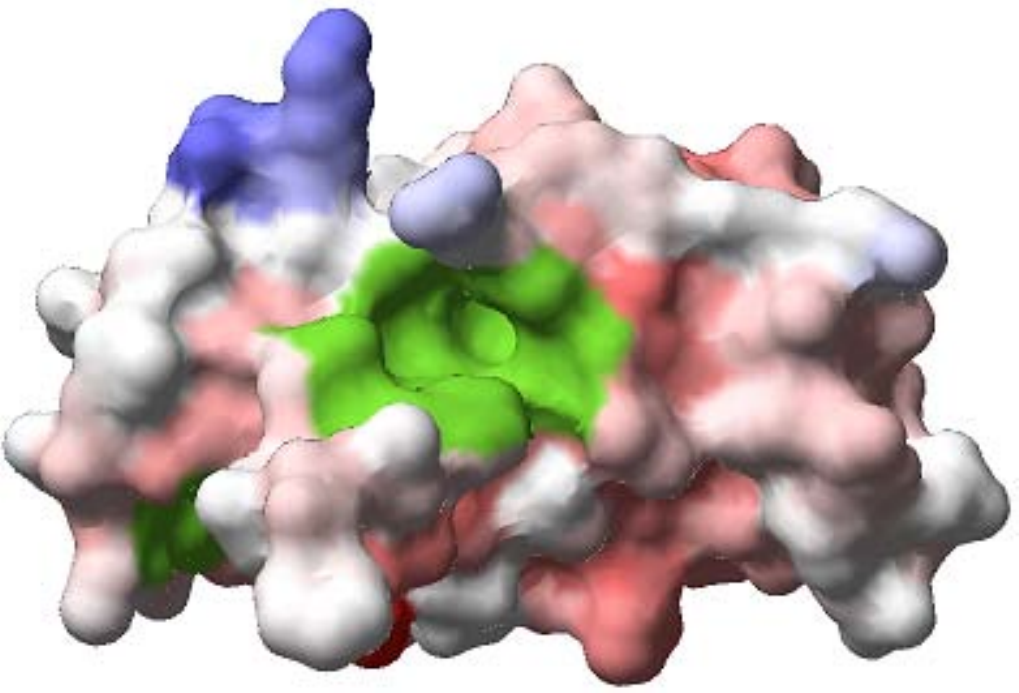}&
\includegraphics[width=0.23\textwidth]{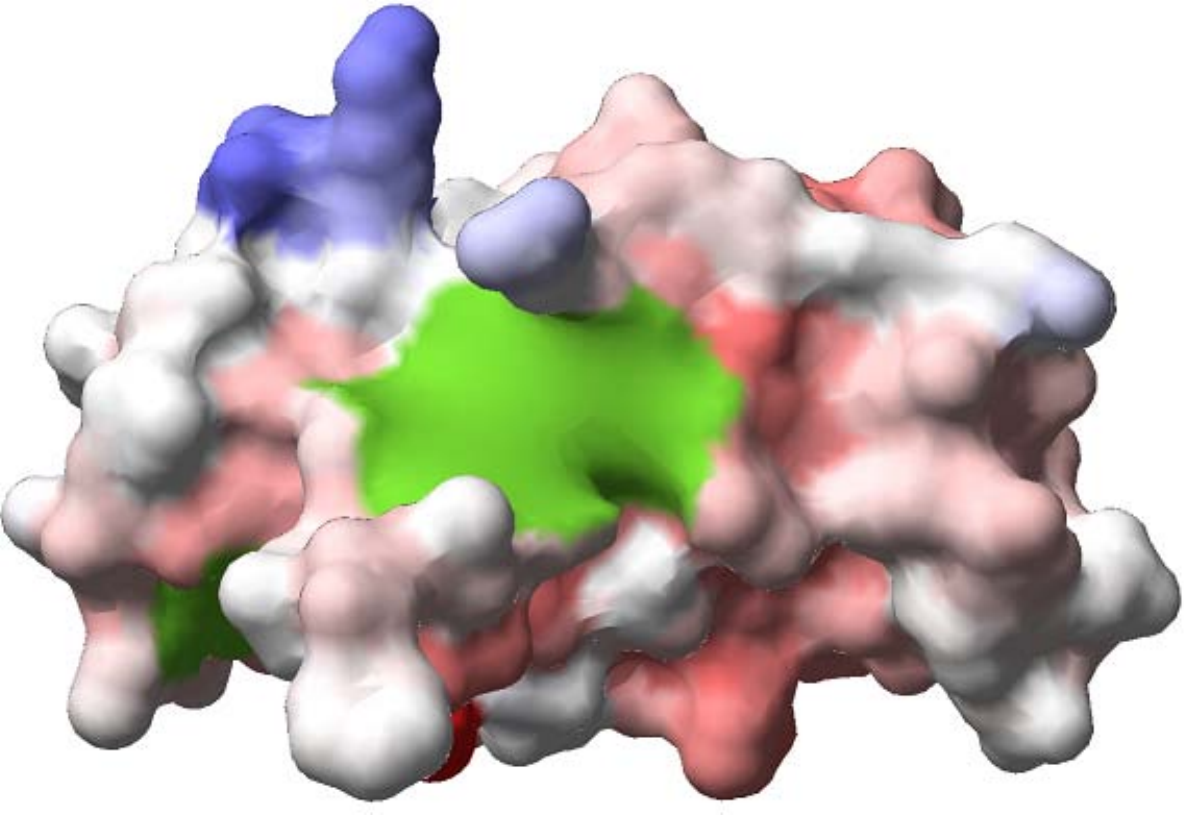}\\
(c) & (d)
\end{tabular}
\end{center}
\vspace{-3mm}
\caption{Molecular surface local mappings for cavities (a) and (c), colored in green. The corresponding local mappings are shown in (b) and (d).  
\label{fig:molecular_surface_cavities}}
\vspace{-5mm}
\end{figure}

\section{Case Studies}
We present the following three challenging case studies for our LMap algorithm.

\subsection{Multimodal Brain Visualization}
The human brain is complex in its shape and connectivity, even at scales at which magnetic resonance imaging can measure. Visualization is used for exploring possible relationships between various brain structures across scan types (multimodal data) or statistical variables (multivariate data). However, the convoluted brain cortical structures induce considerable occlusions making it difficult to visualize these structures in their original form.

Current global brain mapping techniques resolve these occlusions \cite{nadeem2016spherical} and allow for exploration of multiple modalities on a brain surface \cite{nadeem2016multimodal} but at the expense of loosing the geometric context. Moreover, the biggest problem with the global brain mapping techniques is that all the modalities have to be mapped to the common simplified representation and this can be a major bottleneck. An example of the multimodal brain visualization \cite{nadeem2016multimodal} based on angle- and area-preserving spherical parameterization is presented in Figure~\ref{fig:multimodal_brain_comp}. 

In contrast, LMap overcomes these limitations by locally mapping selected ROIs while preserving the overall geometric context and allowing for simultaneous visualization of cortical features, along with structural and functional connectivity information from diffusion and functional magnetic resonance imaging (MRI) data respectively in the native space, as shown in Figure~\ref{fig:multimodal_brain}. We used the Human Connectome Project (HCP) data \cite{van2013wu} along with DSI Studio \cite{yeh2014dsi} to obtain the highest fidelity structural, functional, and diffusion data for our visualizations. 

We also consulted a neuroscientist for feedback on our multimodal brain visualizations. Since the neuroscientists want to study specific cortical structures and the connectivity patterns between them, the ability of LMap to focus on the simplification of these specific structures while preserving the overall geometric context was highly appreciated. Moreover, the local mappings for arbitrary-sized ROIs in the native space were also acknowledged, along with the possibility of visualizing subcortical features in the native space, and accurately overlaying the multimodal information from the resting-state and task-based functional MRI data, and multivariate information from geometric measures (travel depth, curvature, area, thickness) and important landmarks such as sulcal pits and fundi (Figure~\ref{fig:spherical_local_comp}). Currently, neuroscientists visualize deep folds on brain hemispheres by using inflated cortical surfaces or via planar transformation by removing the software-added tiling (shown in Figure~\ref{fig:spherical_local_comp}e) and manually cutting along the cortical surface to lay out the brain on a plane in the original superior view (shown in Figure~\ref{fig:spherical_local_comp}a). We can achieve the tedious planar transformation by just selecting a larger region without requiring any further input. 

\subsection{Optimal Coverage in Virtual Colonoscopy}
Virtual colonsocopy (VC), or CT colonoscopy, is a non-invasive cancer screening technique to navigate and inspect a 3D reconstructed colon surface for polyps (precursors of colon cancer) and if found, to localize and accurately measure these polyps. As part of the standard VC protocol, a radiologist inspects the surface while flying through the colon along the centerline from rectum to cecum and back in both supine (face up) and prone (face down) patient colon datasets. These four centerline flythroughs provide the optimal coverage of the colon surface and hence, minimize the probability of missing a polyp occluded by a fold. In total, a radiologist can spend between 15-30 minutes on these four flythroughs. With the deluge of non-invasive imaging, this interpretation time is a major bottleneck for these radiologists.

Global planar parameterization (or flattening approaches) can be used to completely resolve the occlusions in this context \cite{nadeem2017corresponding} and maximize the coverage. However, the problem with these approaches is that the resultant surface completely looses the geometric context and induces significant areal distortion since all these approaches are only angle-preserving (Figures~\ref{fig:colon_coverage}k and \ref{fig:colon_coverage}l), making these approaches less suitable for polyp localization and visualization tasks; it is for this reason that these flattening approaches are only used for registration purposes. Moreover, these approaches require teniae coli extraction (for boundary mapping) which might not be possible in most cases \cite{nadeem2017corresponding} where the colon is under-distended (a common occurrence). 

We show the effectiveness of LMap in this context by deforming only the detected haustral folds \cite{nadeem2017corresponding} in place while leaving the original geometry of the polyps and the overall colon intact, thus allowing for accurate polyp localization and measurements, as shown in Figure~\ref{fig:colon_coverage}. The coverage of the colon segment, shown in Figure~\ref{fig:colon_coverage}a, increased from 72\% to 94\% using LMap on the detected folds, in Figure~\ref{fig:colon_coverage}c. The coverage of the colon segment, shown in Figure~\ref{fig:colon_coverage}d, is increased from 76\% to 91\% using LMap on the detected folds, in Figure~\ref{fig:colon_coverage}f. The polyps on the colon wall are left intact (distortion-free), making the polyp localization, visualization and diagnosis in the resultant maps as accurate as possible. The polyps can be located on both the colon wall and the folds and hence, shape-preservation during the local mapping is critical to avoid loss of such important information. As demonstrated in Figure~\ref{fig:colon_coverage}g, a fold is detected with a polyp located on it using our algorithm \cite{nadeem2017corresponding}. The detected fold and the polyp located on it are locally deformed using LMap, as shown in Figure~\ref{fig:colon_coverage}h. Due to the angle-preservation and local deformation properties of LMap, we can visualize the true shape of the locally mapped polyp with minimal possible area distortion, as shown in Figure~\ref{fig:colon_coverage}i. The optimal coverage in a single flythrough has a potential to significantly reduce the interpretation time and the associated costs.

We conducted an interview with a VC expert to receive feedback on our current LMap colon visualizations. The expert found the current visualizations with geometric context preservation very promising, especially for training new radiologists on VC systems. The expert reiterated our concern about the current state-of-the-art colon flattening approaches being angle-preserving only which can induce a large area distortion on the flattened surface, making them less desirable for polyp diagnosis. Since LMap only maps local ROIs (folds in the case of VC) the area distortion is much smaller than that induced with a completely flattened colon. 

We use real CT colon data from the publicly available National Institute of Biomedical Imaging and Bioengineering (NIBIB) Image and Clinical Data Repository provided by National Institute of Health (NIH). We perform electronic cleansing incorporating the partial volume effect \cite{wang2006improved}, segmentation with topological simplification \cite{hong2006pipeline}, and reconstruction of the colon surface as a triangular mesh via surface nets \cite{gibson1998constrained} on the original CT images in a pre-processing step. Though the size and resolution of each CT volume varies between clinical datasets, the general data size is approximately 512x512x450 voxels with a resolution of approximately 0.7x0.7x1.0mm.

\subsection{Molecular Surface Visualization}
Solvent excluded molecular surfaces are among the most prevalent visual representations for analyzing molecules, especially in biochemistry and structural biology molecular simulations. Different surface models represent different properties of the molecule and additional information can be overlaid on the surface, for example, to show physico-chemical properties of the underlying atoms \cite{krone2017molecular}. The cavities on the boundary and in the interior of the molecule represent important information about the potential binding sites for other molecules making these cavities critical to study via visualization.  

We demonstrate the effectiveness of LMap on three molecular surface datasets from Protein Data Bank \cite{berman2000protein}, courtesy Krone et al. \cite{krone2017molecular}. The two genus 0 (Figure~\ref{fig:molecular_surface}a) and genus 1 (Figure~\ref{fig:molecular_surface}e) surface datasets were used by Krone et al. \cite{krone2017molecular} but the third genus 65 (Figure~\ref{fig:molecular_surface}i) surface dataset was not; Krone et al. \cite{krone2017molecular} computed the solvent excluded triangular mesh surfaces using the MSMS software by Sanner et al. \cite{sanner1996reduced}. LMap results for the corresponding ROIs (Figures~\ref{fig:molecular_surface}b, \ref{fig:molecular_surface}f and \ref{fig:molecular_surface}j) using standard surface shading and using normal mapped shading from the original mesh are shown in Figure~\ref{fig:molecular_surface}. LMap does not require the removal of interior cavities. It locally parameterizes and deforms only selected ROIs (e.g., boundary cavities) for visualization with minimal angle and area distortion while preserving the overall geometric context.

In contrast, current state-of-the-art molecular surface visualization techniques \cite{krone2017molecular} remove critical information such as channels and fill the resultant holes for using spherical parameterization algorithms which can induce considerable angle and area distortion; higher genus surfaces can induce considerably higher area and angle distortion due to higher loss of interior information. We implemented the two force-directed and parameter-based spherical parameterization algorithms, and the two Lambert equal-area and Plate Carr\'{e}e map projection algorithms, presented in Krone et al. \cite{krone2017molecular}. The results on two molecular genus 0 and genus 1 (with channel removal and hole-filling) surface datasets are shown in Figure~\ref{fig:molecular_surface_map}. The map projections in Figures~\ref{fig:molecular_surface_map}c and \ref{fig:molecular_surface_map}f were used to create space time cube visualizations \cite{krone2017molecular} to visualize the cavities on the boundaries without the need for animation. LMap can similarly be used to visualize the cavities while preserving the geometric context and without removing the critical channel information, as shown in Figure~\ref{fig:molecular_surface_cavities}.
\section{Conclusion and Future Work}
In this work, we presented LMap, a technique to conformally parameterize and deform selected regions of interest on an arbitrary surface. We achieved this using the robust and theoretically-sound extrinsic Ricci flow. We also guarantee the existence of a local map for a selected ROI on an arbitrary surface via the use of the dynamic Ricci flow. The effectiveness and efficacy of our LMap technique was then demonstrated in three challenging use cases: (1) multimodal brain visualization, (2) optimal coverage of virtual colonoscopy flythroughs, and (3) molecular surface visualization.  

In the future, we will incorporate area-preserving techniques into LMap using optimal mass transport for balancing the resultant area- and angle-distortions. Quasi-conformal parameterization \cite{zeng2010supine} allows bounded conformality distortion. In this direction, we will explore extending LMap to quasi-conformal LMaps. Since we can locally parameterize a selected ROI without requiring any global information or spherical constraints \cite{rocha2017decal}, we will also validate our approach in the context of decal mapping. Since LMap allows multimodal brain visualization in the native space, we will integrate subcortical structures into these LMap visualizations for a more comprehensive picture of the brain. In the context of VC, we will conduct a formal study with expert radiologists to test LMap in VC. For molecular surface visualization, we will introduce visualizations to study the evolution of the boundary cavities in place with exploded LMap views of these cavities, and will combine these with context-preserving maps \cite{marino2011context} to visualize the interior cavities, simultaneously.

\ifCLASSOPTIONcompsoc
  \section*{Acknowledgments}
\else
  \section*{Acknowledgment}
\fi

The VC datasets are courtesy of Stony Brook University Hospital (SBUH) and Dr. Richard Choi, Walter Reed Army Medical Center. The brain datasets are courtesy Human Connectome Project, WU-Minn Consortium (PIs: David Van Essen and Kamil Ugurbil) funded by the 16 NIH Institutes and Centers that support the NIH Blueprint for Neuroscience Research; and by the McDonnell Center for Systems Neuroscience at Washington University. The molecular imaging data is courtesy Dr. Michael Krone of Visualization Research Center, University of Stuttgart, Germany. We would like to thank Dr. Matthew Barish and Dr. Kevin Baker of SBUH and Dr. Hoi-Chung Leung of SBU Neuroscience Department for their help in this project. This work has been partially supported by the National Science Foundation grants IIP1069147, CNS1302246, IIS1527200, NRT1633299, CNS1650499, and the Marcus Foundation.


\newpage
\vspace{-12mm}
\begin{IEEEbiography}[{\includegraphics[width=1in,height=1.25in,clip,keepaspectratio]{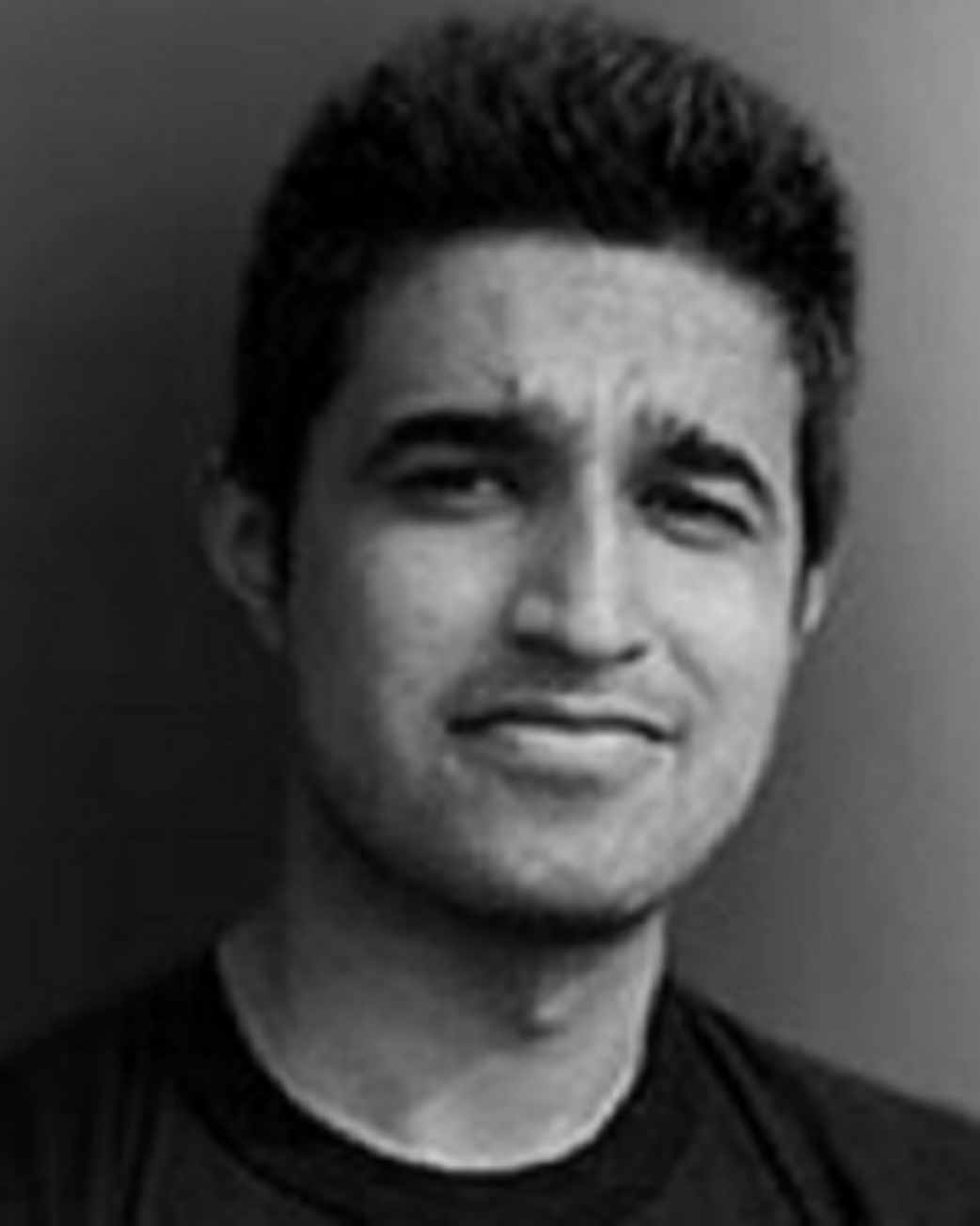}}]{Saad Nadeem}
is a PhD candidate in the Computer Science department, Stony Brook University. He received his BSc Honors in Computer Science and Mathematics from School of Science and Engineering, Lahore University of Management Sciences, Pakistan. His research interests include computer vision, computer graphics, and visualization.
\end{IEEEbiography}

\vspace{-70mm}
\begin{IEEEbiography}[{\includegraphics[width=1in,height=1.25in,clip,keepaspectratio]{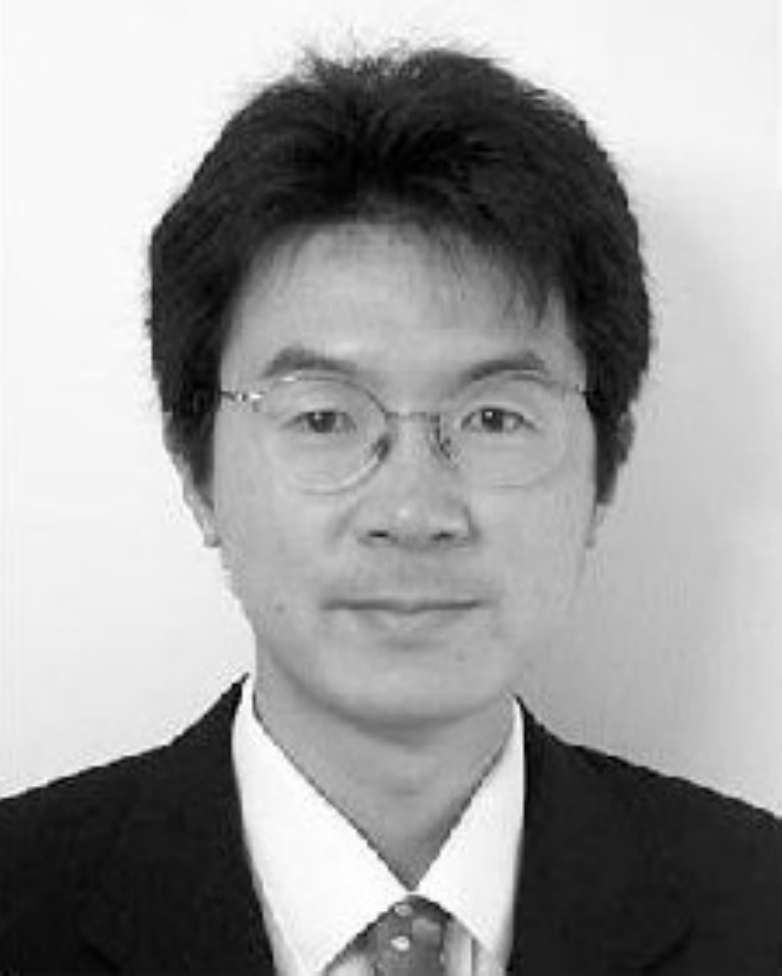}}]{Xianfeng Gu}
received the Ph.D. degree in computer science from Harvard University, Cambridge, MA, USA, in 2003. He is an associate professor of Computer Science and the Director of the 3D Scanning Laboratory with the Department of Computer Science at Stony Brook University, Stony Brook, NY, USA. His current research interests include computer vision, graphics, geometric modeling, and medical imaging. His major works include global conformal surface parameterization in graphics, tracking and analysis of facial expression in vision, manifold splines in modeling, brain mapping and virtual colonoscopy in medical imaging, and computational conformal geometry. He won the U.S. National Science Foundation CAREER Award in 2004.
\end{IEEEbiography}

\vspace{-70mm}
\begin{IEEEbiography}[{\includegraphics[width=1in,height=1.25in,clip,keepaspectratio]{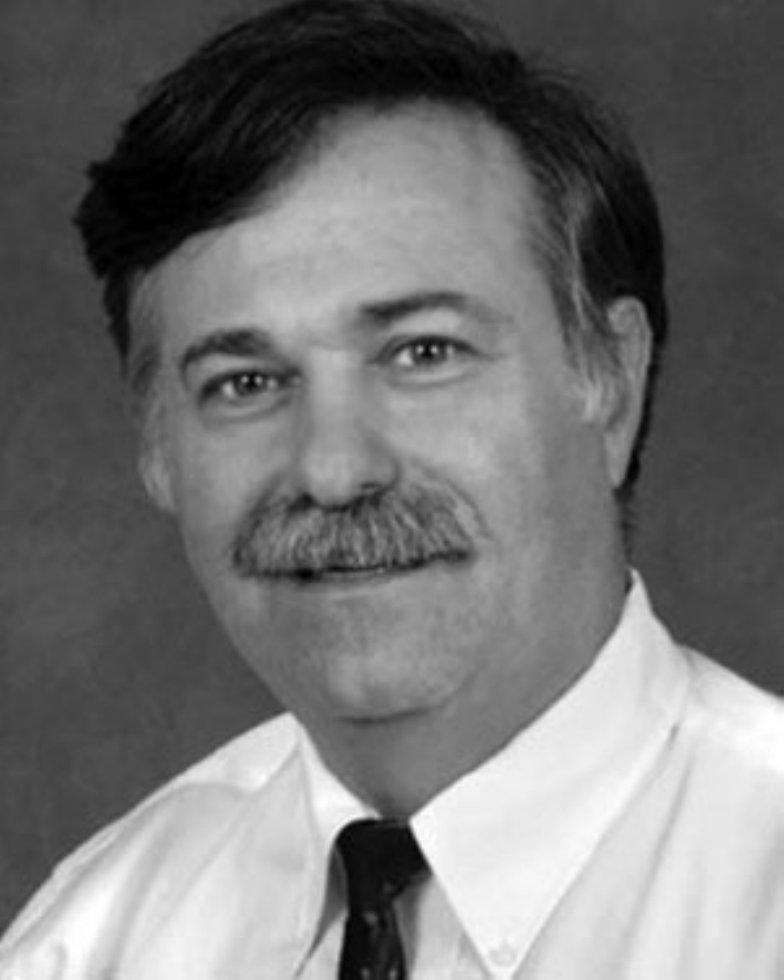}}]{Arie E. Kaufman}
is a Distinguished Professor and Chairman of the Computer Science Department, Director of the Center of Visual Computing (CVC), and Chief Scientist of the Center of Excellence in Wireless and Information Technology (CEWIT) at Stony Brook University. He has conducted research for over 40 years in visualization and graphics and their applications, has published more than 300 refereed papers, has presented more than 20 invited keynote talks, has been awarded/filed more than 40 patents, and has been PI/co-PI on more than 100 grants. He was the founding Editor-in-Chief of IEEE Transaction on Visualization and Computer Graphics (TVCG), 1995-1998. He is a Fellow of IEEE, a Fellow of ACM, the recipient of the IEEE Visualization Career Award (2005), and was inducted into the Long Island Technology Hall of Fame (2013). He received his PhD in Computer Science from Ben-Gurion University, Israel, in 1977.
\end{IEEEbiography}

\end{document}